\documentclass{article}

\usepackage{microtype}
\usepackage{graphicx}
\usepackage{booktabs} 

\usepackage{hyperref}
\usepackage[accepted]{icml2020}

\usepackage[utf8]{inputenc} 
\usepackage[T1]{fontenc}    
\usepackage{hyperref}       
\usepackage{url}            
\usepackage{nicefrac}       
\usepackage{graphicx}
\usepackage{subfigure}
\usepackage{dsfont}
\usepackage{xspace}
\usepackage{amsmath}
\usepackage{amssymb}
\usepackage{dsfont}
\DeclareMathOperator*{\argmax}{arg\,max}

\usepackage{hyperref}
\usepackage{amsfonts}
\usepackage{color}

\usepackage{float}
\usepackage{scalerel,stackengine}
\usepackage{soul}
\usepackage[flushleft]{threeparttable}
\stackMath
\newcommand\reallywidehat[1]{%
\savestack{\tmpbox}{\stretchto{%
  \scaleto{%
    \scalerel*[\widthof{\ensuremath{#1}}]{\kern-.6pt\bigwedge\kern-.6pt}%
    {\rule[-\textheight/2]{1ex}{\textheight}}
  }{\textheight}%
}{0.5ex}}%
\stackon[1pt]{#1}{\tmpbox}%
}
\newcommand{\mX}{\mathcal{X}}

\DeclareMathOperator{\Tr}{Tr}
\newcommand{\SO}[1]{SO($#1$)}
\newcommand{\SE}[1]{SE($#1$)}
\newcommand{\T}[1]{T($#1$)}
\newcommand{\triv}{Trivial}
\DeclareMathOperator{\position}{\mathbf q}
\DeclareMathOperator{\momentum}{\mathbf p}
\DeclareMathOperator{\state}{\mathbf z}

\usepackage[algo2e,ruled,vlined]{algorithm2e}
\include{pythonlisting}
\usepackage{appendix}
\usepackage{amsthm}
\newtheorem{definition}{Definition}
\usepackage{wasysym}
\usepackage{pifont}
\usepackage{bbm}
\usepackage{placeins}

\usepackage{xcolor}
\definecolor{dark-blue}{rgb}{0.15,0.15,0.4}
\definecolor{medium-blue}{rgb}{0,0,0.5}
\hypersetup{
  colorlinks, linkcolor={dark-blue},
  citecolor={dark-blue}, urlcolor={medium-blue}
}

\icmltitlerunning{Generalizing Convolutional Neural Networks for Equivariance to Lie Groups on Arbitrary Continuous Data}

\begin{document}
\twocolumn[
\icmltitle{Generalizing Convolutional Neural Networks for Equivariance \\ to Lie Groups on Arbitrary Continuous Data}

\begin{icmlauthorlist}
\icmlauthor{Marc Finzi}{nyu} \quad
\icmlauthor{Samuel Stanton}{nyu} \quad
\icmlauthor{Pavel Izmailov}{nyu} \quad 
\icmlauthor{Andrew Gordon Wilson}{nyu} \\
\end{icmlauthorlist}

\icmlaffiliation{nyu}{New York University}

\icmlcorrespondingauthor{Marc Finzi}{maf820@nyu.edu}
\icmlcorrespondingauthor{Samuel Stanton}{ss13641@nyu.edu}
\icmlcorrespondingauthor{Pavel Izmalov}{pi390@nyu.edu}
\icmlcorrespondingauthor{Andrew Gordon Wilson}{andrewgw@cims.nyu.edu}

\icmlkeywords{Equivariance, Point Clouds, Hamiltonian Neural Networks, Lie Groups}

\vskip 0.3in
]
\printAffiliationsAndNotice{} 

\begin{abstract}
The translation equivariance of convolutional layers enables convolutional neural networks to generalize well on image problems. While translation equivariance provides a powerful inductive bias for images, we often additionally desire equivariance to other transformations, such as rotations, especially for non-image data.
We propose a general method to construct a convolutional layer that is equivariant to transformations from any specified Lie group with a surjective exponential map. Incorporating equivariance to a new group requires implementing only the group exponential and logarithm maps, enabling rapid prototyping. Showcasing the simplicity and generality of our method, we apply the same model architecture to images, ball-and-stick molecular data, and Hamiltonian dynamical systems. For Hamiltonian systems, the equivariance of our models is especially impactful, leading to exact conservation of linear and angular momentum.
\end{abstract}

\section{Introduction}
Symmetry pervades the natural world. The same law of gravitation governs a game of catch, the orbits of our planets, and the formation of galaxies.
It is precisely because of the order of the universe that we can hope to understand it.  Once we started to understand the symmetries inherent in physical laws, we could predict behavior in galaxies billions of light-years away by studying our own local region of time and space. For statistical models to achieve their full potential, it is essential to incorporate our knowledge of naturally occurring symmetries into the design of algorithms and architectures. An example of this principle is the translation equivariance of convolutional layers in neural networks \citep{lecun1995convolutional}: when an input (e.g. an image) is translated, the output of a convolutional layer is translated in the same way. 

Group theory provides a mechanism to reason about symmetry and equivariance.
Convolutional layers are equivariant to translations, and are a special case of group convolution. A group convolution is a general linear transformation equivariant to a given group, used in group equivariant convolutional networks \citep{cohen2016group}.

\begin{figure}[t]
		\includegraphics[trim=100 250 250 190,clip,height=0.2\textwidth]{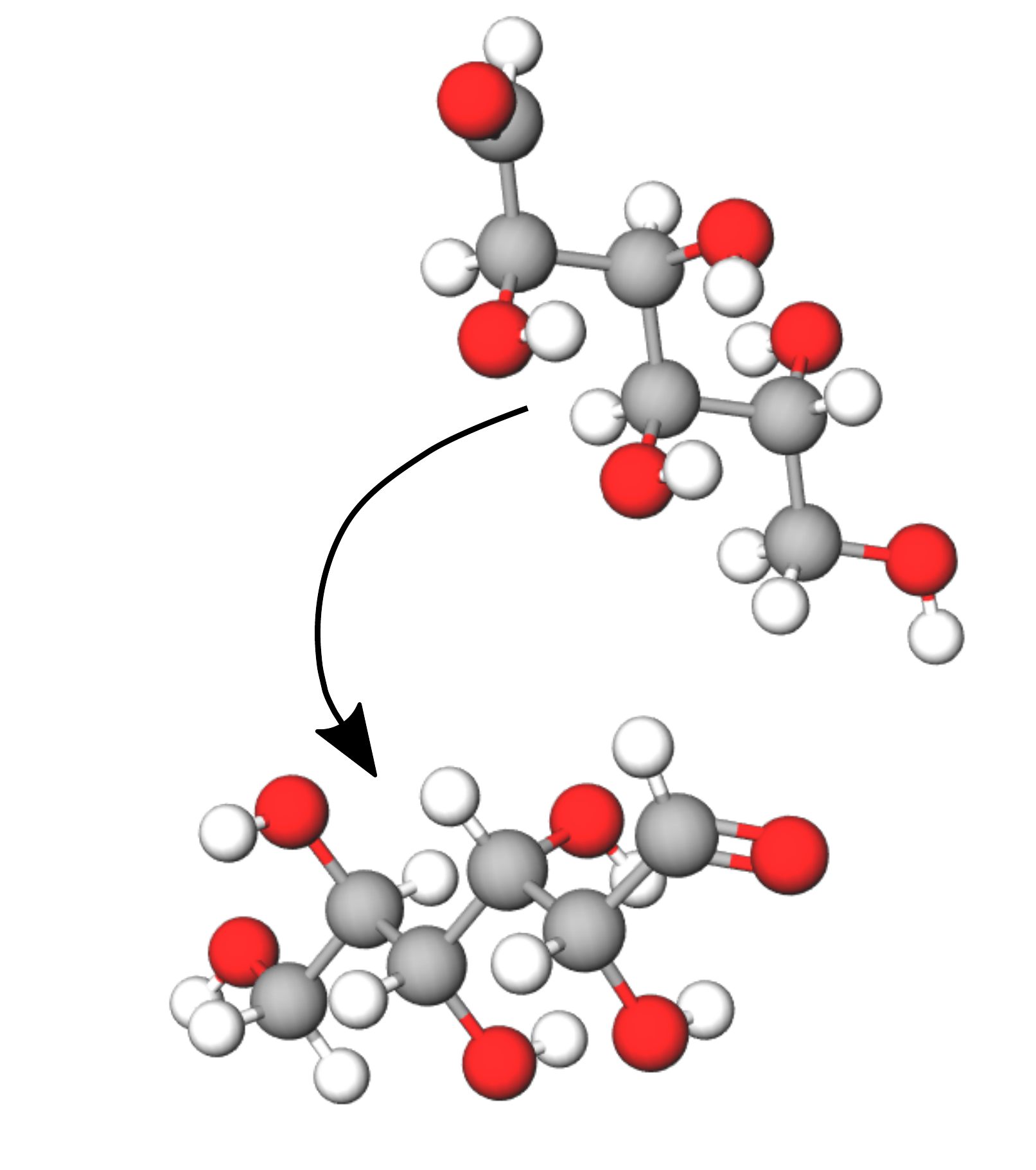}
		\hspace{1mm}
		\includegraphics[height=.2\textwidth]{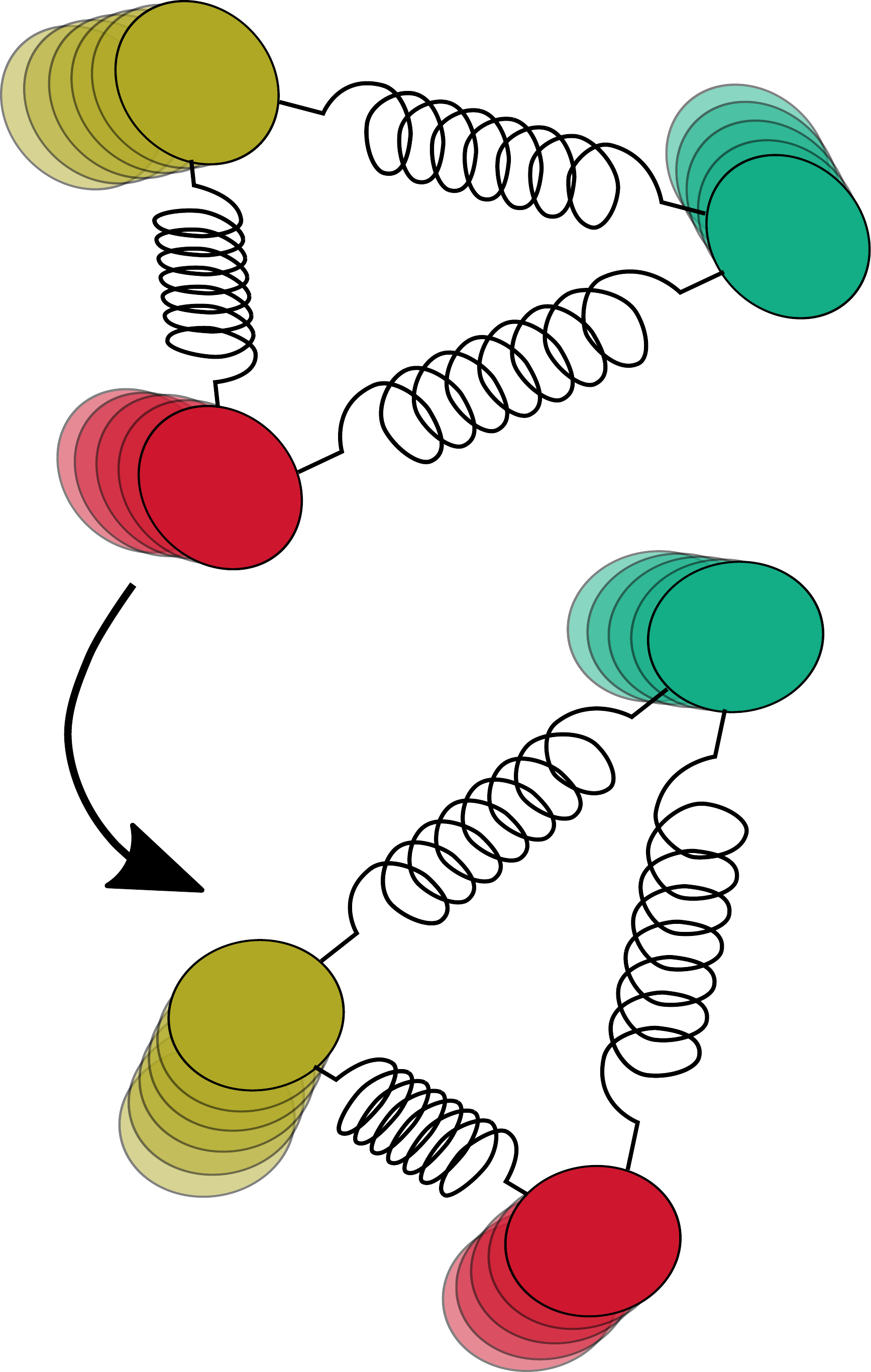}
		\hspace{3mm}
		\includegraphics[trim=50 100 50 100,clip,height=.2\textwidth]{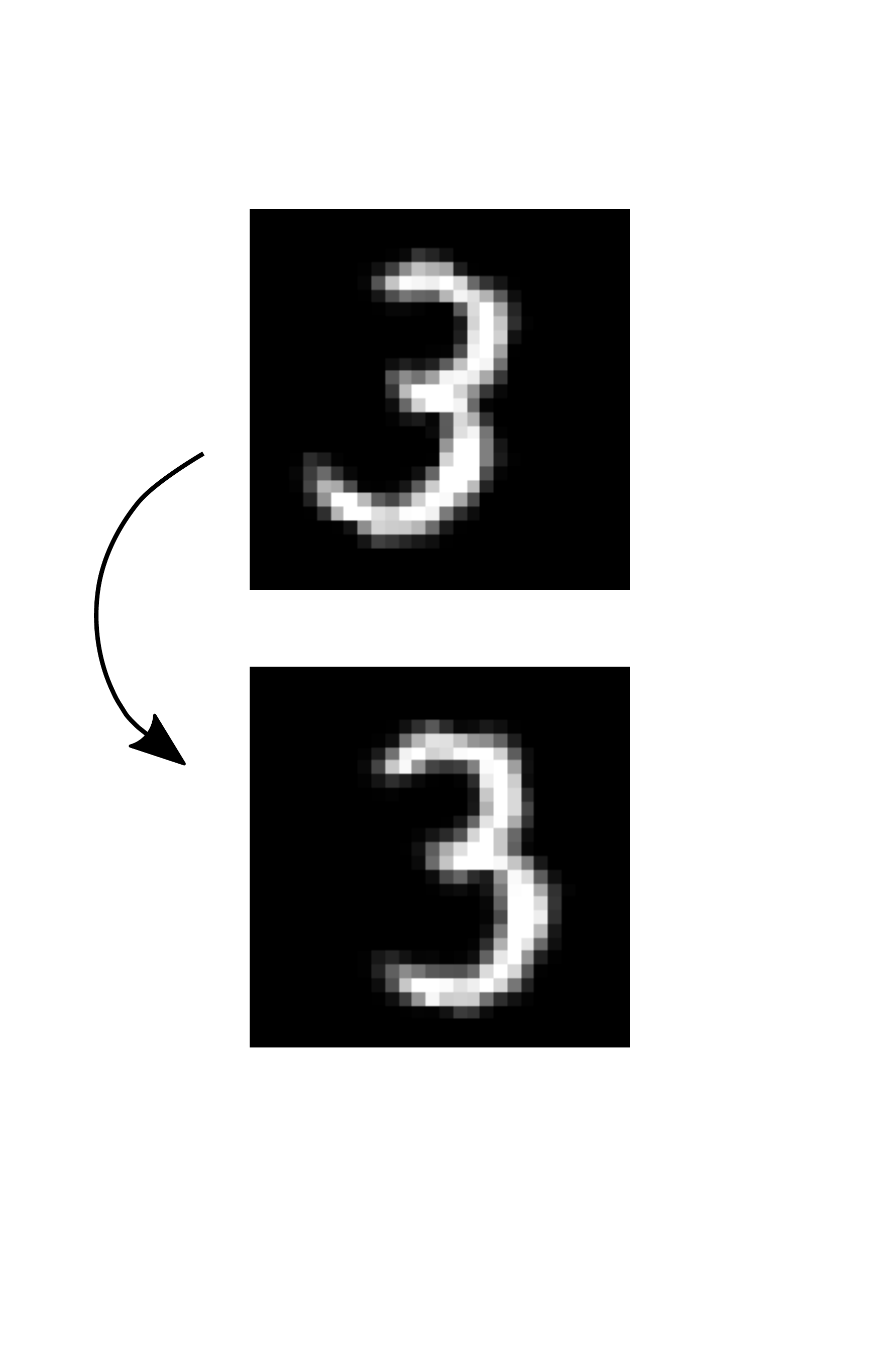}
	\caption{Many modalities of spatial data do not lie on a grid, but still possess important symmetries. We propose a single model to learn from continuous spatial data that can be specialized to respect a given continuous symmetry group.
	}
	\vspace{-4mm}
    \label{fig:intro}
\end{figure}

In this paper, we develop a general framework for equivariant models on arbitrary continuous (spatial) data represented as coordinates and values $\{(x_i, f_i)\}_{i=1}^N$. Spatial data is a broad category, including ball-and-stick representations of molecules, the coordinates of a dynamical system, and images (shown in Figure \ref{fig:intro}). When the inputs or group elements lie on a grid (e.g.,\ image data) one can simply enumerate the values of the convolutional kernel at each group element. But in order to extend to continuous data, we define the convolutional kernel as a continuous function on the group parameterized by a neural network.

We consider the large class of continuous groups known as Lie groups.
In most cases, Lie groups can be parameterized in terms of a vector space of infinitesimal generators (the Lie algebra) via the logarithm and exponential maps. Many useful transformations are Lie groups, including translations, rotations, and scalings. We propose LieConv, a convolutional layer that can be made equivariant to a given Lie group by defining $\exp$ and $\log$ maps. We demonstrate the expressivity and generality of LieConv with experiments on images, molecular data, and dynamical systems. We emphasize that we use the \textit{same network architecture} for all transformation groups and data types. 
LieConv achieves state-of-the-art performance in these domains, even compared to domain-specific architectures. In short, the main contributions of this work are as follows:

\begin{itemize}
    \item We propose LieConv, a new convolutional layer equivariant to transformations from Lie groups. Models composed with LieConv layers can be applied to non-homogeneous spaces and arbitrary spatial data.
    \item We evaluate LieConv on the image classification benchmark dataset rotMNIST \citep{larochelle2007empirical}, and the regression benchmark dataset QM9 \citep{blum, rupp}. LieConv outperforms state-of-the-art methods on some tasks in QM9, and in all cases achieves competitive results.
    \item We apply LieConv to modeling the Hamiltonian of physical systems, where equivariance corresponds to the preservation of physical quantities (energy, angular momentum, etc.). LieConv outperforms state-of-the-art methods for the modeling of dynamical systems.
\end{itemize}
We make code available at \\ \url{https://github.com/mfinzi/LieConv}

\section{Related Work}

One approach to constructing equivariant CNNs, first introduced in \citet{cohen2016group}, is to use standard convolutional kernels and transform them or the feature maps for each of the elements in the group. For discrete groups this approach leads to exact equivariance and uses the so-called regular representation of the group \citep{cohen2019general}. This approach is easy to implement, and has also been used when the feature maps are vector fields \citep{zhou2017oriented,marcos2017rotation}, and with other representations \citep{cohen2016steerable}, but only on image data where locations are discrete and the group cardinality is small. This approach has the disadvantage that the computation grows quickly with the size of the group, and some groups like 3D rotations cannot be easily discretized onto a lattice that is also a subgroup.

Another approach, drawing on harmonic analysis, finds a basis of equivariant functions and parametrizes convolutional kernels in that basis \citep{worrall2017harmonic,weiler2019general,jacobsen2017dynamic}. These kernels can be used to construct networks that are exactly equivariant to continuous groups.
While the approach has been applied on general data types like spherical images \citep{esteves2018learning,cohen2018spherical}, voxel data \citep{weiler20183d}, and point clouds \citep{thomas2018tensor,anderson2019cormorant}, the requirement of working out the representation theory for the group can be cumbersome and is limited to compact groups.
Our approach reduces the amount of work to implement equivariance to a new group, enabling rapid prototyping.

There is also work applying Lie group theory to deep neural networks.
\citet{huang2017deep} define a network where the intermediate activations of the network are 3D rotations representing skeletal poses and embed elements into the Lie algebra using the $\log$ map. \citet{bekkers2019b} use the $\log$ map to express an equivariant convolution kernel through the use of B-splines, which they evaluate on a grid and apply to image problems. While similar in motivation, their method is not readily applicable to point data and can only be used when the equivariance group acts transitively on the input space. Both of these issues are addressed by our work.

\section{Background}

\subsection{Equivariance}

A mapping $h(\cdot)$ is equivariant to a set of transformations $G$ if
when we apply any transformation $g$ to the input of $h$, the output is also transformed by $g$.
The most common example of equivariance in deep learning is the translation equivariance of convolutional
layers: if we translate the input image by an integer number of pixels in $x$ and $y$, the output is also translated by the same amount (ignoring the regions close to the boundary of the image). Formally, if $h: A \rightarrow A$, and $G$ is a set of transformations acting on $A$, we say $h$ is equivariant to $G$ if $\forall a \in A$, $\forall g \in G$,
\begin{align}
    h(g a) = g h(a).
\end{align}
The continuous convolution of a function $f: \mathbb{R} \to \mathbb{R}$ with the kernel $k: \mathbb{R} \to \mathbb{R}$ is equivariant to translations in the sense that $L_t(k*f) = k*L_tf$ where $L_t$ translates the function by $t$: $L_tf(x) = f(x-t)$. 

It is easy to construct invariant functions, where transformations on the input do not affect the output, by simply discarding information. 
Strict invariance unnecessarily limits the expressive power by discarding relevant information, and instead it is necessary to use equivariant transformations that preserve the information.

\subsection{Groups of Transformations and Lie Groups}

Many important sets of transformations form a group. To form a group the set must be closed under composition, include an identity transformation, each element must have an inverse, and composition must be associative.
The set of 2D rotations, \SO{2}, is a simple and instructive example. Composing two rotations $r_1$ and $r_2$, yields another rotation $r = r_2 \circ r_1$. There exists an identity $\textrm{id} \in G$ that maps every point in  $\mathbb R^2$ to itself (i.e., rotation by a zero angle).
And for every rotation $r$, there exists an inverse rotation $r^{-1}$ such that $r \circ r^{-1} = r^{-1} \circ r = \textrm{id}$. Finally, the composition of rotations is an \textit{associative} operation: $(r_1 \circ r_2) \circ r_3 = r_1 \circ (r_2 \circ r_3)$. Satisfying these conditions, \SO{2} is indeed a group.

We can also adopt a more familiar view of \SO{2} in terms of angles, where a rotation matrix $R:\mathbb{R} \rightarrow \mathbb{R}^{2 \times 2}$ is parametrized as $R(\theta) = \exp(J\theta)$. $J$ is the antisymmetric matrix $J = \begin{bmatrix}0 & -1 \\1 & 0 \\\end{bmatrix}$ (an infinitesimal generator of the group) and $\exp$ is the matrix exponential. Note that $\theta$ is totally unconstrained. Using $R(\theta)$ we can add and subtract rotations. Given $\theta_1$, $\theta_2$ we can compute $R(\theta_1)^{-1}R(\theta_2) = \exp(-J\theta_1+J\theta_2) = R(\theta_2-\theta_1)$. $R(\theta) = \exp(J\theta)$ is an example of the Lie algebra parametrization of a group, and
\SO{2} forms a Lie group.

More generally, a Lie group is a group whose elements form a smooth manifold. Since $G$ is not necessarily a vector space, we cannot add or subtract group elements. However, the Lie algebra of $G$, the tangent space at the identity, $\mathfrak{g} = T_{\mathrm{id}} G$, \emph{is} a vector space and can be understood informally as a space of infinitesimal transformations from the group. As a vector space, one can readily expand elements in a basis $A = \sum_k a^ke_k$ and use the components for calculations.

The exponential map $\mathrm{exp}: \mathfrak{g} \to G$ gives a mapping from the Lie algebra to the Lie group, converting infinitesimal transformations to group elements. In many cases, the image of the exponential map covers the group, and an inverse mapping $\mathrm{log}: G \to \mathfrak{g}$ can be defined.  For matrix groups the $\mathrm{exp}$ map coincides with the matrix exponential ($\mathrm{exp}(A) = I + A + A^2/2! + ... \;$), and the $\mathrm{log}$ map with the matrix logarithm. Matrix groups are particularly amenable to our method because in many cases the $\mathrm{exp}$ and $\mathrm{log}$ maps can be computed in closed form. For example, there are analytic solutions for the translation group \T{d}, the $3D$ rotation group \SO{3}, the translation and rotation group \SE{d} for $d=2,3$, the rotation-scale group $\mathbb{R}^* \times$ \SO{2}, and many others \citep{eade2014lie}. In the event that an analytic solution is not available there are reliable numerical methods at our disposal \citep{moler2003nineteen}.

\subsection{Group Convolutions}
\label{sec:groupconv}

Adopting the convention of left equivariance, one can define a \textit{group convolution} between two functions on the group, which generalizes the translation equivariance of convolution to other groups:
\begin{definition} \label{def:group_convolution}
    Let $k, f: G \rightarrow \mathbb{R}$, and $\mu(\cdot)$ be the Haar measure on $G$. For any $u \in G$, the convolution of $k$ and $f$ on $G$ at $u$ is given by
    \begin{equation} \label{eq:group_convolution}
    h(u)= (k*f)(u) = \int_G k(v^{-1}u) f(v) d\mu(v).
    \end{equation}
\end{definition}
\citep{kondor2018generalization,cohen2019general}

\subsection{PointConv Trick}
\label{sec:pointconv}

In order to extend learnable convolution layers to point clouds, not having the regular grid structure in images, 
\citet{dai2017deformable}, \citet{dynamicedge}, and \citet{wu2019pointconv}
 go back to the continuous definition of 
a convolution for a single channel between a learned function (convolutional filter) 
$k_\theta(\cdot): \mathbb{R}^d \to \mathbb{R}^{c_{out} \times c_{in}}$ and an input feature map 
$f(\cdot): \mathbb{R}^d \to \mathbb{R}^{c_{in}}$ yielding the function $h(\cdot): \mathbb{R}^d \to \mathbb{R}^{c_{out}}$,
\begin{equation}
    h(x) = (k_\theta * f)(x) =  \int k_\theta(x-y) f(y) dy.
\end{equation}
We approximate the integral using a discretization:
\begin{equation}
    h(x_i) = (V/n)\sum_{j} k_\theta(x_i-x_j)f(x_j) \,.
\end{equation}

Here $V$ is the volume of the space integrated over and $n$ is the number of quadrature points. In a $3\times 3$ convolutional layer for images, where points fall on a uniform square grid, the filter
$k_\theta$ has independent parameters for each of the inputs $(-1,-1),(-1,0),\dots,(1,1)$.
In order to accommodate points that are not on a regular grid, $k_\theta$ can be parametrized as a small neural network, mapping input offsets to filter matrices, explored with MLPs in \citet{dynamicedge}. The compute and memory costs has severely limited this approach, for typical CIFAR-10 images with $\mathrm{batch size}=32, N=32\times 32, c_{in}=c_{out}=256, n=3\times3$, evaluating a single layer requires computing $20$ billion values for $k$.

In PointConv, \citet{wu2019pointconv} develop a trick where clever reordering of the computation 
cuts memory and computational requirements by $\sim 2$ orders of magnitude, allowing them to scale to the point cloud classification, segmentation datasets ModelNet40 and ShapeNet, and the image dataset CIFAR-10. We review and generalize the Efficient-PointConv trick in 
Appendix \ref{sec:generalized_pointconv}, which we will use to accelerate our method.

\begin{figure*}[h]
	\centering
	\hspace{-0.6cm}
	\subfigure[Data]{
		\includegraphics[width=0.2\textwidth]{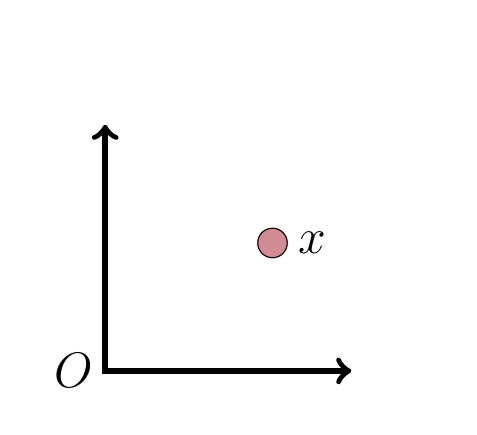}
	}\hspace{-1.3cm}
	\subfigure[Trivial]{
		\includegraphics[width=0.2\textwidth]{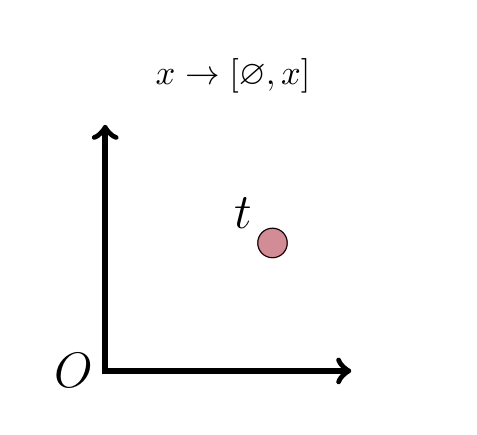}
	}\hspace{-1.3cm}
	\subfigure[$T(2)$]{
		\includegraphics[width=0.2\textwidth]{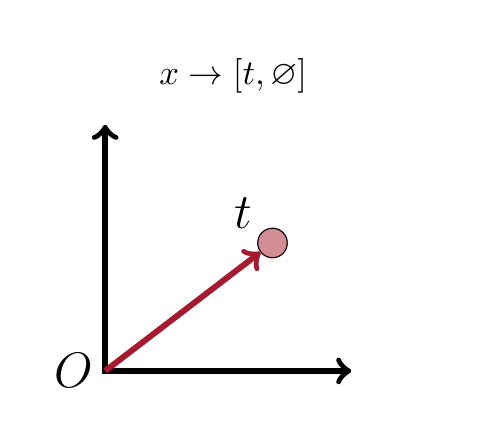}
	}
	\hspace{-1.3cm}
	\subfigure[$SO(2)$]{
		\includegraphics[width=0.2\textwidth]{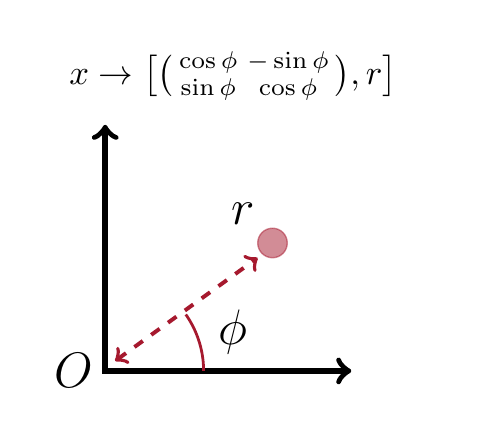}
	}
	\hspace{-0.8cm}
	\subfigure[$\mathbb{R}^*\times SO(2)$]{
		\includegraphics[width=0.2\textwidth]{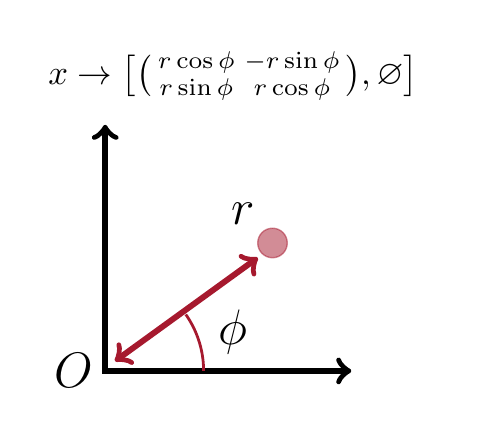}
	}
    \hspace{-0.8cm}
	\subfigure[$SE(2)$]{
		\includegraphics[width=0.23\textwidth]{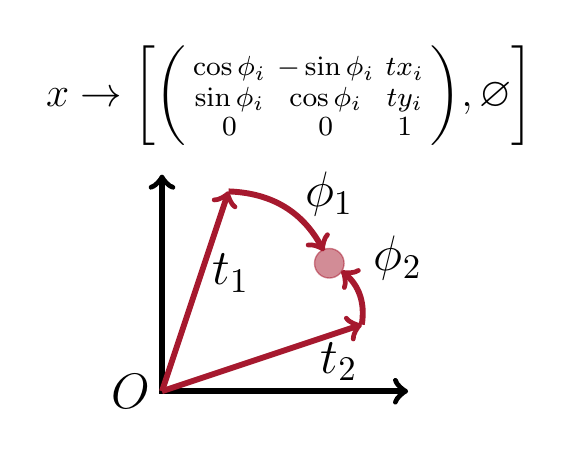}
	}
	\caption{
		Visualization of the lifting procedure. Panel \textbf(a) shows a point $x$ in the original input space $\mathcal X$.
		In panels \textbf{(b)}--\textbf{(f)} we illustrate the lifted embeddings for different groups in the form $[u, q]$, where $u \in G$ is an element of the group and $q \in \mathcal X / G$ identifies the orbit (see Section \ref{sec:nonhomogeneous}). For \SE{2} the lifting is multi-valued.
        }
        \label{fig:lifting}
\end{figure*}

\section{Convolutional Layers on Lie Groups}\label{sec:lie_group_convolutions}

We now introduce LieConv, a new convolutional layer that can be made equivariant to a given Lie group. Models with LieConv layers can act on arbitrary collections of coordinates and values  $\{(x_i,f_i)\}_{i=1}^N$, for $x_i \in \mX$ and $f_i \in V$ where $V$ is a vector space. The domain $\mathcal{X}$ is usually a low dimensional domain like $\mathbb{R}^2$ or $\mathbb{R}^3$ such as for molecules, point clouds, the configurations of a mechanical system, images, time series, videos, geostatistics, and other kinds of spatial data.

We begin with a high-level overview of the method. In Section \ref{sec:lifting} we discuss transforming raw inputs $x_i$ into group elements $u_i$ on which we can perform group convolution. We refer to this process as \textit{lifting}. Section \ref{sec:filter} addresses the irregular and varied arrangements of group elements that result from lifting arbitrary continuous input data by parametrizing the convolutional kernel $k$ as a neural network. In Section \ref{sec:locality}, we show how to enforce the locality of the kernel by defining an invariant distance on the group. In Section \ref{sec:discretization}, we define a Monte Carlo estimator for the group convolution integral in Eq.~\eqref{eq:group_convolution} and show that this estimator is equivariant in distribution. In Section \ref{sec:nonhomogeneous}, we extend the procedure to cases where the group does not act transitively on the input space (when we cannot map any point to any other point with a transformation from the group). Additionally, in Appendix \ref{sec:abelian}, we show that our method generalizes coordinate transform equivariance when $G$ is Abelian. At the end of Section~\ref{sec:nonhomogeneous} we provide a concise algorithmic description of the lifting procedure and our new convolution layer.

\subsection{Lifting from $\mX$ to $G$}
\label{sec:lifting}
If $\mX$ is a homogeneous space of $G$, then every two elements in $\mX$ are connected by an element in $G$, and one can lift elements by simply picking an origin $o$ and defining $\textrm{Lift}(x) = \{u\in G: uo = x\}$: all elements in the group that map the origin to $x$. This procedure enables lifting tuples of coordinates and features $\{(x_i,f_i)\}_{i=1}^N \rightarrow \{(u_{ik},f_i)\}_{i=1,k=1}^{N,K}$, with up to $K$ group elements for each input.\footnote{When $f_i=f(x_i)$, lifting in this way is equivalent to defining $f^\uparrow(u) = f(uo)$ as in \citet{kondor2018generalization}.} To find all the elements $\{u\in G: uo = x\}$, one simply needs to find one element $u_x$ and use the elements in the stabilizer of the origin $H = \{h\in G: ho=o\}$, to generate the rest with $\textrm{Lift}(x) = \{u_xh \textrm{ for } h \in H\}$. For continuous groups the stabilizer may be infinite, and in these cases we sample uniformly using the Haar measure $\mu$ which is described in Appendix \ref{subsec:haar}. We visualize the lifting procedure for different groups in Figure \ref{fig:lifting}.

\subsection{Parameterization of the Kernel}
\label{sec:filter}
The conventional method for implementing an equivariant convolutional network \citep{cohen2016group} requires enumerating the values of $k(\cdot)$ over the elements of the group, with separate parameters for each element. This procedure is infeasible for irregularly sampled data and problematic even for a discretization because there is no generalization between different group elements. Instead of having a discrete mapping from each group element to the kernel values, we parametrize the convolutional kernel as a continuous function $k_\theta$ using a fully connected neural network with Swish activations, varying smoothly over the elements in the Lie group.

However, as neural networks are best suited to learn on euclidean data and $G$ does not form a vector space, we propose to model $k$ by mapping onto the Lie Algebra $\mathfrak{g}$, which \emph{is} a vector space, and expanding in a basis for the space. To do so, we restrict our attention in this paper to Lie groups whose exponential maps are surjective, where every element has a logarithm. This means defining $k_\theta(u) = (k\circ\exp)_\theta(\log u)$, 
where $\tilde{k}_\theta = (k\circ\exp)_\theta$  is the function parametrized by an MLP, $\tilde{k}_\theta: \mathfrak{g} \to \mathbb{R}^{c_{out}\times c_{in}}$. Surjectivity of the $\exp$ map guarantees that $\exp \circ \log = \textrm{id}$, although not in the other order. 

\subsection{Enforcing Locality}
\label{sec:locality}
Important both to the inductive biases of convolutional neural networks and their computational efficiency is the fact that convolutional filters are local, $k_\theta(u_i-u_j) =0$ for $\|u_i-u_j\|>r$. In order to quantify locality on matrix groups, we introduce the function:
\begin{equation}
    d(u,v) := \|\log(u^{-1}v)\|_F,
\end{equation}
where $\log$ is the matrix logarithm, and $F$ is the Frobenius norm. The function is left invariant, since $d(wu,wv) = \|\log(u^{-1}w^{-1}wv)\|_F = d(u,v)$, and is a semi-metric (it does not necessarily satisfy the triangle inequality). In Appendix \ref{subsec:suff_cond_geodesic} we show the conditions under which $d(u,v)$ is additionally the distance along the geodesic connecting $u,v$
, a generalization of the well known formula for the geodesic distance between rotations $\|\log(R_1^TR_2)\|_F$ \citep{kuffner2004effective}.

\begin{figure}[t]
    \centering
    \includegraphics[trim=10 10 10 10,clip,width=0.1\textwidth]{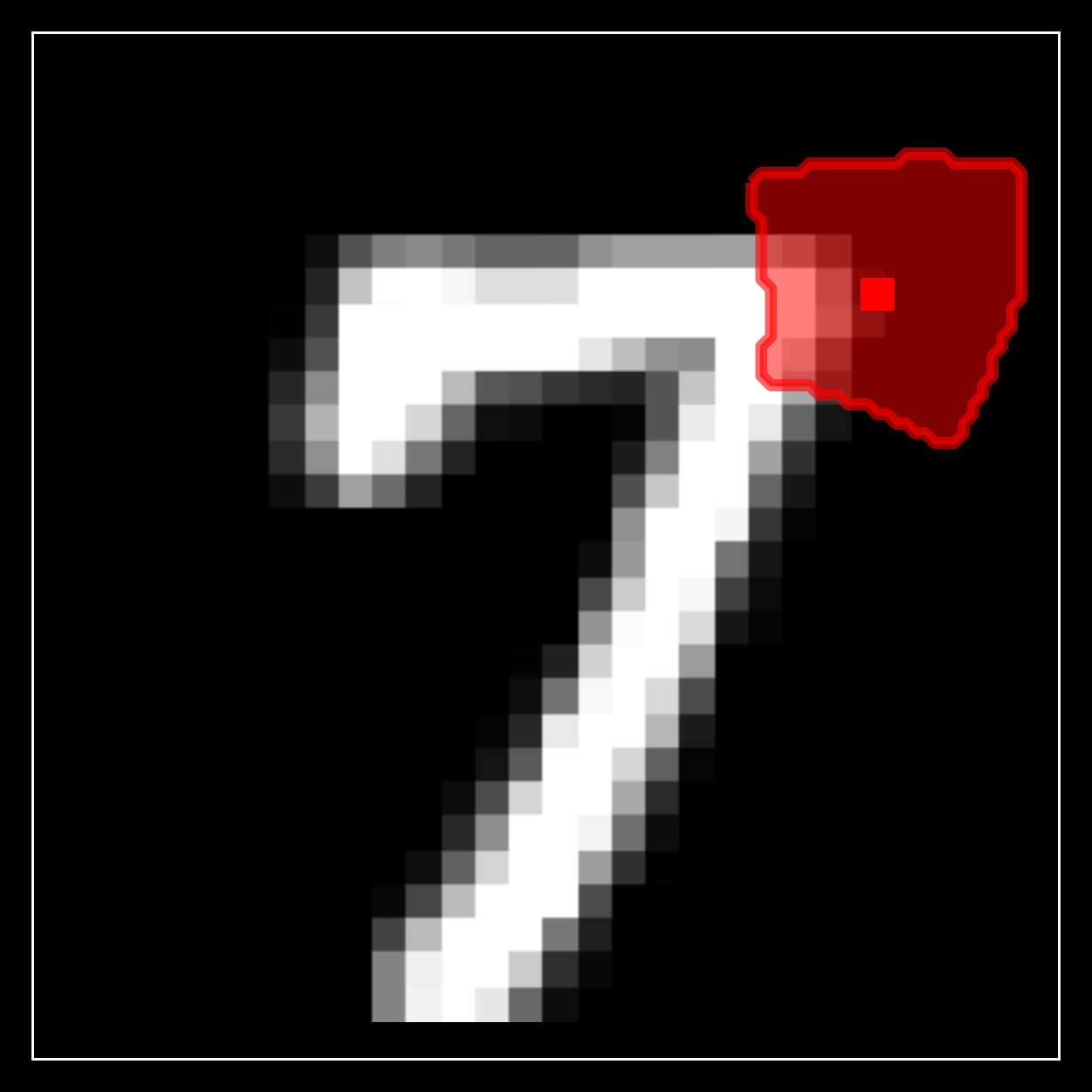}
    ~~
    \includegraphics[trim=10 10 10 10,clip,width=0.1\textwidth]{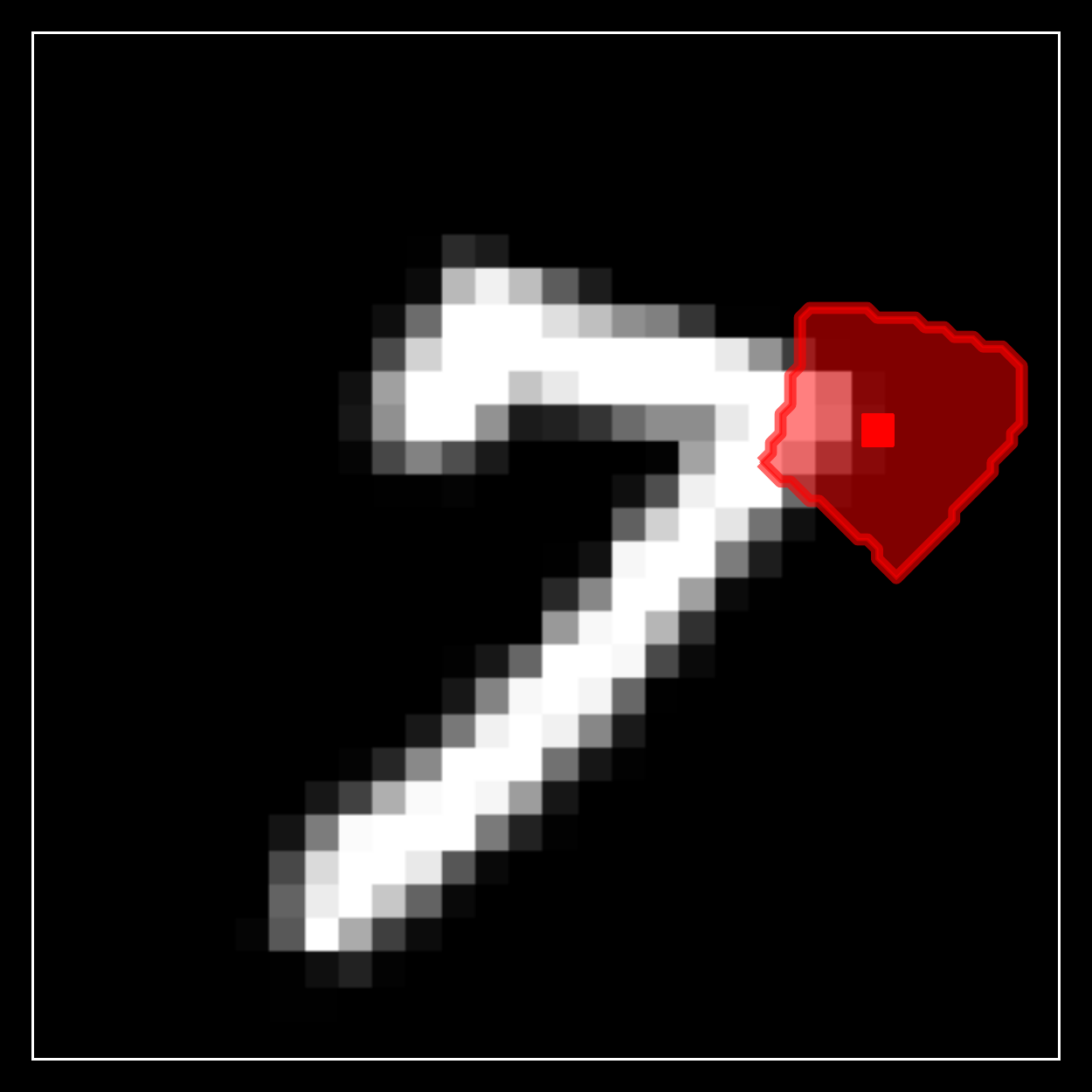}
    ~~
    \includegraphics[trim=10 10 10 10,clip,width=0.1\textwidth]{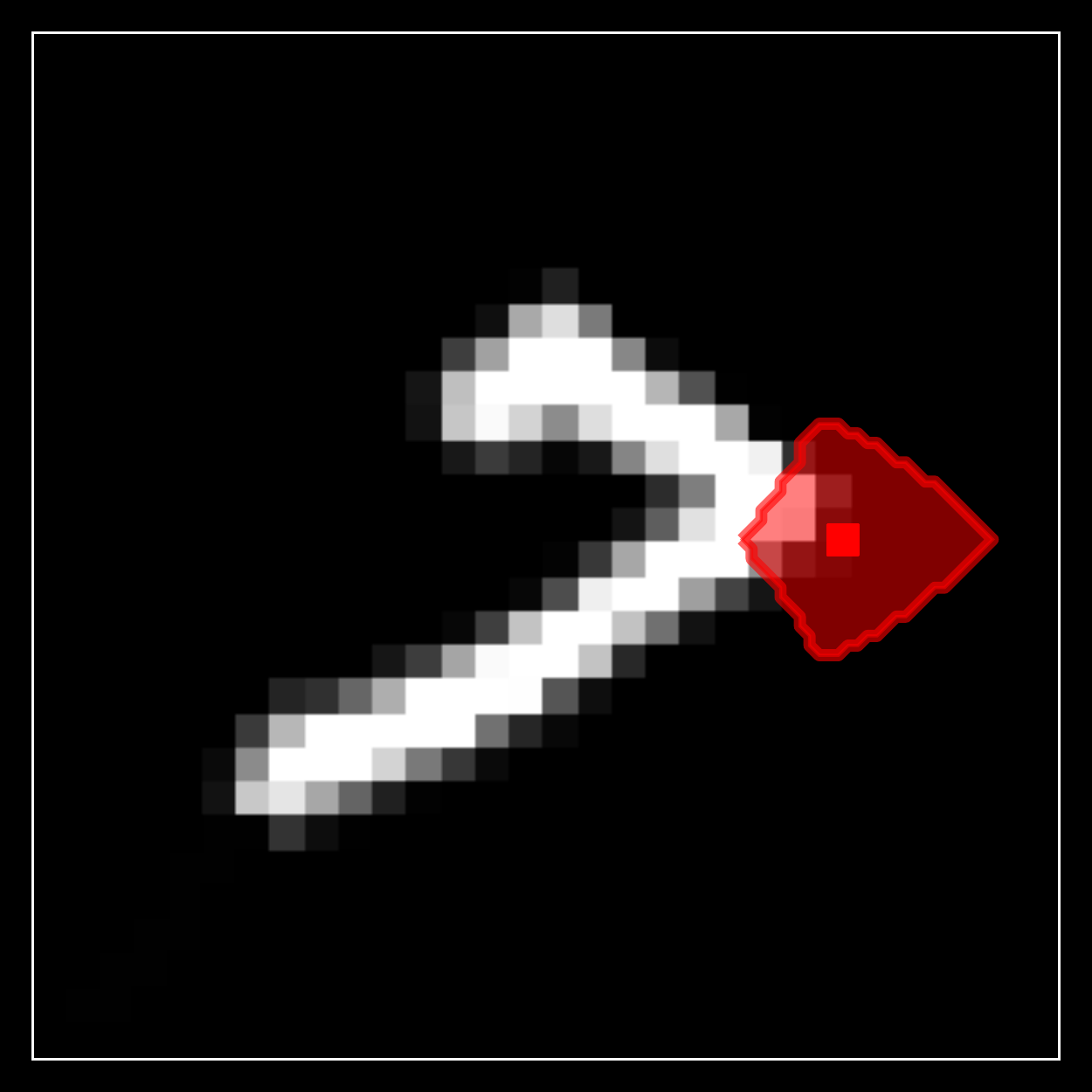}
    ~~
    \includegraphics[trim=10 10 10 10,clip,width=0.1\textwidth]{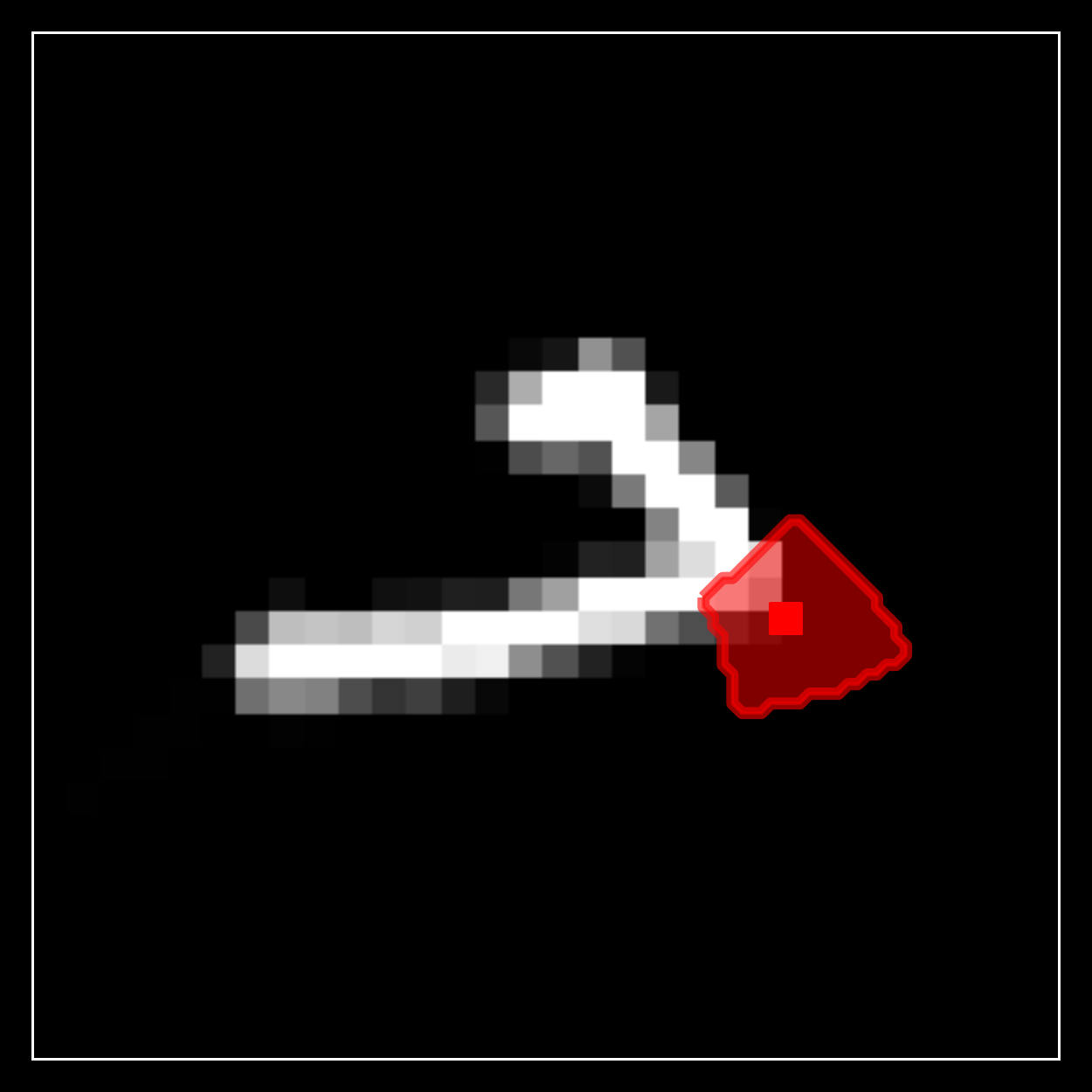}
    \caption{A visualization of the local neighborhood for $\mathbb{R}^* \times$ \SO{2}, in terms of the points in the input space. 
    For the computation of $h$ at the point in orange, elements are sampled from colored region. 
    Notice that the same points enter the calculation when the image is transformed by a rotation and scaling.
    We visualize the neighborhoods for other groups in Appendix \ref{sec:app_nbhd}.
    }
    \label{fig:nbhd}
\end{figure}

To enforce that our learned convolutional filter $k$ is local, we can use our definition of distance to only evaluate the sum for $d(u,v) < r$, implicitly setting $k_\theta(v^{-1}u) = 0$ outside a local neighborhood $\mathrm{nbhd}(u) = \{v: d(u,v)\le r\}$,
\begin{equation}
    h(u) = \int_{v \in \mathrm{nbhd}(u)} k_\theta(v^{-1}u) f(v)d\mu(v).
    \label{eq:lieconv}
\end{equation}
This restriction to a local neighborhood does not break equivariance precisely because $d(\cdot,\cdot)$ is left invariant. Since $d(u,v) = d(v^{-1}u,\mathrm{id})$ this restriction is equivalent to multiplying by the indicator function $k_\theta(v^{-1}u) \to k_\theta(v^{-1}u)\mathds{1}_{[d(v^{-1}u,\mathrm{id})\le r]}$ which depends only on $v^{-1}u$. Note that equivariance \emph{would} have been broken if we used neighborhoods that depend on fixed regions in the input space like the square $3\times3$ region. Figure \ref{fig:nbhd} shows what these neighborhoods look like in terms of the input space.

\subsection{Discretization of the Integral}
\label{sec:discretization}

Assuming that we have a collection of quadrature points $ \{v_j\}_{j=1}^N$ as input
and the function $f_j = f(v_j)$ evaluated at these points, we can judiciously choose to evaluate the convolution at another set of group elements $\{u_i\}_{i=1}^N$, so as to have a set of quadrature points to approximate an integral in a subsequent layer. Because we have restricted the integral \eqref{eq:lieconv} to the compact neighbourhood $\mathrm{nbhd}(u)$, we can define a proper sampling distribution $\mu|_{\mathrm{nbhd}(u)}$ to estimate the integral,
unlike for the possibly unbounded $G$. Computing the outputs only at these target points, we use the Monte Carlo estimator for \eqref{def:group_convolution} as
\begin{equation} \label{eq:group_conv_discretization}
    h(u_i) = (k \ \reallywidehat{*} \ f)(u_i) = \frac{1}{n_i}\sum_{j \in \mathrm{nbhd}(i)} k(v_j^{-1}u_i) f(v_j),
\end{equation}
where $n_i=|\mathrm{nbhd}(i)|$ the number of points sampled in each neighborhood.

For $v_j \sim \mu|_{\mathrm{nbhd}(u)}(\cdot)$, the Monte Carlo estimator is equivariant (in distribution).\\\\
\textbf{Proof:} Recalling that we can absorb the local neighborhood into the definition of $k_\theta$ using an indicator function, we have
\begin{align*}
\begin{split}
        (k \ \reallywidehat{*} \ L_wf)(u_i) &= (1/n_i)\sum_{j} k(v_j^{-1}u_i)f(w^{-1}v_j)\\
        &= (1/n_i)\sum_{j} k(\tilde{v}_j^{-1}w^{-1}u_i)f(\tilde{v}_j)\\
        &\overset{d}{=} (k\ \reallywidehat{*} \ f)(w^{-1}u_i) = L_w(k\ \reallywidehat{*} \ f)(u_i).\\
\end{split}
\end{align*}
Here $\tilde{v}_j := wv_j$, and the last line follows from the fact that the random variables $wv_j\overset{d}{=}v_j$ are equal in distribution because they are sampled from the Haar measure with property $d\mu(wv) = d\mu(v)$. The equivariance also holds deterministically when the sampling locations are transformed along with the function $v_j \to wv_j$.
Now that we have the discretization $h_i = (1/n_i)\sum_{j \in \mathrm{nbhd}(i)} \tilde{k}_\theta (\log(v_j^{-1}u_i))f_i$, we can accelerate this computation using the Efficient-PointConv trick, with the argument of $a_{ij} = \log(v_j^{-1}u_i)$ for the MLP. 
See Appendix \ref{sec:generalized_pointconv} for more details.
Note that we can also apply this discretization of the convolution when the inputs are not functions $f_i=f(x_i)$, but simply coordinates and values $\{(x_i,f_i)\}_{i=1}^N$, and the mapping $\{(u_i,f_i)\}_{i=1}^N \rightarrow \{(u_i,h_i)\}_{i=1}^N$ is still equivariant, which we also demonstrate empirically in Table \ref{table:transformed_test}. We also detail two methods for equivariantly subsampling the elements to further reduce the cost in Appendix \ref{sec:equivariant_subsampling}. 

\subsection{More Than One Orbit?}\label{sec:nonhomogeneous}

\begin{figure}[ht]
    \centering
  \includegraphics[width=0.21\textwidth]{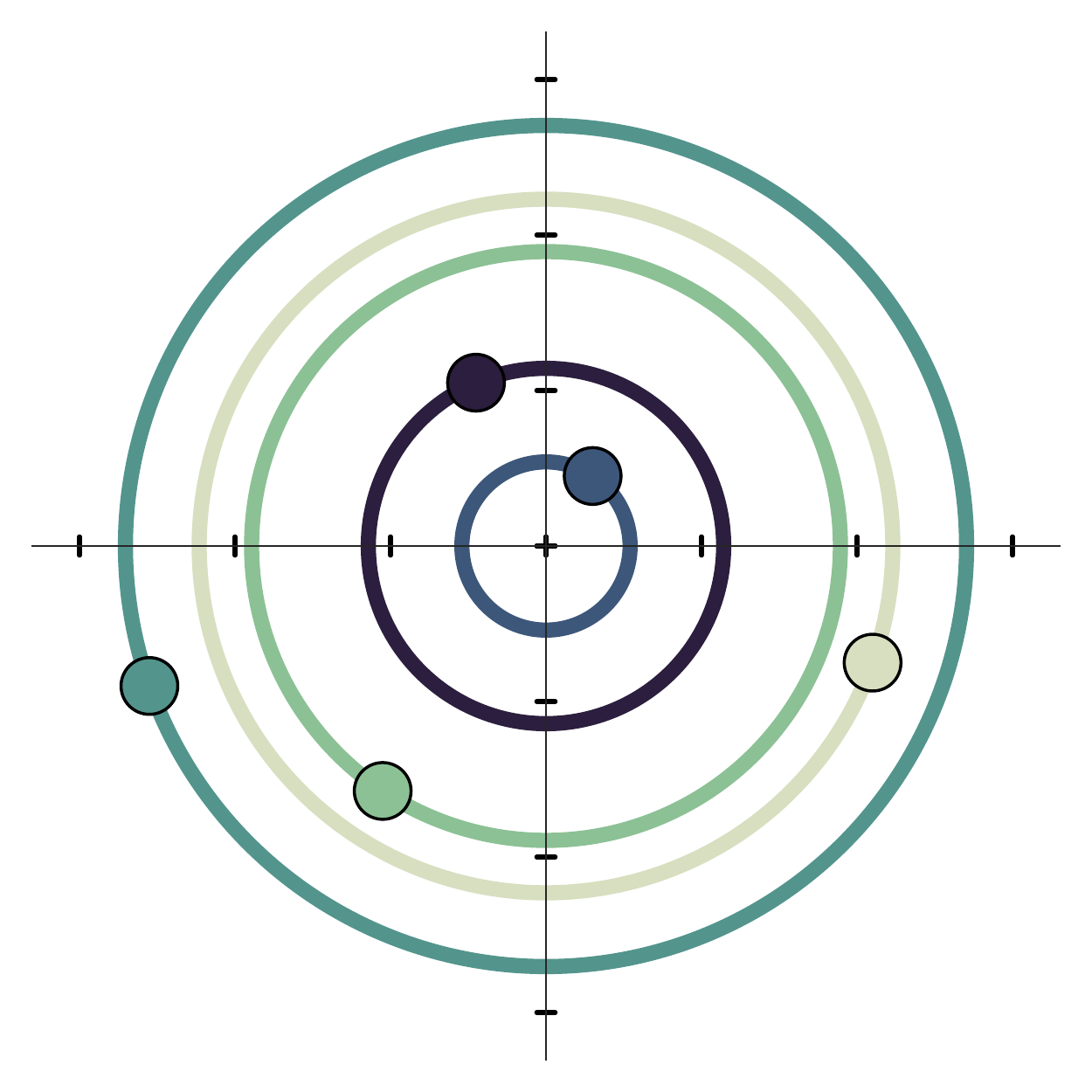}
  \quad
    \includegraphics[width=0.21\textwidth]{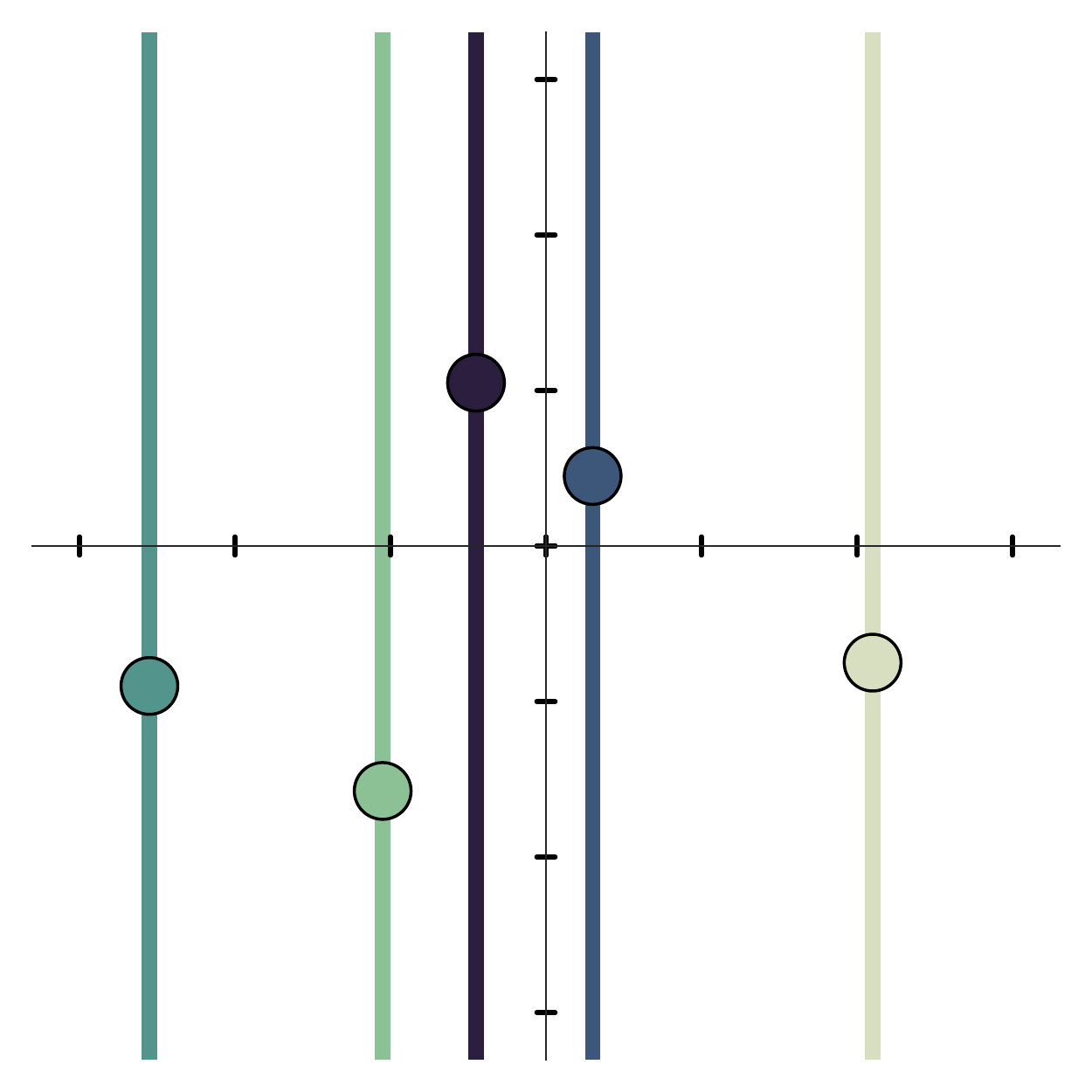}
    \caption{Orbits of \SO{2} and \T{1}$_y$ containing input points in $\mathbb{R}^2$. Unlike \T{2} and \SE{2}, not all points are not contained in a single orbit of these small groups.} 
    \label{fig:orbit}
\end{figure}
In this paper, we consider groups both large and small,
and we require the ability to enable or disable equivariances like translations.
To achieve this functionality, we need to go beyond the usual setting of homogeneous spaces considered in the literature, where every pair of elements in $\mX$ are related by an element in $G$.
Instead, we consider the quotient space $Q = \mX/G$, consisting of the distinct orbits of $G$ in $\mX$ (visualized in Figure \ref{fig:orbit}).\footnote{When $\mX$ is a homogeneous space and the quantity of interest is the quotient with the stabilizer of the origin $H$: $G/H \simeq \mX$, which has been examined extensively in the literature. Here we concerned with the separate quotient space $Q = \mX/G$, relevant when $\mX$ is not a homogeneous space.}
Each of these orbits $q \in Q$ is a homogeneous space of the group, and when $\mX$ is a homogeneous space of $G$ then there is only a single orbit. But in general, there will be many distinct orbits, and lifting should preserve the information on which orbit each point is on.

Since the most general equivariant mappings will use this orbit information, throughout the network the space of elements should not be $G$ but rather $G \times \mX/G$, and $x \in \mX$ is lifted to the tuples $(u,q)$ for $u \in G$ and $q \in Q$. This mapping may be one-to-one or one-to-many depending on the size of $H$, but will preserve the information in $x$ as $u o_q = x$ where $o_q$ is the chosen origin for each orbit. General equivariant linear transforms can depend on both the input and output orbit, and equivariance only constrains the dependence on group elements and not the orbits.

When the space of orbits $Q$ is continuous we can write the equivariant integral transform as
\begin{equation}\label{eq:transform}
    h(u,q)= \int_{G,Q} k(v^{-1}u,q,q') f(v,q') d\mu(v)dq'.
\end{equation}
When $G$ is the trivial group $\{\textrm{id}\}$, this equation simplifies to the integral transform $h(x)= \int k(x,x') f(x') dx'$ where each element in $\mX$ is in its own orbit. 

In general, even if $\mX$ is a smooth manifold and $G$ is a Lie group it is not guaranteed that $\mX/G$ is a manifold \citep{kono1987squaring}. However in practice this is not an issue as we will only have a finite number of orbits present in the data. All we need is an invertible way of embedding the orbit information into a vector space to be fed into $k_\theta$. One option is to use an embedding of the orbit origin $o_q$, or simply find enough invariants of the group to identify the orbit. To give a few examples:
\begin{enumerate}
    \item $\mX = \mathbb{R}^d$ and $G = \textrm{SO}(d): \textrm{Embed}(q(x)) = \|x\|$
    \item $\mX = \mathbb{R}^d$ and $G = \mathbb{R}^*: \textrm{Embed}(q(x)) = \cfrac{x}{\|x\|}$
    \item $\mX= \mathbb{R}^d$ and $G = \textrm{T}(k): \textrm{Embed}(q(x)) = x_{[k+1:d]}$
\end{enumerate}

Discretizing \eqref{eq:transform} as we did in \eqref{eq:group_conv_discretization}, we get
\begin{equation}
    h_i = \frac{1}{n_i}\sum_{j \in \mathrm{nbhd(i)}} \tilde{k}_\theta(\log(v_j^{-1}u_i),q_i,q_j) f_j,
\end{equation}
which again can be accelerated with the Efficient-PointConv trick by feeding in $a_{ij} = \mathrm{Concat}([\log(v_j^{-1}u_i),q_i,q_j])$ as input to the MLP.
If we want the filter to be local over orbits also, we can extend the distance $d((u_i,q_i),(v_j,q_j))^2 = d(u_i,v_j)^2 + \alpha \|q_i-q_j\|^2$, which need not be invariant to transformations on $q$.
To the best of our knowledge, we are the first to systematically address equivariances of this kind, where $\mX$ is not a homogeneous space of $G$. 

To recap, Algorithms \ref{alg:lifting} and \ref{alg:lie_conv} give a concise overview of our lifting procedure and our new convolution layer respectively. Please consult Appendix \ref{app:practical_details} for additional implementation details. 

\begin{algorithm}[H]
\SetAlgoLined
\textbf{Inputs:} spatial data $\{(x_i, f_i)\}_{i=1}^N$ ($x_i \in \mX$, $f_i \in \mathbb{R}^{\mathrm{c}_{in}}$).\\
\textbf{Returns:} matrix-orbit-value tuples $\{(u_j,q_j, f_j)\}_{j=1}^{NK}$. \\
For each orbit $q \in \mX/G$, choose an origin $o_q$. \\
For each $o_q$, compute its stabilizer $H_q$. \\
\For{$i = 1, \dots, N$}{
    Find the orbit $q_i\in \mX/G$ , s.t. $x_i \in q_i$. \\
    Sample $\{v_j\}_{j=1}^K$, where $v_j \sim \mu(H_{q_i})$ (see \ref{subsec:haar}).\\
    Compute an element $u_{i} \in G$ s.t. $u_i o_q = x_i$. \\
    $Z_i = \{(u_i v_j,q_i, f_i)\}_{j=1}^K$.
    }
\Return $Z$
\caption{Lifting from $\mX$ to $G\times \mX/G$}
\label{alg:lifting}
\end{algorithm}

\begin{algorithm}[H]
\SetAlgoLined
\textbf{Inputs:} matrix-orbit-value tuples $\{(u_j,q_j, f_j)\}_{j=1}^{m}$ \\
\textbf{Returns:} convolved matrix-orbit-values $\{ (u_i,q_i, h_i) \}_{i=1}^{m}$ \\
\For{$i = 1, \dots, m$}{
$u_i^{-1} = \exp(-\log(u_i))$.\\
$\mathrm{nbhd}_{i} = \{j: d((u_i,q_i),(u_j,q_j)) < r)\}$.\\
\For{$j = 1, \dots, m$}{
    $a_{ij} = \mathrm{Concat}([\log(u_i^{-1}u_j),q_i,q_j])$.
    }
$h_i = (1/n_i)\sum_{j \in \mathrm{nbhd}_{i}} k_\theta(a_{ij}) f_j$ (see \ref{sec:generalized_pointconv}).
}
\Return $(\mathbf{u}, \mathbf{q}, \mathbf h)$
\caption{The Lie Group Convolution Layer}
\label{alg:lie_conv}
\end{algorithm}

\begin{table*}[h]
    \centering
    \caption{Classification Error (\%) on RotMNIST dataset for LieConv with different group equivariances and baselines:
    G-CNN \citep{cohen2016group},
    H-Net \citep{worrall2017harmonic},
    ORN \citep{zhou2017oriented},
    TI-Pooling \citep{laptev2016ti},
    RotEqNet \citep{marcos2017rotation},
    E(2)-Steerable CNNs \citep{weiler2019general}
    .}
    \vspace{0.25cm}
    
    \small
    \begin{tabular}{cccccccccc ccccc}
    \toprule
    \multicolumn{6}{c}{Baseline Methods}&
    \multicolumn{6}{c}{LieConv (Ours)}\\
    \cmidrule(r){1-6} \cmidrule(r){7-12}
     G-CNN & H-NET & ORN & TI-Pooling &  RotEqNet & E(2)-Steerable &
     Trivial & \T{1}$_y$ & \T{2} & \SO{2} & \SO{2}$\times \mathbb{R}^*$ & \SE{2} \\
     \midrule
     2.28 & 1.69 & 1.54 & 1.2 & 1.09 & \textbf{0.68} & 1.58 & 1.49 & 1.44 & 1.42 &  1.27 & \textbf{1.24} \\
    \bottomrule
    \end{tabular}
    
    \label{table:rotmnist}
\end{table*}

\begin{table*}[ht]
    \centering
    \caption{QM9 Molecular Property Mean Absolute Error}
    \begin{tabular}{lr r r r r r r r r r r r}
    \toprule
     Task & $\alpha$ & $\Delta\varepsilon$  &$\varepsilon_{\textrm{HOMO}}$& $\varepsilon_{\textrm{LUMO}}$& $\mu$&$C_\nu$&$G$&$H$&$R^2$&$U$&$U_0$& ZPVE\\
     
     Units & bohr$^3$&meV &meV  &meV &D &cal/mol K &meV&meV&bohr$^2$&meV&meV&meV\\
    \midrule
    NMP& .092&69&43&38&\textbf{.030}&.040&19&17&.180&20&20&\textbf{1.500}\\
    SchNet&.235 &63&41&34&.033&.033&\textbf{14}&\textbf{14}&\textbf{.073}&\textbf{19}&\textbf{14}&1.700\\
    Cormorant&.085&61&34&38&.038&\textbf{.026}&20&21&.961&21&22&2.027\\
    LieConv(T3)&\textbf{.084}&\textbf{49}&\textbf{30} &\textbf{25} &.032&.038&22&24 &.800 &\textbf{19}&19 &2.280 \\
    \bottomrule
    \end{tabular}
    
    \label{table:qm9}
\end{table*}

\section{Applications to Image and Molecular Data}

First, we evaluate LieConv on two types of problems: classification on image data
and regression on molecular data. With LieConv as the convolution layers, we implement a bottleneck ResNet architecture with a final global pooling layer (Figure \ref{fig:architecture}). For a detailed architecture description, see Appendix \ref{sec:architecture}. We use the same model architecture for all tasks and achieve performance competitive with task-specific specialized methods.

\subsection{Image Equivariance Benchmark}
\label{sec:exp_img}

The RotMNIST dataset consists of 12k randomly rotated MNIST digits with rotations sampled uniformly from $\mathrm{SO}(2)$, separated into 10k for training and 2k for validation. 
This commonly used dataset has been a standard benchmark for equivariant CNNs on image data.
To apply LieConv to image data we interpret each input image as a collection of $N=28\times28$ points on $\mX = \mathbb{R}^2$ with associated binary values: $\{x_i,f(x_i)\}_{i=1}^{784}$ to which we apply a circular center crop. We note that LieConv is broadly targeting generic spatial data, and more practical equivariant methods exist specialized to images (e.g. \citet{weiler2019general}). However, as we demonstrate in Table \ref{table:rotmnist}, we are able to easily incorporate equivariance to a variety of different groups without changes to the method or the architecture of the network, while achieving performance competitive with methods that are not applicable beyond image data.

\begin{figure}[t]
    \centering
	\includegraphics[width=0.42\textwidth]{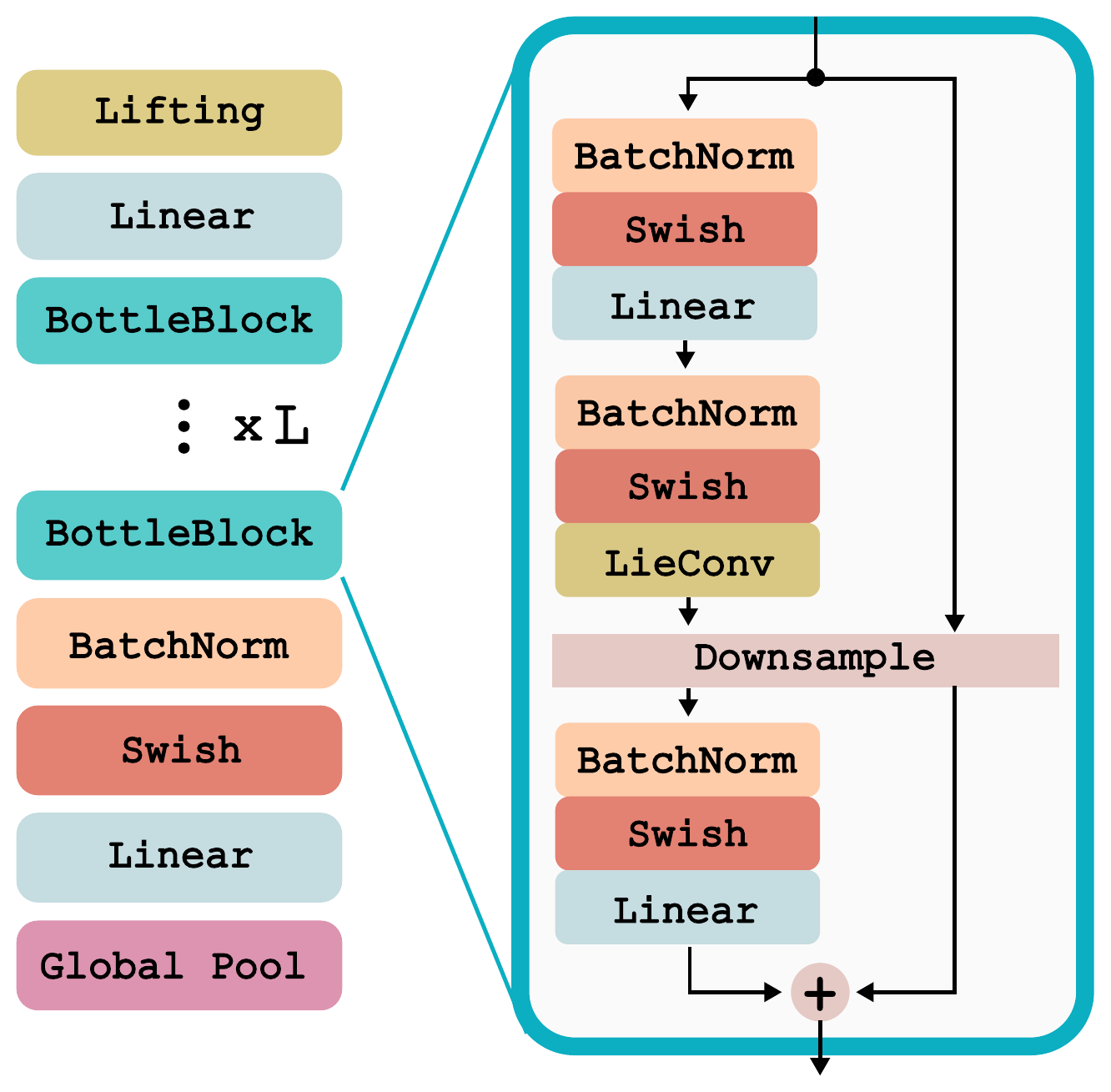}
	\caption{A visual overview of the LieConv model architecture, which is composed of $L$ LieConv bottleneck blocks that couple the values at different group elements together. The BottleBlock is a residual block with a LieConv layer between two linear layers.
	}\vspace*{-3mm}
    \label{fig:architecture}
\end{figure}

\subsection{Molecular Data}
\label{sec:exp_molec}
Now we apply LieConv to the QM9 molecular property learning task \citep{wu2018moleculenet}. 
The QM9 regression dataset consists of small inorganic molecules encoded as a collection of 3D spatial coordinates for each of the atoms, and their atomic charges. The labels consist of various properties of the molecules such as heat capacity. This is a challenging task as there is no canonical origin or orientation for each molecule, and the target distribution is invariant to $\mathrm{E}(3)$ (translation, rotation, and reflection) transformations of the coordinates. Successful models must generalize across different spatial locations and orientations.

We first perform an ablation study on the \texttt{Homo} problem of predicting the energy of the highest occupied molecular orbital for the molecules. We apply LieConv with different equivariance groups, combined with \SO{3} data augmentation. The results are reported in Table \ref{table:molecular_ablation}. Of the three groups, our \SE{3} network performs the best.
We then apply \T{3}-equivariant LieConv layers to the full range of tasks in the QM9 dataset and report the results in Table \ref{table:qm9}. We perform competitively with state-of-the-art methods \citep{gilmer2017neural,schutt2018schnet,anderson2019cormorant}, with lowest MAE on several of the tasks. See \ref{table:transformed_test} for a demonstration of the equivariance property and efficiency with limited data.

\begin{table}[t]
    \centering
    \label{table:molecular_ablation}
    \begin{tabular}{c cccc}
    \toprule
     Cormorant  & \triv & \SO{3} & \T{3} & \SE{3}\\
     \midrule
     34 & 31.7 & 65.4 & 29.6 &  \textbf{26.8}\\
     \bottomrule
    \end{tabular}
    \caption{LieConv performance (Mean Absolute Error in meV) for different groups on the \texttt{HOMO} regression problem.
    }
    
\end{table}
\hspace{-15mm}
\begin{figure}[ht]
    \centering
	\includegraphics[width=0.23\textwidth]{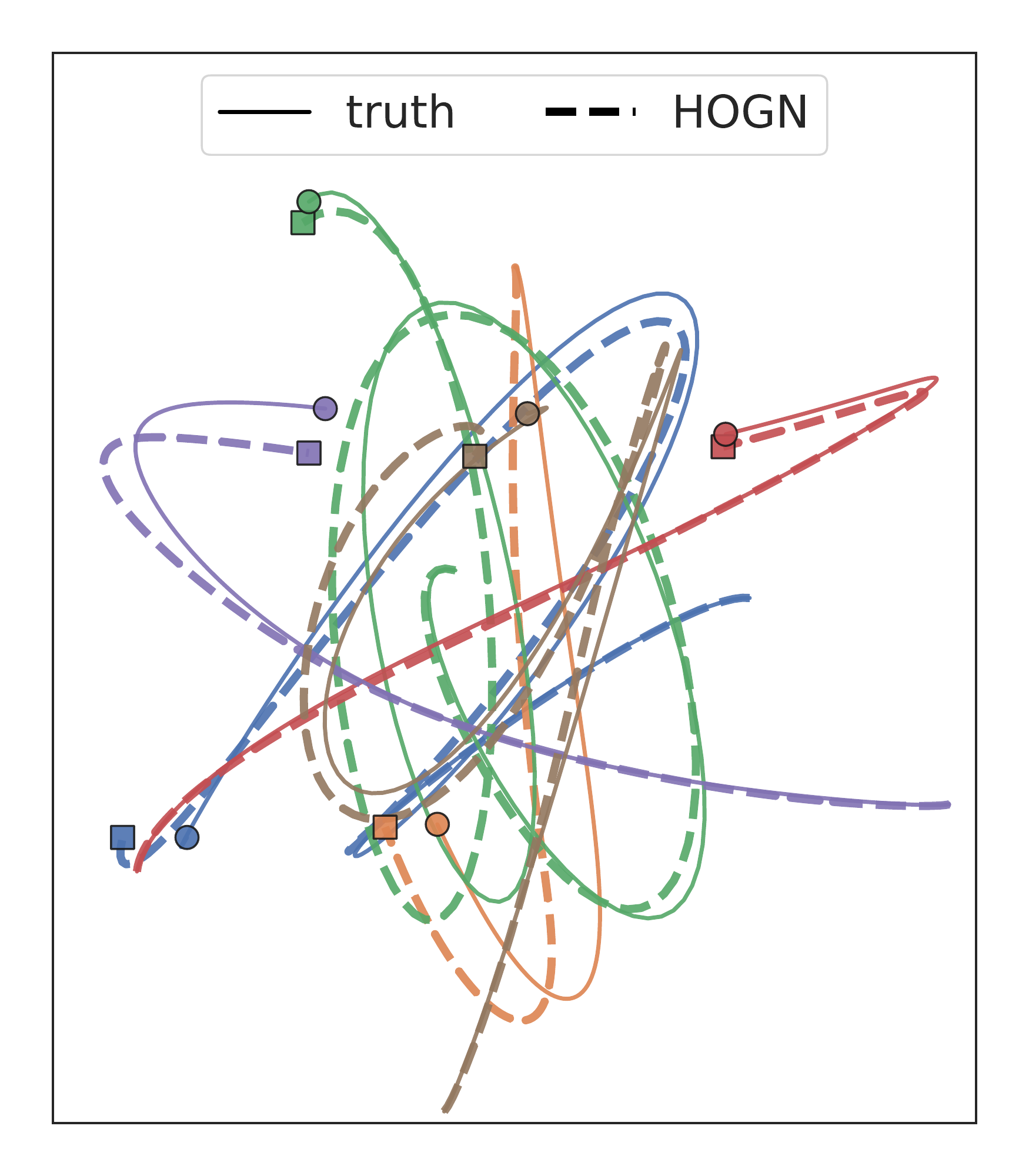}
	\hspace{-3mm}
	\includegraphics[width=0.23\textwidth]{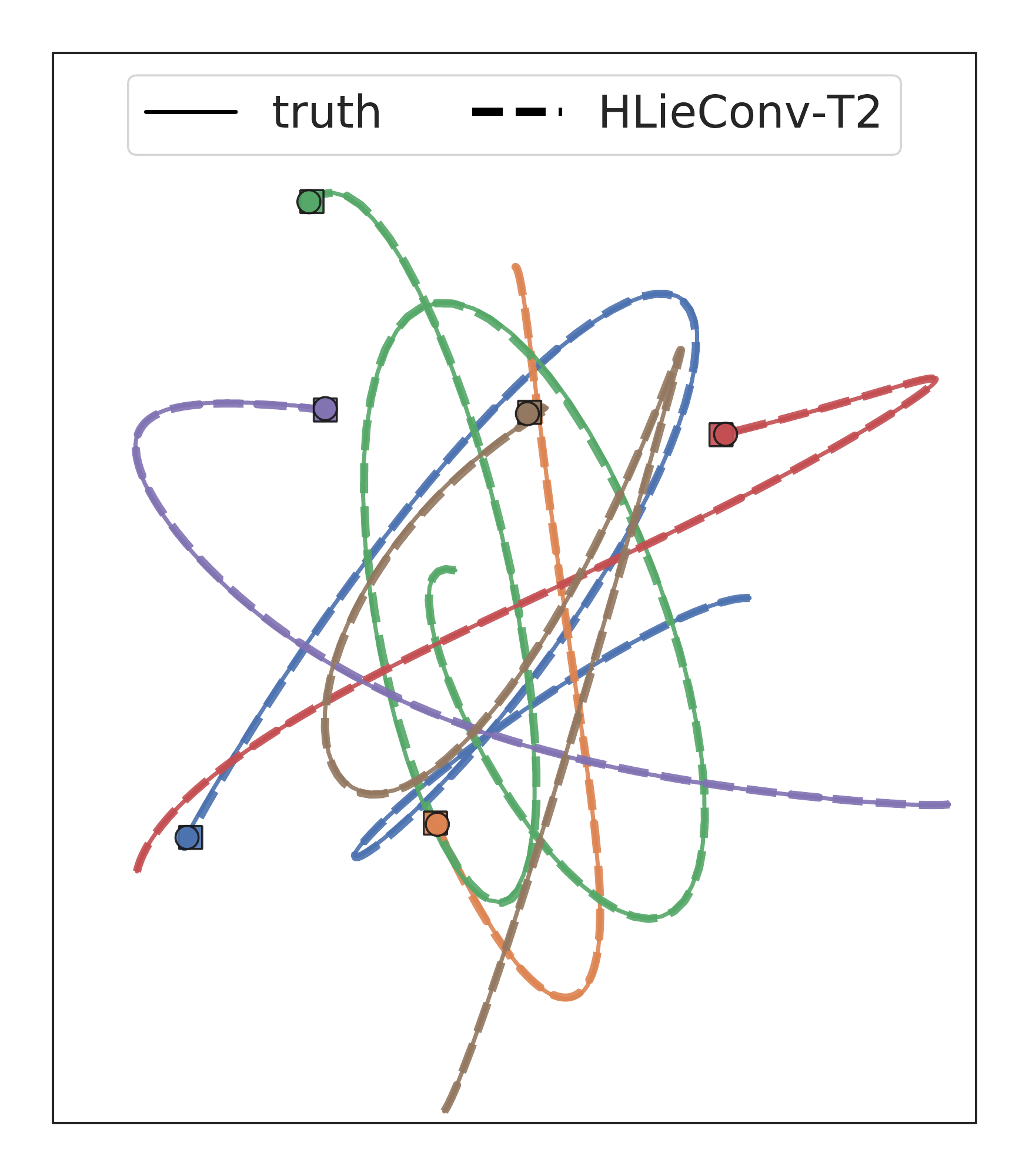}
	\label{fig:trajectories}
    \caption{A qualitative example of the trajectory predictions over 100 time steps on the 2D spring problem given a set of initial conditions.     We see that HLieConv (right) yields predictions that are accurate over a longer time than HOGN (left), a SOTA architecture for modeling interacting physical systems. 
    }
    \label{fig:spring_trajectories}
\end{figure}

\section{Modeling Dynamical Systems}
Accurate transition models for macroscopic physical systems are critical components in control systems \citep{lenz2015deepmpc, kamthe2017data, chua2018deep} and data-efficient reinforcement learning algorithms \citep{nagabandi2018neural, janner2019trust}. In this section we show how to enforce conservation of quantities such as linear and angular momentum in the modeling of Hamiltonian systems through LieConv symmetries.
 
\subsection{Predicting Trajectories with Hamiltonian Mechanics}\label{subsec:predicting_trajectories}
For dynamical systems, the equations of motion can be written in terms of the state $\state$ and time $t$:
$\dot{\state} = F(\state,t)$. Many physically occurring systems have Hamiltonian structure, meaning that the state can be split into generalized coordinates and momenta $\state = (\position,\momentum)$, and the dynamics can be written as
\begin{equation} \label{eq:hamiltons_equations}
\frac{d\position}{dt} = \frac{\partial \mathcal{H}}{\partial \momentum} \hspace{8mm}
\frac{d\momentum}{dt} = - \frac{\partial \mathcal{H}}{\partial \position}
\end{equation}
for some choice of scalar Hamiltonian $\mathcal{H}(\position, \momentum,t)$. $\mathcal{H}$ is often the total energy of the system, and can sometimes be split into kinetic and potential energy terms 
$\mathcal{H}(\position,\momentum) = K(\momentum) + V(\position)$. 
The dynamics can also be written compactly as $\dot{\state} = J \nabla_{\state} \mathcal{H}$ for 
$J = \begin{bmatrix}0 & I \\-I & 0 \\\end{bmatrix}$.

As shown in \citet{greydanus2019hamiltonian}, a neural network parametrizing $\hat{\mathcal{H}}_\theta(\state, t)$ can be learned directly from trajectory data, providing substantial benefits in generalization over directly modeling $F_\theta(\state, t)$, and with better energy conservation. We follow the approach of \citet{sanchez2019hamiltonian} and \citet{zhong2019symplectic}. Given an initial condition $\state_0$ and $F_\theta(\state,t) = J\nabla_{\state}\hat{\mathcal{H}}_\theta$, we employ a twice-differentiable model architecture and a differentiable ODE solver \citep{chen2018neural} to compute predicted states $(\hat\state_1, \dots, \hat\state_T) = \mathrm{ODESolve}(\state_0,F_\theta,(t_1,t_2,...,t_T))$. The parameters of the Hamiltonian model $\hat{\mathcal{H}}_\theta$ can be trained directly through the $L2$ loss,
\begin{equation}\label{eq:trajectory_loss}
    L(\theta) = \frac{1}{T} \sum\limits_{t=1}^T ||\hat\state_t -\state_t||_2^2.
\end{equation}

\begin{figure*}[h]
        \subfigure[]{
		\includegraphics[width=0.33\textwidth]{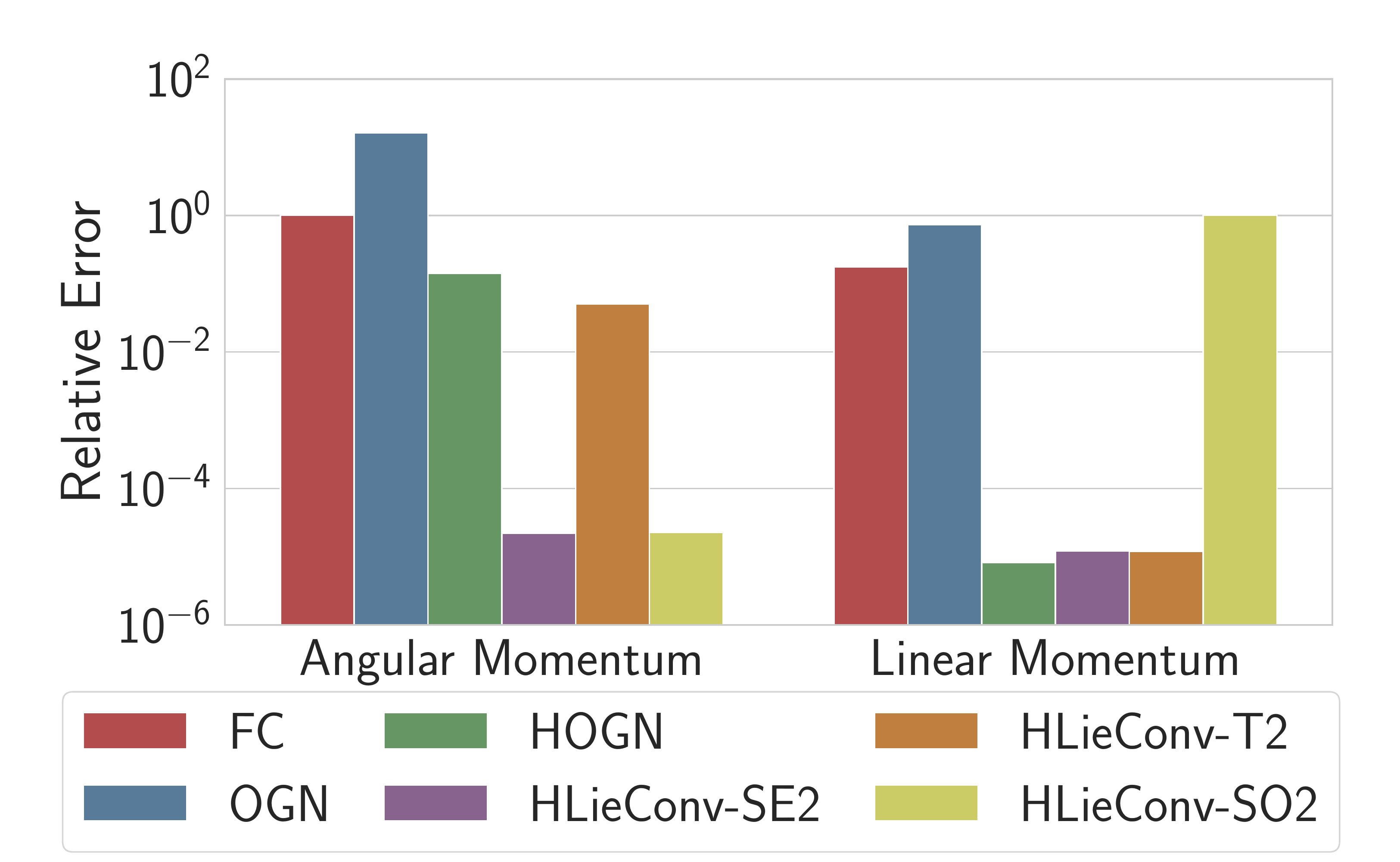}%
		\label{subfig:spring_conservation}
    }
        \subfigure[]{
		\includegraphics[width=0.33\textwidth]{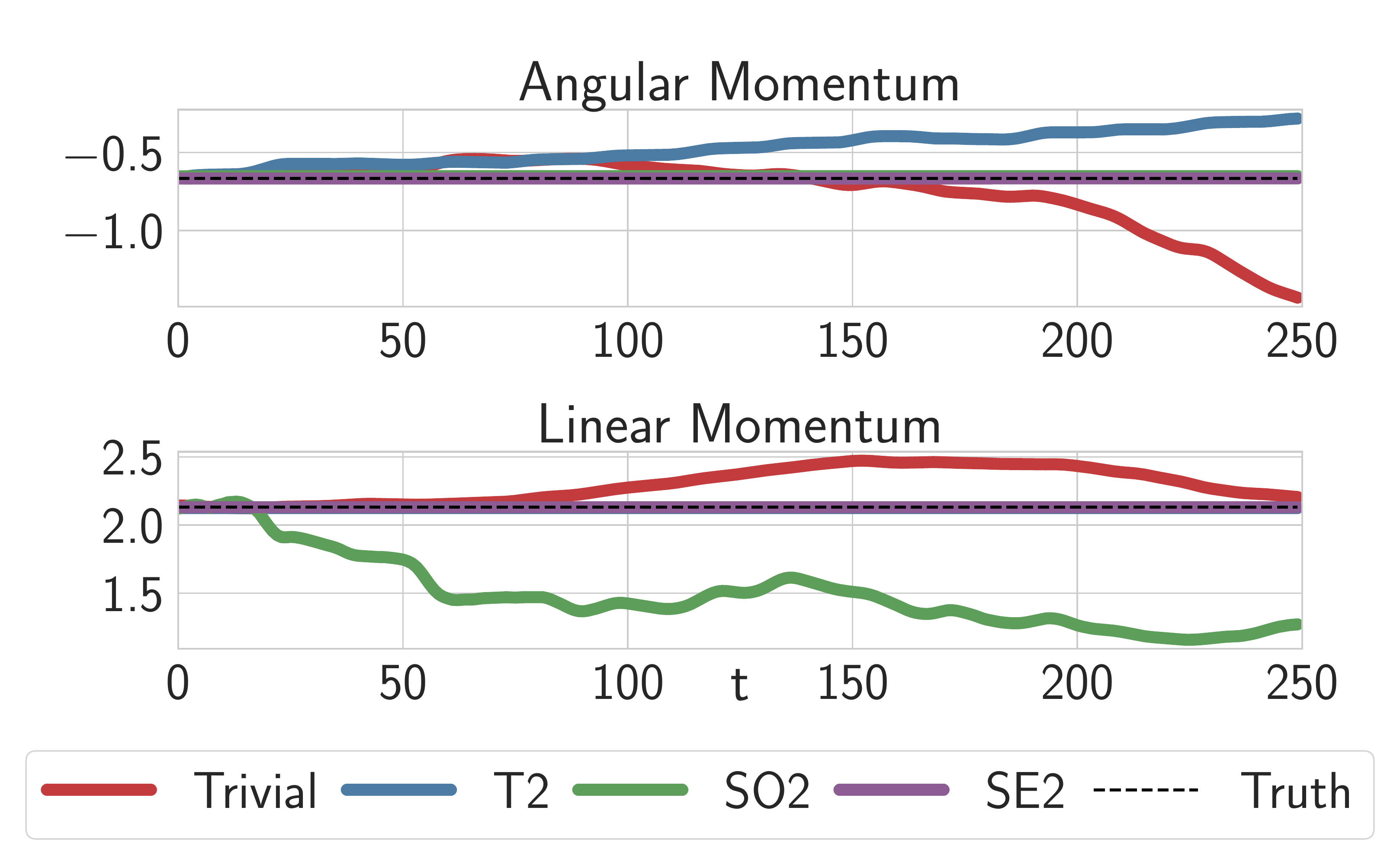}
    }
        \subfigure[]{
		\includegraphics[width=0.33\textwidth]{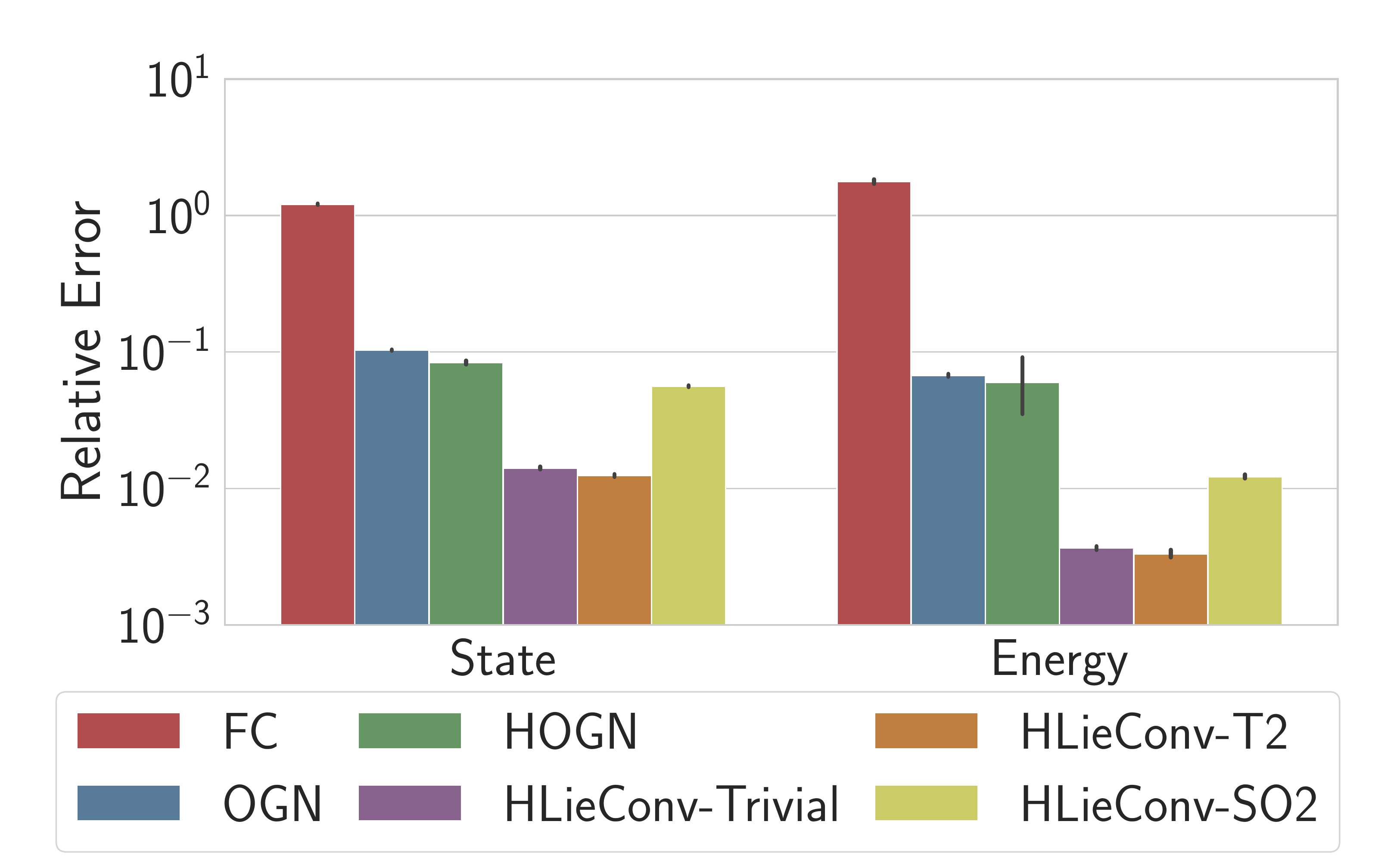}
		\label{subfig:state_energy}
    }
	\caption{\textbf{Left:} 
	We can directly control whether linear and angular momentum is conserved by changing the model's symmetries. The components of linear and angular momentum of the integrated trajectories are conserved up to integrator tolerance. \textbf{Middle:} Momentum along the rollout trajectories for the LieConv models with different imposed symmetries. 
	\textbf{Right:} Our method outperforms HOGN, a state-of-the-art model, on both state rollout error and system energy conservation.}
    \label{fig:dynamics}
\end{figure*}

\subsection{Exact Conservation of Momentum}
While equivariance is broadly useful as an inductive bias, it has a very special implication for the modeling of Hamiltonian systems. Noether's Hamiltonian theorem states that each continuous symmetry in the Hamiltonian of a dynamical system has a corresponding conserved quantity \citep{noether1971invariant, butterfield2006symmetry}. 
Symmetry with respect to the continuous transformations of translations and rotations lead directly to conservation of the total linear and angular momentum of the system, an extremely valuable property for modeling dynamical systems. In fact, all models that exactly conserve linear and angular momentum must have a corresponding translational and rotational symmetry. See Appendix \ref{subsec:noethers_theorem} for a primer on Hamiltonian symmetries, Noether's theorem, and the implications in the current setting.

As showed in Section \ref{sec:lie_group_convolutions}, we can construct models that are equivariant to a large variety of \emph{continuous} Lie Group symmetries, and therefore we can exactly conserve associated quantities like linear and angular momentum. Figure \ref{subfig:spring_conservation} shows that using LieConv layers with a given \T{2} and/or \SO{2} symmetry, the model trajectories conserve linear and/or angular momentum with relative error close to machine epsilon, determined by the integrator tolerance. As there is no corresponding Noether conservation for discrete symmetry groups, discrete approaches to enforcing symmetry \citep{cohen2016group,marcos2017rotation} would not be nearly as effective. 

\subsection{Results}

For evaluation, we compare a fully-connected (FC) Neural-ODE model \citep{chen2018neural}, ODE graph networks (OGN) \citep{battaglia2016interaction}, Hamiltonian ODE graph networks (HOGN) \citep{sanchez2019hamiltonian},
and our own LieConv architecture on predicting the motion of point particles connected by springs as described in \citep{sanchez2019hamiltonian}.
Figure \ref{fig:spring_trajectories} shows example rollout trajectories, and our quantitative results are presented in Figure \ref{fig:dynamics}. 
In the spring problem $N$ bodies with mass $m_1, \dots, m_N$ interact through pairwise spring forces with constants 
$k_1, \dots, k_{N \times N}$. The system preserves energy, linear momentum, and angular momentum. 
The behavior of the system depends both the values of the system parameters 
($\mathbf s = (k, m)$) and the initial conditions $\state_0$. 
The dynamics model must learn not only to predict trajectories across a broad range of initial conditions, 
but also infer the dependence on varied system parameters, which are additional inputs to the model. 
We compare models that attempt to learn the dynamics $F_\theta(\state,t) = d \mathbf z / dt$ directly 
against models that learn the Hamiltonian as described in section \ref{subsec:predicting_trajectories}. 

\begin{figure}[h]
    \centering
    \includegraphics[width=0.43\textwidth]{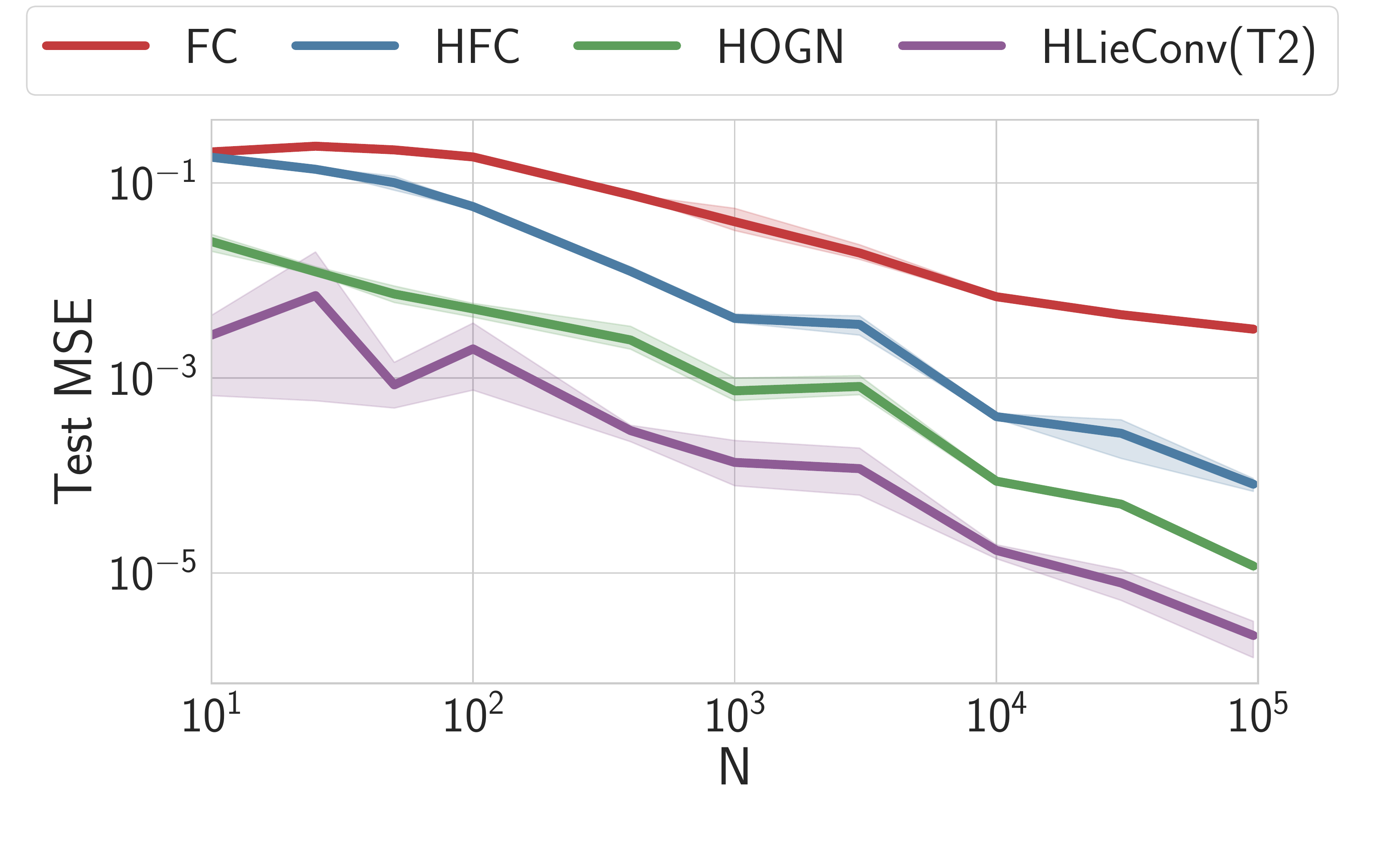}
    \vspace*{-3.5mm}
    \caption{Test MSE as a function of the number of examples in the training dataset, $N$. As the inductive biases of Hamiltonian, Graph-Network, and LieConv equivariance are added, generalization performance improves. LieConv outperforms the other methods across all dataset sizes.
    The shaded region corresponds to a 95\% confidence interval, estimated across 3 trials.}
    \label{fig:dynamics_scaling}
\end{figure}

In Figure \ref{fig:dynamics}(a) and \ref{fig:dynamics}(b) we show that by changing the invariance of our Hamiltonian models, we have direct control over the conservation of linear and angular momentum in the predicted trajectories. Figure \ref{fig:dynamics}(c) demonstrates that our method outperforms HOGN, a SOTA architecture for dynamics problems, and achieves significant improvement over the na\"ive fully-connected (FC) model. We summarize the various models and their symmetries in Table \ref{table:model_variants}. Finally, in Figure \ref{fig:dynamics_scaling} we evaluate test MSE of the different models over a range of training dataset sizes, highlighting the additive improvements in generalization from the Hamiltonian, Graph-Network, and equivariance inductive biases successively.

\section{Discussion}

We presented a convolutional layer to build networks that can handle a wide variety of data types, and flexibly swap out the equivariance of the model. While the image, molecular, and dynamics experiments demonstrate the generality of our method, there are many exciting application domains (e.g. time-series, geostats, audio, mesh) and directions for future work. 
We also believe that it will be possible to benefit from the inductive biases of HLieConv models even for systems that do not exactly preserve energy or momentum, such as those found in control systems and reinforcement learning.

The success of convolutional neural networks on images has highlighted the power of encoding symmetries in models for learning from raw sensory data. But the variety and complexity of other modalities of data is a significant challenge in further developing this approach. More general data may not be on a grid, it may possess other kinds of symmetries, or it may contain quantities that cannot be easily combined. We believe that central to solving this problem is a decoupling of convenient computational representations of data as dense arrays from the set of geometrically sensible operations they may have. We hope to move towards models that can `see' molecules, dynamical systems, multi-scale objects, heterogeneous measurements, and higher mathematical objects, in the way that convolutional neural networks perceive images.

\subsection*{Acknowledgements}
MF, SS, PI and AGW are supported by an Amazon Research  Award,  Amazon Machine Learning Research Award, 
Facebook  Research, NSF I-DISRE 193471, NIH R01 DA048764-01A1, NSF IIS-1910266, NSF 1922658 NRT-HDR: 
FUTURE Foundations, Translation, and Responsibility for Data Science, and by the United States Department of Defense through the National Defense Science \& Engineering Graduate (NDSEG) Fellowship Program. We thank Alex Wang for helpful comments.

\bibliographystyle{plainnat}
\bibliography{main}

\cleardoublepage

\appendix
\appendixpage
\section{Derivations and Additional Methodology}
\subsection{Generalized PointConv Trick}\label{sec:generalized_pointconv}
The matrix notation becomes very cumbersome for manipulating these higher order $n$-dimensional arrays, so we will instead use index notation with Latin indices $i,j,k$ indexing points, Greek indices $\alpha,\beta,\gamma$ indexing feature channels, and $c$ indexing the coordinate dimensions of which there are $d=3$ for PointConv and $d=\textrm{dim}(G)+ 2 \textrm{ dim}(Q)$ for LieConv.\footnote{$\textrm{dim}(Q)$ is the dimension of the space into which $Q$, the orbit identifiers, are embedded.} As the objects are not geometric tensors but simply $n$-dimensional arrays, we will make no distinction between upper and lower indices. After expanding into indices, it should be assumed that all values are scalars, and that any free indices can range over all of the values.

Let $k_{ij}^{\alpha,\beta}$ be the output of the MLP $k_\theta$ which takes $\{a_{ij}^c\}$ as input and acts independently over the locations $i,j$. For PointConv, the input $a_{ij}^c = x_i^c-x_j^c$ and for LieConv the input $a_{ij}^c = \mathrm{Concat}([\log(v_j^{-1}u_i),q_i,q_j])^c$.

We wish to compute
\begin{equation}\label{eq:trick}
    h_i^\alpha = \sum_{j,\beta} k^{\alpha,\beta}_{ij} f_j^\beta.
\end{equation}
In \citet{wu2019pointconv}, it was observed that since $k_{ij}^{\alpha,\beta}$ is the output of an MLP, $k_{ij}^{\alpha,\beta} = \sum_{\gamma} W^{\alpha,\beta}_{\gamma}s_{i,j}^{\gamma}$ for some final weight matrix $W$ and penultimate activations $s_{i,j}^{\gamma}$ ($s_{i,j}^{\gamma}$ is simply the result of the MLP after the last nonlinearity). With this in mind, we can rewrite \eqref{eq:trick}
\begin{align}
    h_i^\alpha &= \sum_{j,\beta} \big(\sum_\gamma W^{\alpha,\beta}_{\gamma}s_{i,j}^{\gamma}\big) f_j^\beta\\
    &=\sum_{\beta,\gamma} W^{\alpha,\beta}_{\gamma}\big(\sum_j s_{i,j}^{\gamma} f_j^\beta\big)
\end{align}
    
In practice, the intermediate number of channels is much less than the product of $c_{in}$ and $c_{out}$: $|\gamma|<|\alpha||\beta|$ and so this reordering of the computation leads to a massive reduction in both memory and compute. Furthermore, $b_i^{\gamma,\beta} = \sum_j s_{i,j}^{\gamma} f_j^\beta$ can be implemented with regular matrix multiplication and $h_i^\alpha = \sum_{\beta,\gamma} W^{\alpha,\beta}_{\gamma}b_i^{\gamma,\beta}$ can be also by flattening $(\beta,\gamma)$ into a single axis $\varepsilon$:  $h_i^\alpha = \sum_{\varepsilon}W^{\alpha,\varepsilon}b_i^\varepsilon$. 

The sum over index $j$ can be restricted to a subset $j(i)$ (such as a chosen neighborhood) by computing $f^\beta_{(\cdot)}$ at each of the required indices and padding to the size of the maximum subset with zeros, and computing $b_i^{\gamma,\beta} = \sum_j s_{i,j(i)}^\gamma f^\beta_{j(i)}$ using dense matrix multiplication. Masking out of the values at indices $i$ and $j$ is also necessary when there are different numbers of points per minibatch but batched together using zero padding. The generalized PointConv trick can thus be applied in batch mode when there may be varied number of points per example and varied number of points per neighborhood.

\subsection{Abelian $G$ and Coordinate Transforms}
\label{sec:abelian}
For Abelian groups that cover $\mX$ in a single orbit, the computation is very similar to ordinary Euclidean convolution. Defining $a_i = \log(u_i)$, $b_j = \log(v_j)$, and using the fact that $e^{-b_j}e^{a_i} = e^{a_i-b_j}$ means that $\log(v_j^{-1}u_i) = (\log\circ \exp)(a_i-b_j)$.
Defining $\tilde{f} = f \circ \exp$, $\tilde{h} = h \circ \exp$; we get
\begin{equation}\label{eq:proj_abelian}
    \tilde{h}(a_i) = \frac{1}{n}\sum_{j\in \mathrm{nbhd}(i)} (\tilde{k}_\theta \circ\mathrm{proj})(a_i-b_j) \tilde{f}(b_j),
\end{equation}
where $\mathrm{proj} = \log\circ \exp$ projects to the image of the logarithm map.
Apart from a projection and a change to logarithmic coordinates, this is equivalent to Euclidean convolution in a vector space with dimensionality of the group. When the group is Abelian and $\mX$ is a homogeneous space, then the dimension of the group is the dimension of the input. In these cases we have a trivial stabilizer group $H$ and single origin $o$, so we can view $f$ and $h$ as acting on the input $x_i = u_io$.

This directly generalizes some of the existing coordinate transform methods for achieving equivariance from the literature such as log polar coordinates for rotation and scaling equivariance \cite{esteves2017polar}, and using hyperbolic coordinates for squeeze and scaling equivariance.

\textbf{Log Polar Coordinates:}
Consider the Abelian Lie group of positive scalings and rotations: $G=\mathbb{R}^* \times$ \SO{2} acting on $\mathbb{R}^2$. Elements of the group $M\in G$ can be expressed as a $2\times 2$ matrix 
\[
    M(r,\theta) = \begin{bmatrix}r\cos(\theta) & -r\sin(\theta) \\r\sin(\theta) & r\cos(\theta) \\\end{bmatrix}
\]
for $r \in \mathbb{R}^+$ and $\theta \in \mathbb{R}$.
The matrix logarithm is\footnote{Here $\theta \textrm{ mod } 2\pi$ is defined to mean $\theta +2\pi n$ for the integer $n$ such that the value is in $(-\pi,\pi)$, consistent with the principal matrix logarithm. $(\theta+\pi)\%2\pi-\pi$ in programming notation.
}
\[
    \log \bigg(\begin{bmatrix}r\cos(\theta) & -r\sin(\theta) \\r\sin(\theta) & r\cos(\theta) \\\end{bmatrix}\bigg) = \begin{bmatrix}\log(r) & -\theta \textrm{ mod } 2\pi \\\theta\textrm{ mod }2\pi & \log(r) \\\end{bmatrix},
\]
or more compactly $\log(M(r,\theta)) = \log(r) I + (\theta \textrm{ mod } 2\pi) J$, which is $[\log(r),\theta \textrm{ mod } 2\pi]$ in the basis for the Lie algebra $[I,J]$. It is clear that $\mathrm{proj} = \log\circ\exp$ is simply $\textrm{ mod } 2\pi$ on the $J$ component.

As $\mathbb{R}^2$ is a homogeneous space of $G$, one can choose the global origin $o=[1,0] \in \mathbb{R}^2$. A little algebra shows that lifting to the group yields the transformation $u_i = M(r_i,\theta_i)$ for each point $p_i = u_io$, where $r = \sqrt{x^2+y^2}$, and $\theta = \mathrm{atan2}(y,x)$ are the polar coordinates of the point $p_i$. Observe that the logarithm of $v_j^{-1}u_i$ has a simple expression highlighting the fact that it is invariant to scale and rotational transformations of the elements, 
\begin{align*}
    \log(v_j^{-1}u_i) &= \log(M(r_j,\theta_j)^{-1}M(r_i,\theta_i))\\
    &= \log(r_i/r_j)I + (\theta_i-\theta_j\textrm{ mod } 2\pi)J.\\
\end{align*}

Now writing out our Monte Carlo estimation of the integral:
\begin{equation*}
  h(p_i) = \frac{1}{n}\sum_{j}
    \tilde{k}_\theta(\log(r_i/r_j),\theta_i-\theta_j\textrm{ mod } 2\pi) f(p_j),
\end{equation*}
which is a discretization of the log polar convolution from \citet{esteves2017polar}. This can be trivially extended to encompass cylindrical coordinates with the group $\T{1}\times\mathbb{R}^* \times$ \SO{2}.

\textbf{Hyperbolic coordinates:}
For another nontrivial example, consider the group of scalings and squeezes $G = \mathbb{R}^* \times \textrm{SQ}$ acting on the positive orthant $\mX = \{(x,y) \in \mathbb{R}^2: x>0,y>0\}$. Elements of the group can be expressed as the product of a squeeze mapping and a scaling
\begin{equation*}
    M(r,s) = \begin{bmatrix}s & 0 \\0 & 1/s \\\end{bmatrix}\begin{bmatrix}r & 0 \\0 & r \\\end{bmatrix}= \begin{bmatrix}rs & 0 \\0 & r/s \\\end{bmatrix}
\end{equation*}
for any $r,s \in \mathbb{R}^+$. As the group is abelian, the logarithm splits nicely in terms of the two generators $I$ and $A$:
\begin{equation*}
    \log \bigg(\begin{bmatrix}rs & 0 \\0 & r/s \\\end{bmatrix}\bigg) = (\log r)\begin{bmatrix}1 & 0 \\0 & 1 \\\end{bmatrix} + (\log s)\begin{bmatrix}1 & 0 \\0 & -1 \\\end{bmatrix}.
\end{equation*}
Again $\mX$ is a homogeneous space of $G$, and we choose a single origin $o = [1,1]$. With a little algebra, it is clear that $M(r_i,s_i)o = p_i$ where $r = \sqrt{xy}$ and $s = \sqrt{x/y}$ are the hyperbolic coordinates of $p_i$.

Expressed in the basis $\mathcal{B} = [I,A]$ for the Lie algebra above, we see that
\begin{equation*}
    \log (v_j^{-1}u_i) = \log (r_i/r_j) I + \log (s_i/s_j) A
\end{equation*}
yielding the expression for convolution
\begin{equation*}
    h(p_i) = \frac{1}{n}\sum_{j}
    \tilde{k}_\theta(\log(r_i/r_j),\log(s_i/s_j)) f(p_j),
\end{equation*}
  which is equivariant to squeezes and scalings.
  
As demonstrated, equivariance to groups that contain the input space in a single orbit and are abelian can be achieved with a simple coordinate transform; however our approach generalizes to groups that are both 'larger' and 'smaller' than the input space, including coordinate transform equivariance as a special case.

\subsection{Sufficient Conditions for Geodesic Distance}\label{subsec:suff_cond_geodesic}
In general, the function $d(u,v) = \|\log(v^{-1}u)\|_F$, defined on the domain of $\mathrm{GL}(d)$ covered by the exponential map, satisfies the first three conditions of a distance metric but not the triangle inequality, making it a semi-metric:
\begin{enumerate}
    \item $d(u,v) \ge 0$
    \item $d(u,v) = 0 \Leftrightarrow \log(u^{-1}v) = 0 \Leftrightarrow u = v$
    \item $d(u,v) = \|\log(v^{-1}u)\| = \|-\log(u^{-1}v)\| = d(v,u)$.
\end{enumerate}
However for certain subgroups of $\mathrm{GL}(d)$ with additional structure, the triangle inequality holds and the function is the distance along geodesics connecting group elements $u$ and $v$ according to the metric tensor
\begin{equation}\label{eq:metric}
    \langle A,B \rangle_u := \Tr(A^Tu^{-T}u^{-1}B),
\end{equation} 
where $-T$ denotes inverse and transpose.

Specifically, if the subgroup $G$ is in the image of the $\exp: \mathfrak{g} \to G$ map and each infinitesmal generator commutes with its transpose: $[A,A^T] = 0$ for $\forall A \in  \mathfrak{g}$, then $d(u,v) = \|\log(v^{-1}u)\|_F$ is the geodesic distance between $u,v$.

\textbf{Geodesic Equation:}
Geodesics of \eqref{eq:metric} satisfying $\nabla_{\dot{\gamma}} \dot{\gamma}=0$ can equivalently be derived by minimizing the energy functional
\[E[\gamma] =\int_\gamma \langle \dot{\gamma},\dot{\gamma} \rangle_\gamma dt = \int_0^1 \Tr(\dot{\gamma}^T\gamma^{-T}\gamma^{-1}\dot{\gamma})dt\]
using the calculus of variations. Minimizing curves $\gamma(t)$, connecting elements $u$ and $v$ in $G$ ($\gamma(0)=v, \gamma(1)=u$) satisfy
\[0=\delta E = \delta \int_0^1 \Tr(\dot{\gamma}^T\gamma^{-T}\gamma^{-1}\dot{\gamma})dt\]
Noting that $\delta (\gamma^{-1}) = -\gamma^{-1}\delta \gamma \gamma^{-1}$ and the linearity of the trace,
\[2 \int_{0}^1 \Tr(\dot{\gamma}^T\gamma^{-T}\gamma^{-1}\delta \dot{\gamma}) - \Tr(\dot{\gamma}^T\gamma^{-T}\gamma^{-1}\delta\gamma\gamma^{-1}\dot{\gamma}) dt=0.\]

Using the cyclic property of the trace and integrating by parts, we have that
\[-2\int_{0}^1 \Tr\bigg( \big(\frac{d}{dt}(\dot{\gamma}^T\gamma^{-T}\gamma^{-1}) + \gamma^{-1}\dot{\gamma}\dot{\gamma}^T\gamma^{-T}\gamma^{-1}\big)\delta\gamma\bigg) dt=0,\]
where the boundary term $\Tr(\dot{\gamma}\gamma^{-T}\gamma^{-1}\delta\gamma)\big|_0^1$ vanishes since $(\delta \gamma)(0)=(\delta\gamma)(1) = 0$.

As $\delta\gamma$ may be chosen to vary arbitrarily along the path, $\gamma$ must satisfy the geodesic equation: \begin{equation}\label{eq:geodesic}
    \frac{d}{dt}(\dot{\gamma}^T\gamma^{-T}\gamma^{-1}) + \gamma^{-1}\dot{\gamma}\dot{\gamma}^T\gamma^{-T}\gamma^{-1} = 0.
\end{equation}
\textbf{Solutions:}
When $A = \log(v^{-1}u)$ satisfies $[A,A^T]=0$, the curve $\gamma(t) = v\exp(t\log(v^{-1}u))$ is a solution to the geodesic equation \eqref{eq:geodesic}.
Clearly $\gamma$ connects $u$ and $v$, $\gamma(0)=v$ and $\gamma(1)=u$. Plugging in $\dot{\gamma} = \gamma A$ into the left hand side of equation \eqref{eq:geodesic}, we have
\begin{align*}
    &= \frac{d}{dt}(A^T\gamma^{-1}) + AA^T\gamma^{-1}\\
 &= -A^T\gamma^{-1}\dot{\gamma}\gamma^{-1} + AA^T\gamma^{-1}\\
 &= [A,A^T]\gamma^{-1} = 0
\end{align*}
\textbf{Length of $\gamma$:}
The length of the curve $\gamma$ connecting $u$ and $v$ is $\|\log(v^{-1}u)\|_F$, 
\begin{align*}
    L[\gamma] &=\int_\gamma \sqrt{\langle \dot{\gamma},\dot{\gamma} \rangle_\gamma} dt = \int_0^1 \sqrt{\Tr(\dot{\gamma}^T\gamma^{-T}\gamma^{-1}\dot{\gamma})}dt\\
    &= \int_0^1 \sqrt{\Tr(A^TA)}dt = \|A\|_F = \|\log(v^{-1}u)\|_F
\end{align*}

Of the Lie Groups that we consider in this paper, all of which have a single connected component, the groups 
$G = \T{d}, \SO{d}, \mathbb{R}^*\times \SO{d}, \mathbb{R}^* \times\mathrm{SQ}$ satisfy this property that $[\mathfrak{g},\mathfrak{g}^T] =0$; however, the \SE{d} groups do not.

\subsection{Equivariant Subsampling}
\label{sec:equivariant_subsampling}

Even if all distances and neighborhoods are precomputed, the cost of computing equation \eqref{eq:lieconv} for $i=1,...,N$ is still quadratic, $O(nN) = O(N^2)$, because the number of points in each neighborhood $n$ grows linearly with $N$ as $f$ is more densely evaluated. So that our method can scale to handle a large number of points, we show two ways two equivariantly subsample the group elements, which we can use both for the locations at which we evaluate the convolution and the locations that we use for the Monte Carlo estimator. Since the elements are spaced irregularly, we cannot readily use the coset pooling method described in \cite{cohen2016group}, instead we can perform:

\textbf{Random Selection:} Randomly selecting a subset of $p$ points from the original $n$ preserves the original sampling distribution, so it can be used.

\textbf{Farthest Point Sampling:} 
Given a set of group elements $S = \{u_i\}_{i=1}^k \in G$, we can select a subset $S_p^*$ of size $p$ by maximizes the minimum distance between any two elements in that subset,
\begin{equation}
    \textrm{Sub}_p(S) := S_p^* = \argmax_{S_p \subset S} \min_{u,v \in S_p: u\ne v} d(u,v),
\end{equation}
farthest point sampling on the group. 
Acting on a set of elements, $\textrm{Sub}_p: S \mapsto S_p^*$, the farthest point subsampling is equivariant $\textrm{Sub}_p(wS) = w\textrm{Sub}_p(S)$ for any $w\in G$. Meaning that applying a group element to each of the elements does not change the chosen indices in the subsampled set because the distances are left invariant $d(u_i,u_j) = d(wu_i,wu_j)$.

Now we can use either of these methods for $\textrm{Sub}_p(\cdot)$ to equivariantly subsample the quadrature points in each neighborhood used to estimate the integral to a fixed number $p$,
\begin{equation}
    h_i = \frac{1}{p}\sum_{j \in \textrm{Sub}_p(\textrm{nbhd}(u_i))} k_\theta(v_j^{-1}u_i) f_j.
    \label{eq:lieconv2}
\end{equation}
Doing so has reduced the cost of estimating the convolution from $O(N^2)$ to $O(pN)$, ignoring the cost of computing $\textrm{Sub}_p$ and $\{\textrm{nbhd}(u_i)\}_{i=1}^{N}$.

\subsection{Review and Implications of Noether's Theorem}\label{subsec:noethers_theorem}
In the Hamiltonian setting, Noether's theorem relates the continuous symmetries of the Hamiltonian of a system with conserved quantities, and has been deeply impactful in the understanding of classical physics. We give a review of Noether's theorem, loosely following \citet{butterfield2006symmetry}.

\textbf{More on Hamiltonian Dynamics}

As introduced earlier, the Hamiltonian is a function acting on the state $H(z) = H(q,p)$, (we will ignore time dependence for now) can be viewed more formally as a function on the cotangent bundle $(q,p) = z \in M=T^*C$ where $C$ is the coordinate configuration space, and this is the setting for Hamiltonian dynamics. 

In general, on a manifold $\mathcal{M}$, a vector field $X$ can be viewed as an assignment of a directional derivative along $\mathcal{M}$ for each point $z \in \mathcal{M}$. It can be expanded in a basis using coordinate charts $X = \sum_\alpha X^\alpha \partial_\alpha$, where $\partial_\alpha = \frac{\partial}{\partial z^\alpha}$ and acts on functions $f$ by $X(f) = \sum_\alpha X^\alpha \partial_\alpha f$. In the chart, each of the components $X^\alpha$ are functions of $z$.

In Hamiltonian mechanics, for two functions on $M$, there is the Poisson bracket which can be written in terms of the canonical coordinates $q_i,p_i$, \footnote{Here we take the definition of the Poisson bracket to be negative of the usual definition in order to streamline notation.}
\begin{equation*}
    \{f,g\} = \sum_i \frac{\partial f}{\partial p_i}  \frac{\partial g}{\partial q_i} - \frac{\partial f}{\partial q_i}  \frac{\partial g}{\partial p_i}.
\end{equation*}

The Poisson bracket can be used to associate each function $f$ to a vector field
\begin{equation*}
X_f = \{f, \cdot\} = \sum_i \frac{\partial f}{\partial p_i}  \frac{\partial }{\partial q_i} - \frac{\partial f}{\partial q_i}  \frac{\partial }{\partial p_i},
\end{equation*}
 which specifies, by its action on another function $g$, the directional derivative of $g$ along $X_f$: $X_f(g) = \{f,g\}$. Vector fields that can be written in this way are known as Hamiltonian vector fields, and the Hamiltonian dynamics of the system is a special example $X_H = \{H,\cdot\}$. This vector field in canonical coordinates $z=(p,q)$ is the vector field $X_H = F(z) = J\nabla_zH$ (i.e. the symplectic gradient, as discussed in Section \ref{subsec:predicting_trajectories}). Making this connection clear, a given scalar quantity evolves through time as $\dot{f} = \{H,f\}$. But this bracket can be used to evaluate the rate of change of a scalar quantity along the flows of vector fields other than the dynamics, such as the flows of continuous symmetries.

\textbf{Noether's Theorem}

The flow $\phi_\lambda^X$ by $\lambda \in \mathbb{R}$ of a vector field $X$ is the set of integral curves, the unique solution to the system of ODEs $\dot{z}^\alpha = X^\alpha$ with initial condition $z$ and at parameter value $\lambda$, or more abstractly the iterated application of $X$: $\phi_\lambda^X = \exp(\lambda X)$.
Continuous symmetries transformation are the transformations that can be written as the flow $\phi_\lambda^X$ of a vector field. The directional derivative characterizes how a function such as the Hamiltonian changes along the flow of $X$ and is a special case of the Lie Derivative $\mathcal{L}$.

\begin{equation*}
    \mathcal{L}_{X}H = \frac{d}{d\lambda} (H\circ\phi^{X}_\lambda)\big|_{\lambda=0} = X(H)
\end{equation*}
A scalar function is invariant to the flow of a vector field if and only if the Lie Derivative is zero 
\begin{equation*}
    H(\phi^{X}_\lambda(z)) = H(z) \Leftrightarrow \mathcal{L}_{X}H = 0.
\end{equation*}
For all transformations that respect the Poisson Bracket\footnote{More precisely, the Poisson Bracket can be formulated in a coordinate free manner in terms of a symplectic two form $\omega$, $\{f,g\} = \omega(X_f,X_g)$. In the original coordinates $\omega = \sum_i dp_i \wedge dq^i$, and this coordinate basis, $\omega$ is represented by the matrix $J$ from earlier. The dynamics $X_H$ are determined by d$H=\omega(X_H,\cdot) = \iota_{X_H}\omega$. Transformations which respect the Poisson Bracket are symplectic, $\mathcal{L}_X\omega = 0$. With Cartan's magic formula, this implies that d$(\iota_X\omega) =0$. Because the form $\iota_X\omega$ is closed, Poincare's Lemma implies that locally $(\iota_X\omega)=\mathrm{d}f)$ for some function $f$ and hence $X=X_f$ is (locally) a Hamiltonian vector field. For more details see \citet{butterfield2006symmetry}.}, which we add as a requirement for a symmetry, the vector field $X$ is (locally) Hamiltonian and there exists a function $f$ such that $X=X_f = \{f,\cdot\}$. If $M$ is a contractible domain such as $\mathbb{R}^{2n}$, then $f$ is globally defined. 
For every continuous symmetry $\phi^{X_f}_\lambda$,

\begin{equation*}
    \mathcal{L}_{X_f}H = X_f(H) = \{f,H\} = -\{H,f\} = -X_H (f),
\end{equation*}
by the antisymmetry of the Poisson bracket. So if $\phi^{X}_\lambda$ is a symmetry of $H$, then $X=X_f$ for some function $f$, and $H(\phi^{X_f}_\lambda(z)) = H(z)$ implies
\begin{equation*}
\mathcal{L}_{X_f}H = 0\Leftrightarrow \mathcal{L}_{X_H}f=0 \Leftrightarrow  f(\phi^{X_H}_\tau(z))=f(z)
\end{equation*}
or in other words $f(z(t+\tau)) = f(z(t))$ and $f$ is a conserved quantity of the dynamics.

This implication goes both ways, if $f$ is conserved then $\phi^{X_f}_\lambda$ is necessarily a symmetry of the Hamiltonian, and if $\phi^{X_f}_\lambda$ is a symmetry of the Hamiltonian then $f$ is conserved.

\textbf{Hamiltonian vs Dynamical Symmetries}

So far we have been discussing Hamiltonian symmetries, invariances of the Hamiltonian. But in the study of dynamical systems there is a related concept of dynamical symmetries, symmetries of the equations of motion. This notion is also captured by the Lie Derivative, but between vector fields.
A dynamical system $\dot{z} = F(z)$, has a continuous dynamical symmetry $\phi^X_\lambda$ if the flow along the dynamical system commutes with the symmetry:
\begin{equation}\label{eq:flow_commutes}
    \phi^X_\lambda (\phi^F_t (z)) = \phi^F_t (\phi^X_\lambda (z)).
\end{equation}
Meaning that applying the symmetry transformation to the state and then flowing along the dynamical system is equivalent to flowing first and then applying the symmetry transformation.
Equation \eqref{eq:flow_commutes} is satisfied if and only if the Lie Derivative is zero:
\begin{equation*}
    \mathcal{L}_X F = [X,F] = 0,
\end{equation*}
where $[\cdot,\cdot]$ is the Lie bracket on vector fields.\footnote{The Lie bracket on vector fields produces another vector field and is defined by how it acts on functions, for any smooth function $g$: $[X,F](g) = X(F(g)) - F(X(g))$} 

For Hamiltonian systems, every Hamiltonian symmetry is also a dynamical symmetry.
In fact, it is not hard to show that the Lie and Poisson brackets are related,
$$[X_f,X_g] = X_{\{f,g\}}$$ and this directly shows the implication. If $X_f$ is a Hamiltonian symmetry, $\{f,H\} = 0$, and then
\begin{equation*}
    [X_f,F] = [X_f,X_H] = X_{\{f,H\}} = 0.
\end{equation*}

However, the converse is not true, dynamical symmetries of a Hamiltonian system are not necessarily Hamiltonian symmetries and thus might not correspond to conserved quantities.
Furthermore even if the system has a dynamical symmetry which is the flow along a Hamiltonian vector field $\phi^X_\lambda$, $X=X_f = \{f,\cdot\}$, but the dynamics $F$ are not Hamiltonian, then the dynamics will not conserve $f$ in general. Both the symmetry and the dynamics must be Hamiltonian for the conservation laws. 

\begin{figure}[t]
	\includegraphics[width=0.45\textwidth]{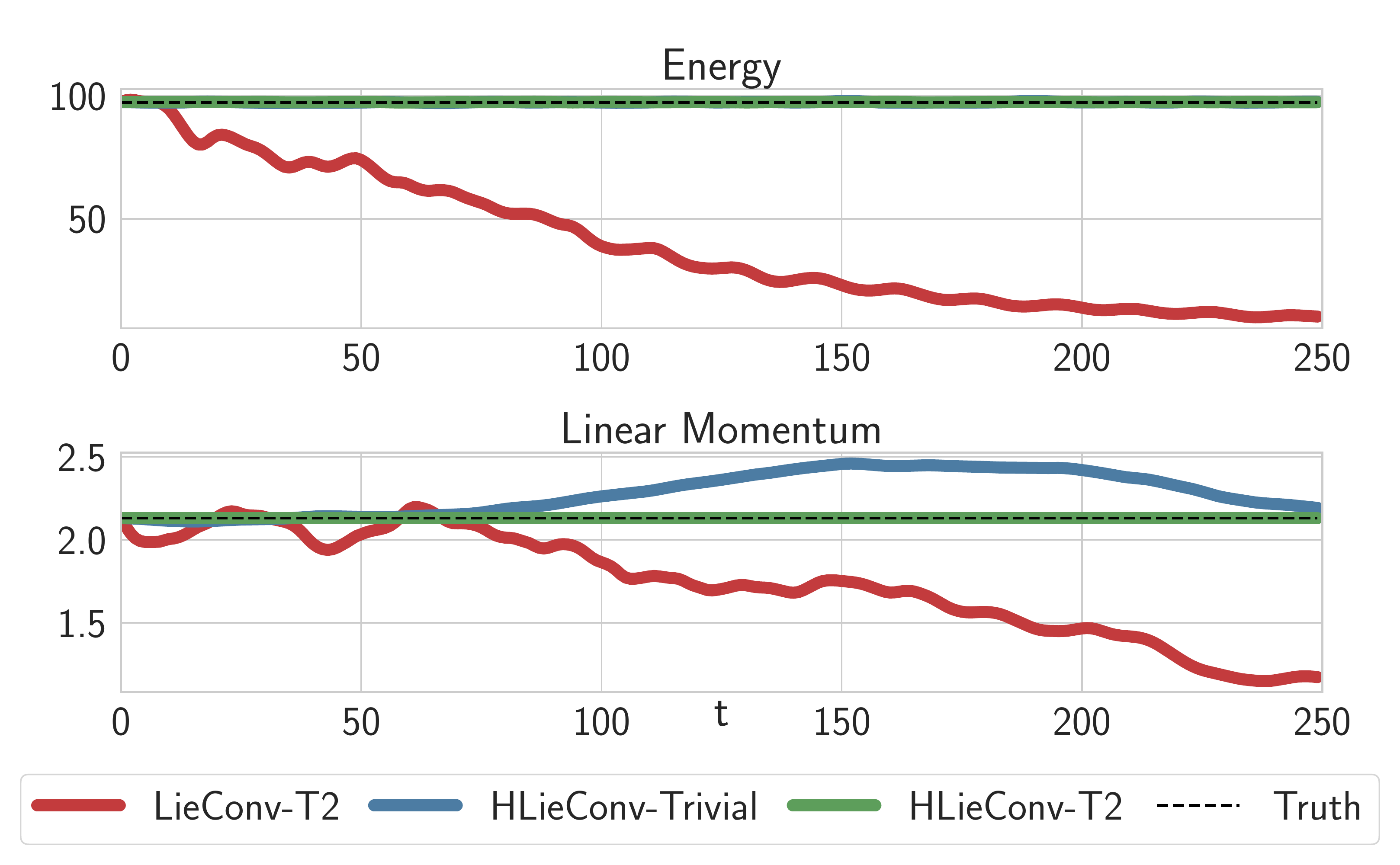}
	\caption{Equivariance alone is not sufficient, for conservation we need \textit{both} to model $\mathcal{H}$ and incorporate the given symmmetry. For comparison, LieConv-T(2) is T(2)-equivariant but models $F$, and HLieConv-Trivial models $\mathcal{H}$ but is not T(2)-equivariant. Only HLieConv-T(2) conserves linear momentum.
	}
	\label{fig:dynamical_symmetry}
\end{figure}

This fact is demonstrated by Figure \ref{fig:dynamical_symmetry}, where the dynamics of the (non-Hamiltonian) equivariant LieConv-\T{2} model has a \T{2} dynamical symmetry with the generators $\partial_{x}, \partial_{y}$ which are Hamiltonian vector fields for $f=p_x,f=p_y$, and yet linear momentum is not conserved by the model.

\newpage
\textbf{Conserving Linear and Angular Momentum}

Consider a system of $N$ interacting particles described in Euclidean coordinates with position and momentum $q_{im},p_{im}$, such as the multi-body spring problem.  Here the first index $i=1,2,3$ indexes the spatial coordinates and the second $m=1,2,...,N$ indexes the particles. We will use the bolded notation $\mathbf{q}_m,\mathbf{p}_m$ to suppress the spatial indices, but still indexing the particles $m$ as in Section \ref{subsec:predicting_trajectories}.

The total linear momentum along a given direction $\mathbf{n}$ is $\mathbf{n}\cdot\mathbf{P} = \sum_{i,m} n_{i} p_{im} = \mathbf{n}\cdot(\sum_m \mathbf{p}_m)$. Expanding the Poisson bracket, the Hamiltonian vector field 
\begin{equation*}
    X_{\mathbf{n}\mathbf{P}} = \{\mathbf{n}\cdot\mathbf{P},\cdot\} = \sum_{i,m}n_i \frac{\partial}{\partial q_{im}} = \mathbf{n}\cdot \sum_m \frac{\partial}{\partial \mathbf{q}_m}
\end{equation*} which has the flow $\phi^{X_{\mathbf{n}\mathbf{P}}}_\lambda(\mathbf{q}_m,\mathbf{p}_m) = (\mathbf{q}_m + \lambda \mathbf{n},\mathbf{p}_m)$, a translation of all particles by $\lambda \mathbf{n}$. So our model of the Hamiltonian conserves linear momentum if and only if it is invariant to a global translation of all particles, (e.g. \T{2} invariance for a 2D spring system).

The total angular momentum along a given axis $\mathbf{n}$ is 
\begin{equation*}
    \mathbf{n}\cdot\mathbf{L} =\mathbf{n}\cdot \sum_m \mathbf{q}_m\times \mathbf{p}_m = \sum_{i,j,k,m}\epsilon_{ijk}n_i q_{jm}p_{km} =  \sum_m\mathbf{p}_m^TA\mathbf{q}_m
\end{equation*}, where $\epsilon_{ijk}$ is the Levi-Civita symbol and we have defined the antisymmetric matrix $A$ by $A_{kj} = \sum_{i} \epsilon_{ijk}n_i$.
\begin{align*}
     X_{\mathbf{n}\mathbf{L}} &= \{\mathbf{n}\cdot\mathbf{L},\cdot\} = \sum_{j,k,m}A_{kj}q_{jm} \frac{\partial}{\partial q_{km}} - A_{jk}p_{jm}\frac{\partial}{\partial p_{km}}\\
     X_{\mathbf{n}\mathbf{L}}&= \sum_m \big(\mathbf{q}_m^TA^T\frac{\partial}{\partial \mathbf{q}_m} + \mathbf{p}_m^TA^T\frac{\partial}{\partial \mathbf{p}_m}\big)
\end{align*}
where the second line follows from the antisymmetry of $A$. We can find the flow of $X_{\mathbf{n}\mathbf{L}}$ from the differential equations $\dot{\mathbf{q}}_m = A\mathbf{q}, \dot{\mathbf{p}}_m = A\mathbf{q}$ which have the solution 
\begin{equation*}
    \phi^{X_{\mathbf{n}\mathbf{L}}}_\theta(\mathbf{q}_m,\mathbf{p}_m) = (e^{\theta A} \mathbf{q}_m,e^{\theta A}\mathbf{p}_m) = (R_\theta\mathbf{q}_m,R_\theta\mathbf{p}_m),
\end{equation*}
 where $R_\theta$ is a rotation about the axis $\mathbf{n}$ by the angle $\theta$, which follows from the Rodriguez rotation formula. Therefore, the flow of the Hamiltonian vector field of angular momentum along a given axis is a global rotation of the position and momentum of each particle about that axis. Again, the dynamics of a neural network modeling a Hamiltonian conserve total angular momentum if and only if the network is invariant to simultaneous rotation of all particle positions and momenta.

\section{Additional Experiments}
\subsection{Equivariance Demo}
While \eqref{eq:group_conv_discretization} shows that the convolution estimator is equivariant, we have conducted the ablation study below examining the equivariance of the network empirically. We trained LieConv (\triv, \T{3}, \SO{3}, \SE{3}) models on a limited subset of 20k training examples (out of 100k) of the HOMO task on QM9 without any data augmentation. We then evaluate these models on a series of modified test sets where each example has been randomly transformed by an element of the given group (the test translations in \T{3} and \SE{3} are sampled from a normal with stddev 0.5). In table \ref{table:transformed_test} the rows are the models configured with a given group equivariance and the columns N/G denote no augmentation at training time and transformations from G applied to the test set (test translations in \T{3} and \SE{3} are sampled from a normal with stddev 0.5).

\begin{table}[h]
    \centering
    \label{table:transformed_test}
    \small
    \begin{tabular}{ccccc}
    \toprule
     Model & N/N & N/\T{3} & N/\SO{3} & N/\SE{3}\\
     \midrule
     \triv & \textbf{173} &   183 &    239   &   243   \\
     \T{3}& \textbf{113}  &   \textbf{113} &   133    &   133   \\
     \SO{3}&\textbf{159}  &  238 &   \textbf{160}   &    240   \\
     \SE{3}&\textbf{62}  &  \textbf{62} &   \textbf{63}   &    \textbf{62}   \\
    \end{tabular}
    \caption{Test MAE (in meV) on HOMO test set randomly transformed by elements of $G$. Despite no data augmentation (N), $G$ equivariant models perform as well on $G$ transformed test data.}
    
\end{table}

Notably, the performance of the LieConv-G models do not degrade when random G transformations are applied to the test set. Also, in this low data regime, the added equivariances are especially important. 

\subsection{RotMNIST Comparison}
While the RotMNIST dataset consists of 12k rotated MNIST digits, it is standard to separate out 10k to be used for training and 2k for validation. However, in Ti-Pooling and E(2)-Steerable CNNs,
it appears that after hyperparameters were tuned the validation set is folded back into the training set to be used as additional training data, a common approach used on other datasets. Although in table \ref{table:rotmnist} we only use 10k training points, in the table below we report the performance with and without augmentation trained on the full 12k examples. 
\begin{table}[h]
    \centering
    \label{table:mnist_12k}
    \vspace{0.25cm}
    \small
    \begin{tabular}{ccccccc}
    \toprule
     Aug &Trivial & $\mathrm{T}_y$ & \T{2} & \SO{2} & \SO{2}$\times \mathbb{R}^*$ & \SE{2} \\
     \midrule
     \SO{2}&1.44 & 1.35 & 1.32 & 1.27 &  1.13 & 1.13 \\
     None &1.60 & 2.64 & 2.34 & 1.26 &  1.25 & 1.15 \\
    \end{tabular}
    \caption{Classification Error (\%) on RotMNIST dataset for LieConv with different group equivariances and baselines:}
\end{table}

\section{Implementation Details}

\subsection{Practical Considerations} \label{app:practical_details}

While the high-level summary of the lifting procedure (Algorithm \ref{alg:lifting}) and the LieConv layer (Algorithm \ref{alg:lie_conv}) provides a useful conceptual understanding of our method, there are some additional details that are important for a practical implementation.
\begin{enumerate}
    \item According to Algorithm \ref{alg:lie_conv}, $a_{ij}$ is computed in every LieConv layer, which is 
both highly redundant and costly. In practice, we precompute $a_{ij}$ once after 
lifting and feed it through the network with layers operating on the state 
$\big(\{a_{ij}\}_{i,j}^{N,N}, \{f_i\}_{i=1}^N\big)$ instead of $\{(u_i,q_i,f_i)\}_{i=1}^N$. 
Doing so requires fixing the group elements that will be used at each layer for a given forwards pass.
    \item In practice only $p$ elements of $\mathrm{nbhd}_i$ are sampled (randomly) for computing the Monte Carlo estimator in order to limit the computational burden (see Appendix \ref{sec:equivariant_subsampling}). 
    \item We use the analytic forms for the exponential and logarithm maps of the various groups as described in \citet{eade2014lie}.
\end{enumerate}

\subsection{Sampling from the Haar Measure for Various groups}\label{subsec:haar}
When the lifting map from $\mX \to G \times \mX/G$ is multi-valued, we need to sample elements
of $u \in G$ that project down to $x$: $uo =x$ in a way consistent with the Haar measure $\mu(\cdot)$. In other words, since the restriction $\mu(\cdot)|_\mathrm{nbhd}$ is a distribution, then we must sample from the conditional distribution $u \sim \mu(u|uo=x)|_\mathrm{nbhd}$. In general this can be done by parametrizing the distribution of $\mu$ as a collection of random variables that includes $x$, and then sampling the remaining variables. 

In this paper, the groups we use in which the lifting map is multi-valued are \SE{2}, \SO{3}, and \SE{3}. The process is especially straightforward for \SE{2} and \SE{3} as these groups can be expressed as a semi-direct product of two groups $G = H\ltimes N$,  
\begin{equation}
    d\mu_{G}(h,n) = \delta(h)d\mu_H(h)d\mu_N(n),
\end{equation}
where $\delta(h) = \frac{d\mu_N(n)}{d\mu_N(hnh^{-1})}$ \citep{willson2009reiter}.
For $G=\mathrm{SE}(d) = \mathrm{SO}(d)\ltimes \mathrm{T}(d)$, $\; \delta(h)=1$ since the 
Lebesgue measure $d\mu_{\mathrm{T}(d)}(x) = d\lambda(x) = dx$ is invariant to rotations. So simply $d\mu_{\mathrm{SE}(d)}(R,x) = d\mu_{\mathrm{SO}(d)}(R)dx$.

So lifts of a point $x\in \mX$ to \SE{d} consistent with the $\mu$ are just $T_x R$, the multiplication of a translation by $x$ and randomly sampled rotations $R \sim \mu_{\mathrm{SO}(d)}(\cdot)$. There are multiple easy methods to sample uniformly from \SO{d} given in \cite{kuffner2004effective}, for example sampling uniformly from \SO{3} can be done by sampling a unit quaternion from the 3-sphere, and identifying it with the corresponding rotation matrix.

\subsection{Model Architecture}\label{sec:architecture}

We employ a ResNet-style architecture \citep{he2016deep}, using bottleneck blocks \citep{zagoruyko2016wide}, and replacing ReLUs with Swish activations \citep{ramachandran2017searching}. The convolutional kernel $g_\theta$ internal to each LieConv layer is parametrized by a 3-layer MLP with 32 hidden units, batch norm, and Swish nonlinearities. Not only do the Swish activations improve performance slightly, but unlike ReLUs they are twice differentiable which is a requirement for backpropagating through the Hamiltonian dynamics. The stack of elementwise linear and bottleneck blocks is followed by a global pooling layer that computes the average over all elements, but not over channels.
Like for regular image bottleneck blocks, the channels for the convolutional layer in the middle are smaller by a factor of 4 for increased parameter and computational efficiency.

\textbf{Downsampling:}  As is traditional for image data, we increase the number of channels and the receptive field at every downsampling step. The downsampling is performed with the farthest point downsampling method described in Appendix \ref{sec:equivariant_subsampling}. 
For a downsampling by a factor of $s < 1$, the radius of the neighborhood is scaled up by $s^{-1/2}$ and the channels are scaled up by $s^{-1/2}$. When an image is downsampled with $s=(1/2)^2$ that is typical in a CNN, this results in 2x more channels and a radius or dilation of 2x. In the bottleneck block, the downsampling operation is fused with the LieConv layer, so that the convolution is only evaluated at the downsampled query locations. We perform downsampling only on the image datasets, which have more points.

\textbf{BatchNorm:}
In order to handle the varied number of group elements per example and within each neighborhood, we use a modified batchnorm that computes statistics only over elements from a given mask. The batch norm is computed per channel, with statistics averaged over the batch size and each of the valid locations.
 \FloatBarrier
\subsection{Details for Hamiltonian Models}\label{appendix:ham_models}

\textbf{Model Symmetries:}

 As the position vectors are mean centered in the model forward pass $\position_i' = \position_i - \; \bar\position$, HOGN and HLieConv-SO2* have additional \T{2} invariance, yielding \SE{2} invariance for HLieConv-SO2*. We also experimented with a HLieConv-SE2 equivariant model, but found that the exponential map for SE2 (involving taylor expands and masking) was not numerically stable enough for for second derivatives, required for optimizing through the Hamiltonian dynamics. So instead we benchmark the HLieConv-SO2 (without centering) and the HLieConv-SO2* (with centering) models separately. Layer equivariance is preferable for not prematurely discarding useful information and for better modeling performance, but invariance alone is sufficient for the conservation laws. Additionally, since we know \textit{a priori} that the spring problem has Euclidean coordinates, we need not model the kinetic energy $K(\momentum, m) = \sum_{j=1}^n \|\momentum_j\|^2/m_j$ and instead focus on modeling the potential $V(\position, k)$. We observe that this additional inductive bias of Euclidean coordinates improves model performance. Table \ref{table:model_variants} shows the invariance and equivariance properties of the relevant models and baselines. For Noether conservation, we need both to model the Hamiltonian and have the symmetry property.

\begin{center}
\begin{table}
    \begin{tabular}{|c|c|c|c|c|}
         \hline
         & $F(\state, t)$ & $\mathcal{H}(\state, t)$ & $T(2)$ & $SO(2)$ \\
         \hline
         FC & $\bullet$ & & & \\ \hline
         OGN & $\bullet$ & & &\\ \hline
         HOGN & & $\bullet$ & \ding{72} & \\ \hline
         LieConv-T(2) & $\bullet$ &  & \ding{74} & \\ \hline
         HLieConv-Trivial & & $\bullet$ & & \\ \hline
         HLieConv-T(2) & & $\bullet$ & \ding{74} &  \\ \hline
         HLieConv-SO(2) & & $\bullet$ &  & \ding{74} \\ \hline
         HLieConv-SO(2)* & & $\bullet$ & \ding{72} & \ding{74} \\ \hline
    \end{tabular}
\caption{Model characteristics. Models with layers invariant to $G$ are denoted with \ding{72}, and those with equivariant layers with \ding{74}.}
\label{table:model_variants}
\end{table}
\end{center}

\textbf{Dataset Generation:} To generate the spring dynamics datasets we generated $D$ systems each with $N=6$ particles connected by springs. The system parameters, mass and spring constant, are set by sampling $\{m_1^{(i)}, \dots m_6^{(i)}, k_1^{(i)}, \dots, k_6^{(i)}\}_{i=1}^N$, $m_j^{(i)} \sim \mathcal{U}(0.1, 3.1)$, $k_j^{(i)} \sim \mathcal{U}(0, 5)$. Following \citet{sanchez2019hamiltonian}, we set the spring constants as $k_{ij} = k_ik_j$. For each system $i$, the position and momentum of body $j$ were distributed as $\position_j^{(i)} \sim \mathcal{N}(0, 0.16I)$, $\momentum_j^{(i)} \sim \mathcal{N}(0, 0.36I)$. Using the analytic form of the Hamiltonian for the spring problem, $\mathcal{H}(\position, \momentum) = K(\momentum, m) + V(\position, k)$, we use the RK4 numerical integration scheme to generate $5$ second ground truth trajectories broken up into 500 evaluation timesteps.
We use a fixed step size scheme for RK4 chosen automatically (as implemented in \citet{chen2018neural}) with a relative tolerance of 1e-8 in double precision arithmetic. We then randomly selected a single segment for each trajectory, consisting of an initial state $\state_t$ and $\tau=4$ transition states: $(\state_{t+1}^{(i)}, \dots, \state_{t + \tau}^{(i)})$.

\textbf{Training:}
All models were trained in single precision arithmetic (double precision did not make any appreciable difference) with an integrator tolerance of 1e-4. We use a cosine decay for the learning rate schedule and perform early stopping over the validation MSE. We trained with a minibatch size of $200$ and for $100$ epochs each using the Adam optimizer \citep{kingma2014adam} without batch normalization. With 3k training examples, the HLieConv model takes about 20 minutes to train on one 1080Ti.

For the examination of performance over the range of dataset sizes in \ref{fig:dynamics_scaling}, we cap the validation set to the size of the training set to make the setting more realistic, and we also scale the number of training epochs up as the size of the dataset shrinks (epochs $= 100(\sqrt{10^3/D})$) which we found to be sufficient to fit the training set. For $D\le200$ we use the full dataset in each minibatch.

\textbf{Hyperparameters:}
\begin{center}
    \begin{tabular}{cccc}
         \toprule
         & channels &  layers & lr \\
         \midrule
         (H)FC & 256 & 4 & 1e-2 \\
         (H)OGN & 256 & 1 & 1e-2 \\
         (H)LieConv & 384 & 4 & 1e-3 \\
         \bottomrule
    \end{tabular}
\end{center}

\textbf{Hyperparameter tuning:} 
Model hyperparameters were tuned by grid search over channel width, number of layers, and learning rate. The models were tuned with training, validation, and test datasets consisting of 3000, 2000, and 2000 trajectory segments respectively.

\subsection{Details for Image and Molecular Experiments}

\textbf{RotMNIST Hyperparameters:}
For RotMNIST we train each model for 500 epochs using the Adam optimizer with learning rate 3e-3 and batch size 25. The first linear layer maps the 1-channel grayscale input to $k=128$ channels, and the number of channels in the bottleneck blocks follow the scaling law from Appendix \ref{sec:architecture} as the group elements are downsampled. We use 6 bottleneck blocks, and the total downsampling factor $S=1/10$ is split geometrically between the blocks as $s = (1/10)^{1/6}$ per block. The initial radius $r$ of the local neighborhoods in the first layer is set so as to include 1/15 of the total number of elements in each neighborhood and is scaled accordingly. The subsampled neighborhood used to compute the Monte Carlo convolution estimator uses $p=25$ elements.  The models take less than 12 hours to train on a 1080Ti.

\textbf{QM9 Hyperparameters:}
For the QM9 molecular data, we use the featurization from \citet{anderson2019cormorant}, where the input features $f_i$ are determined by the atom type (C,H,N,O,F) and the atomic charge. The coordinates $x_i$ are simply the raw atomic coordinates measured in angstroms. A separate model is trained for each prediction task, all using the same hyperparameters and early stopping on the validation MAE. We use the same train, validation, test split as \citet{anderson2019cormorant}, with 100k molecules for train, 10\% for test and the remaining for validation. Like with the other experiments, we use a cosine learning rate decay schedule.
Each model is trained using the Adam optimizer for $1000$ epochs with a learning rate of 3e-3 and batch size of 100. We use \SO{3} data augmentation, 6 bottleneck blocks, each with $k=1536$ channels. The radius of the local neighborhood is set to $r=\infty$ to include all elements. The model takes about 48 hours to train on a single 1080Ti.

\begin{figure*}[h]
    \def \panelwidth {0.08\textwidth}
    \centering
    \subfigure[Trivial]{
    \begin{tabular}{ccc}
        \includegraphics[trim=10 10 10 10,clip,width=\panelwidth]{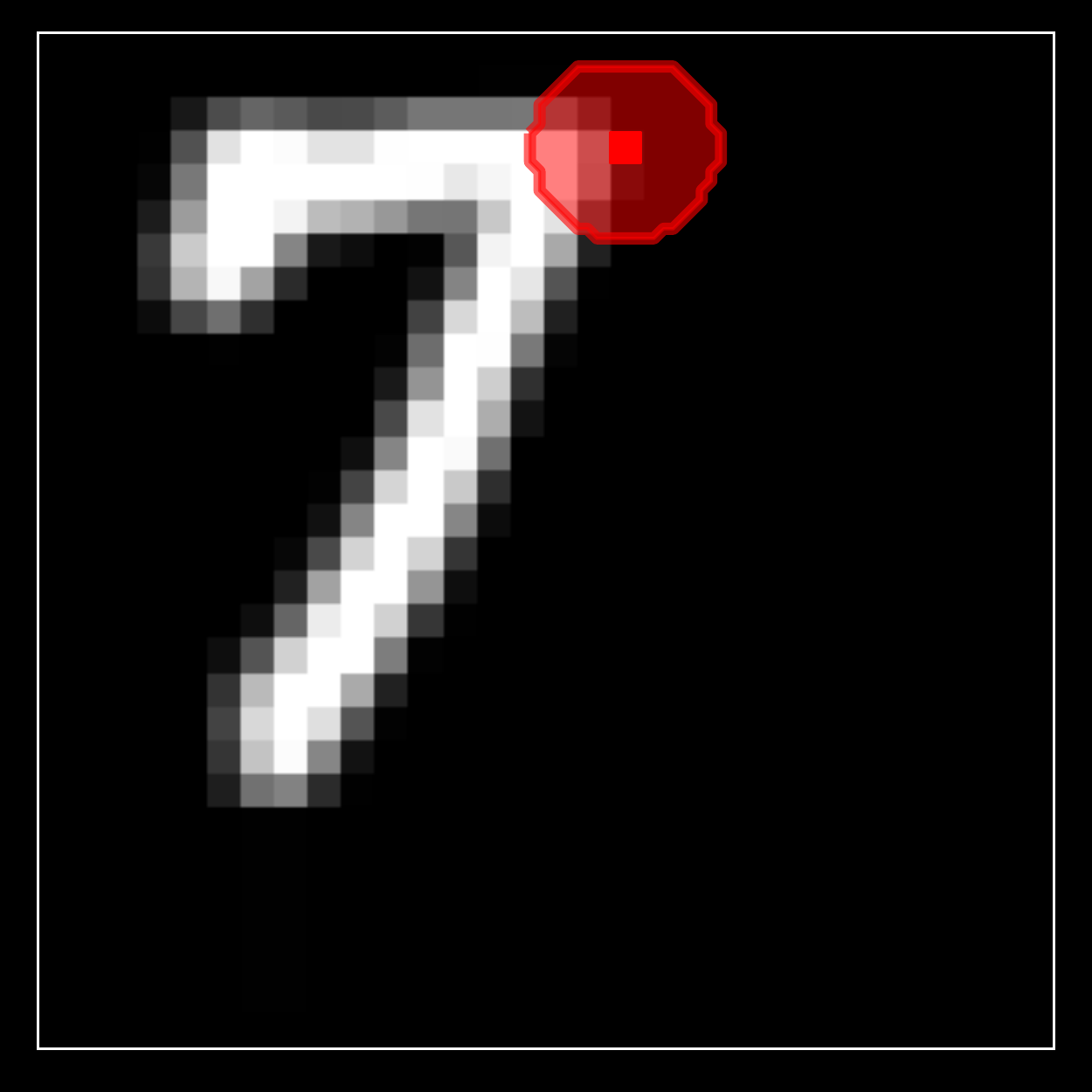}
        &
        \includegraphics[trim=10 10 10 10,clip,width=\panelwidth]{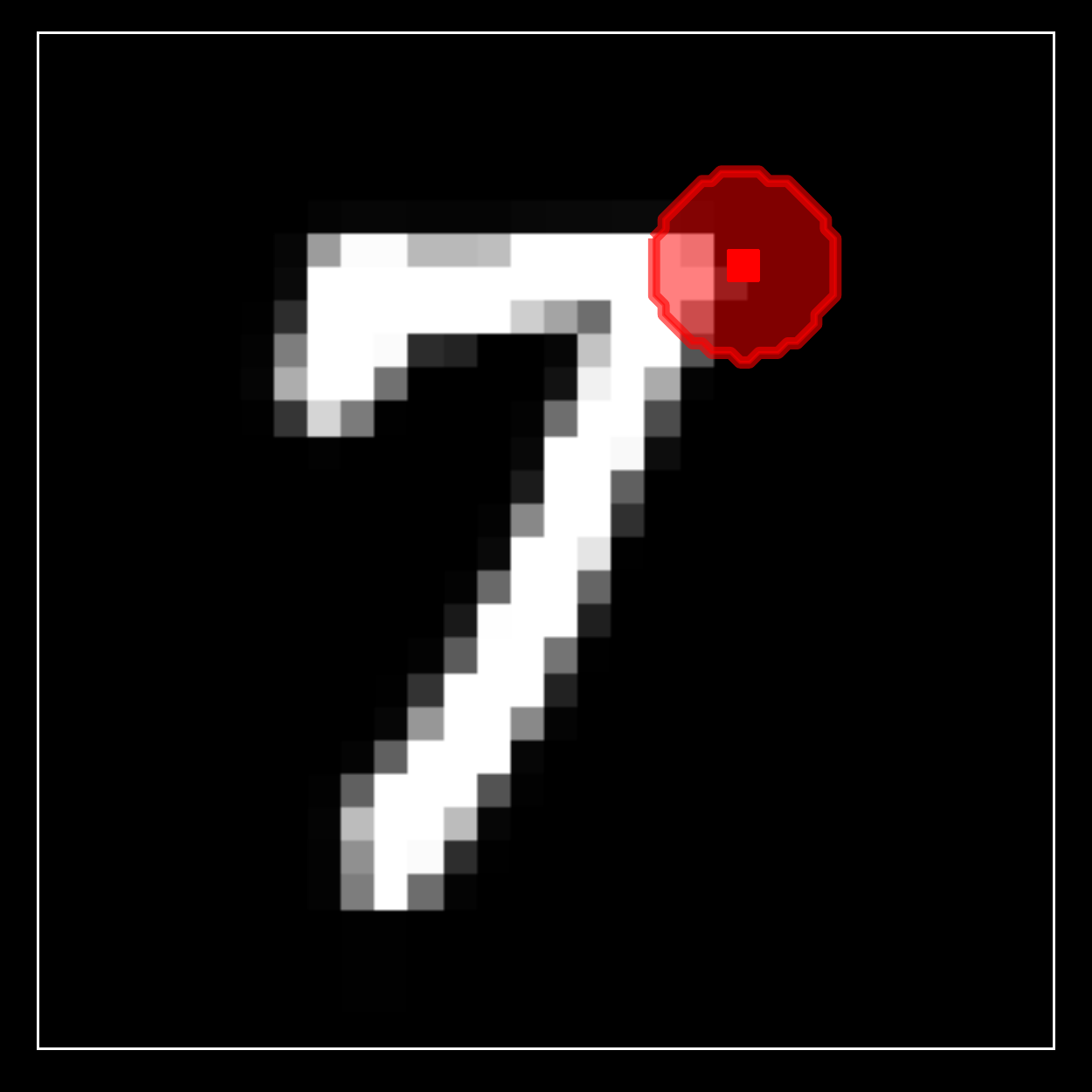}
        &
        \includegraphics[trim=10 10 10 10,clip,width=\panelwidth]{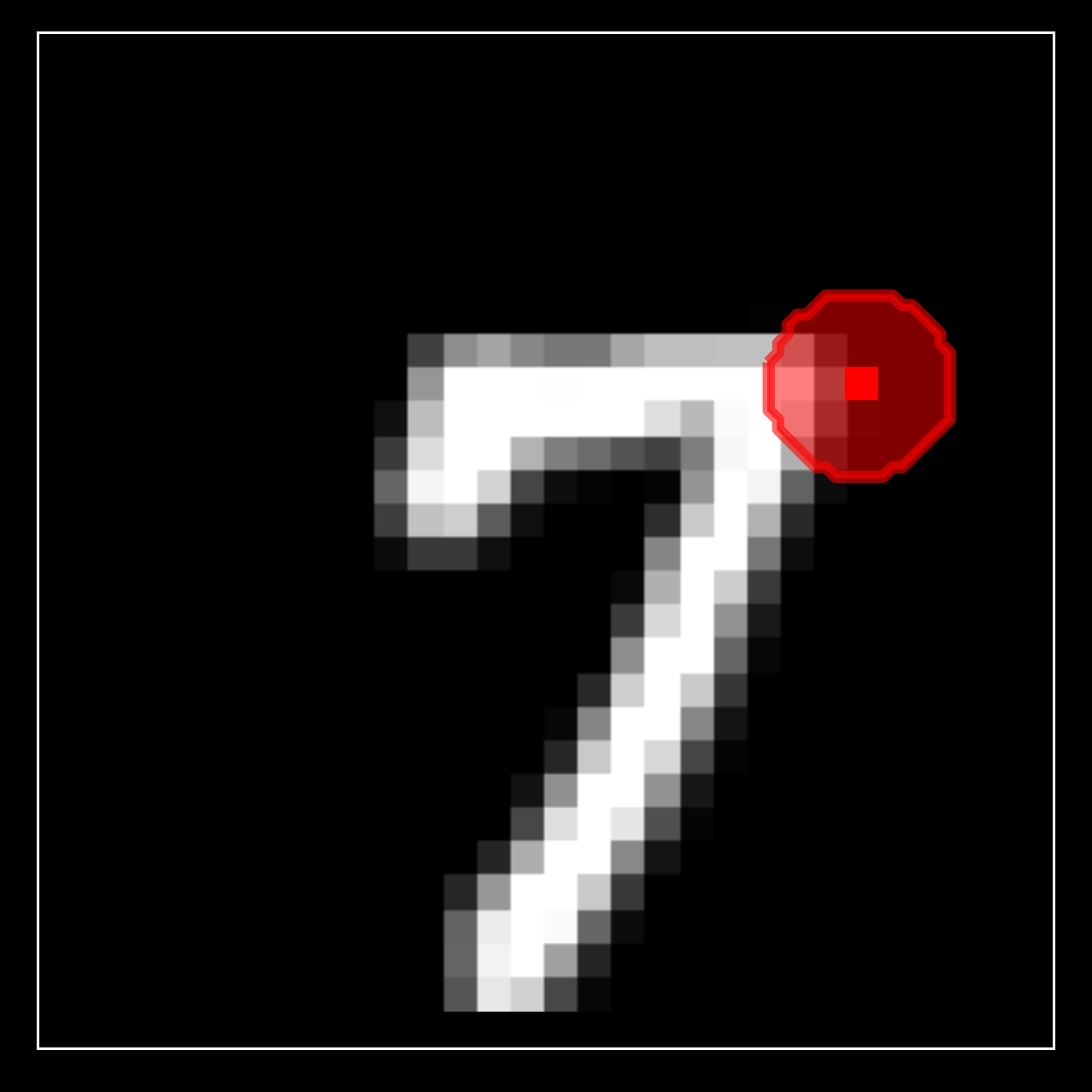}
        \\
        \includegraphics[trim=10 10 10 10,clip,width=\panelwidth]{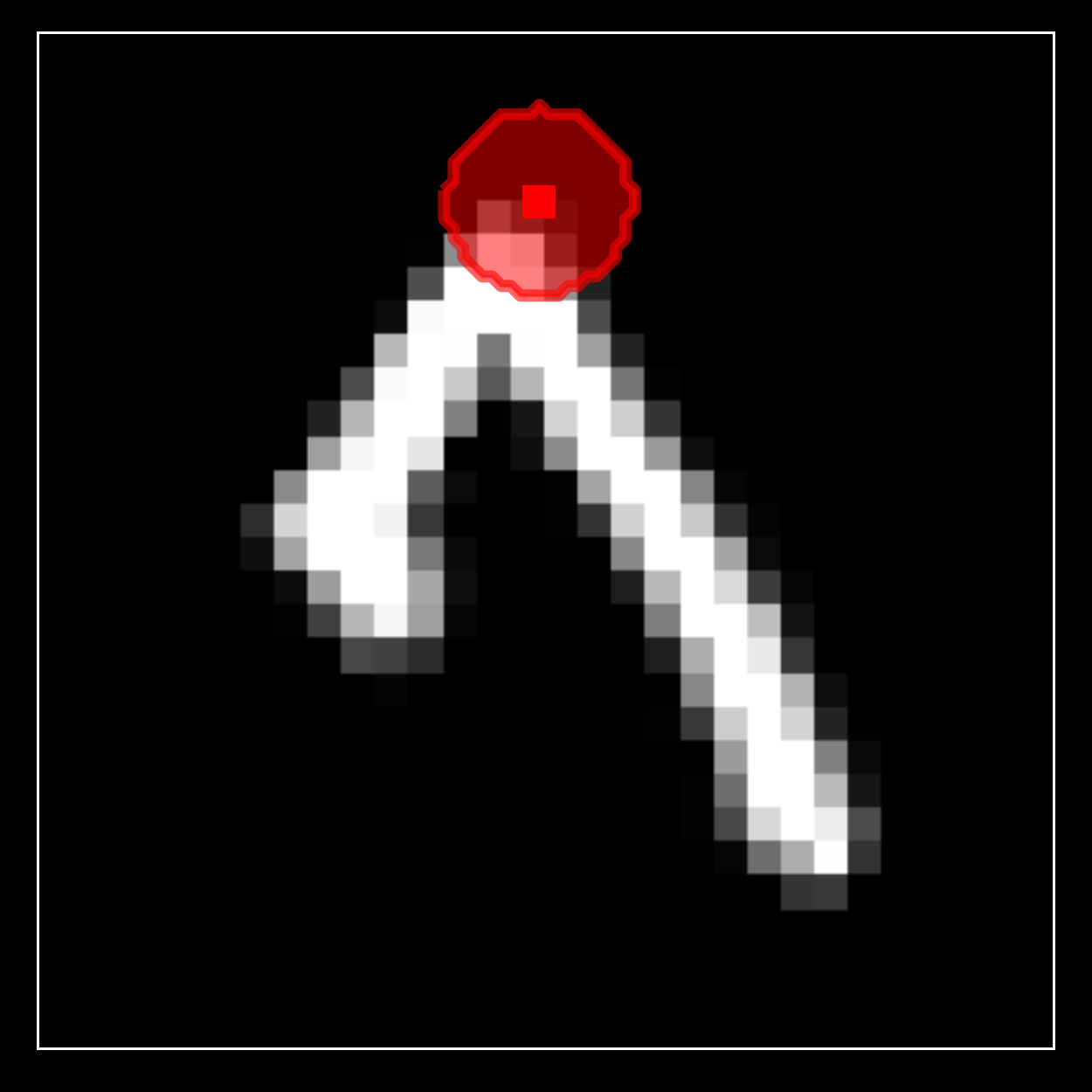}
        &
        \includegraphics[trim=10 10 10 10,clip,width=\panelwidth]{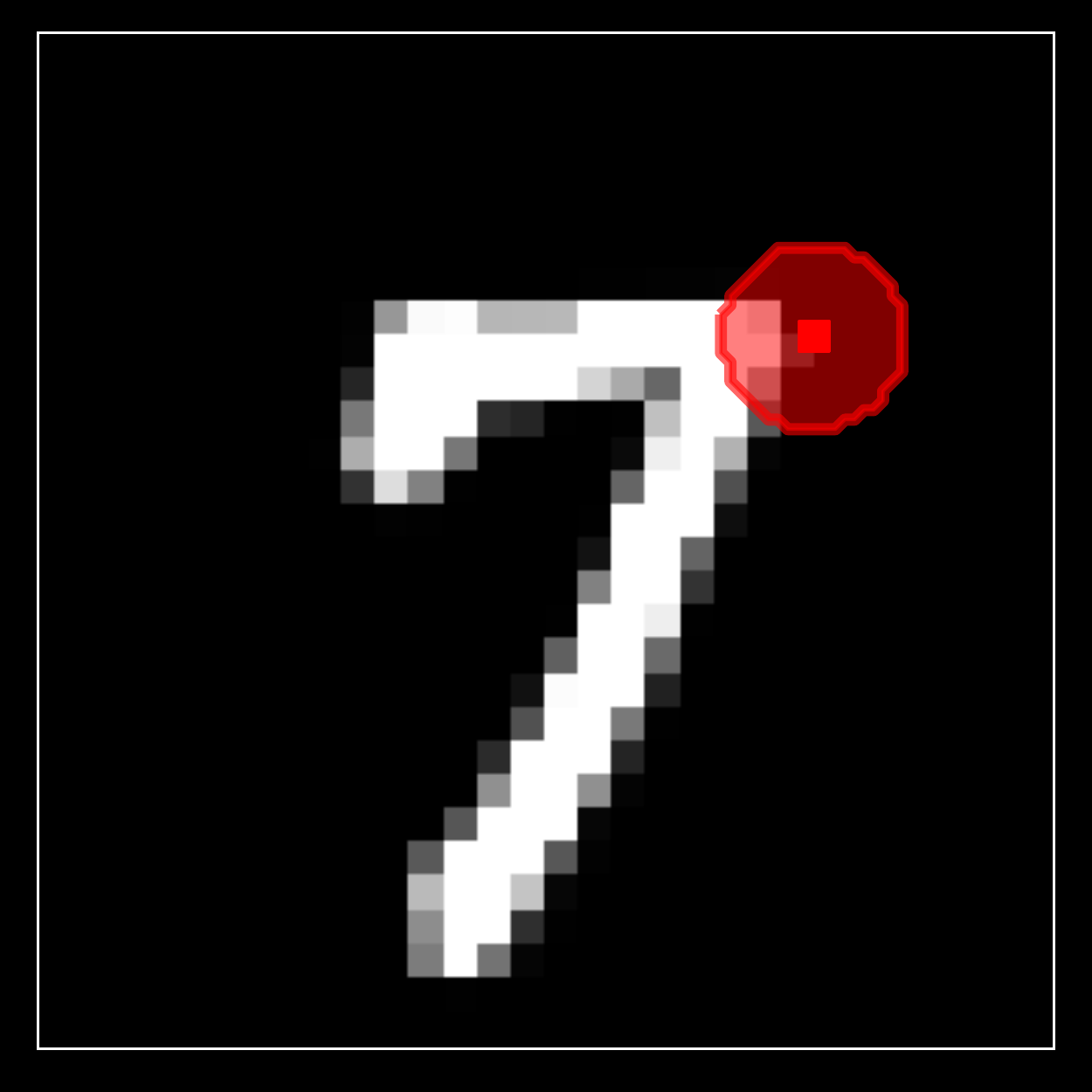}
        &
        \includegraphics[trim=10 10 10 10,clip,width=\panelwidth]{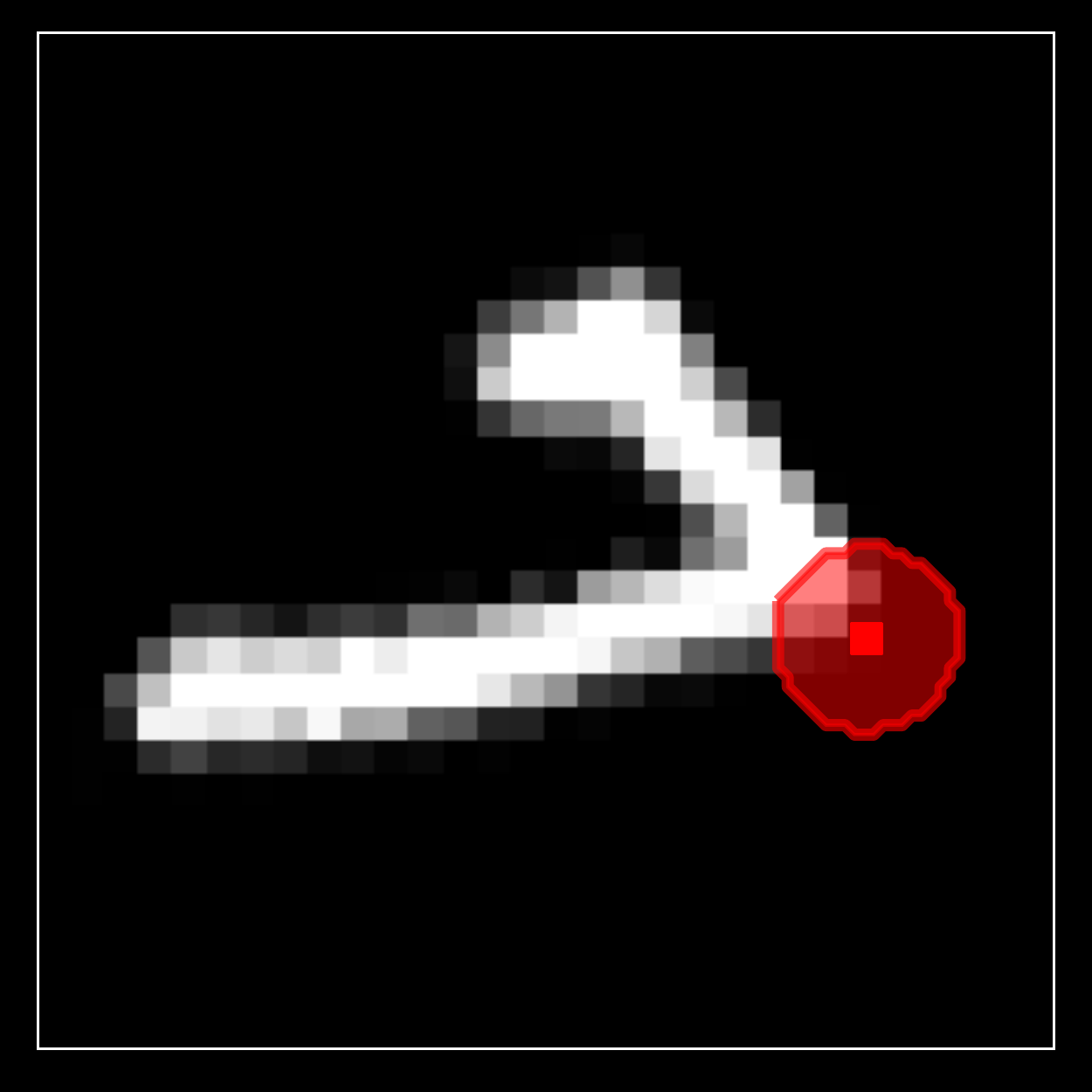}
        \\
        \includegraphics[trim=10 10 10 10,clip,width=\panelwidth]{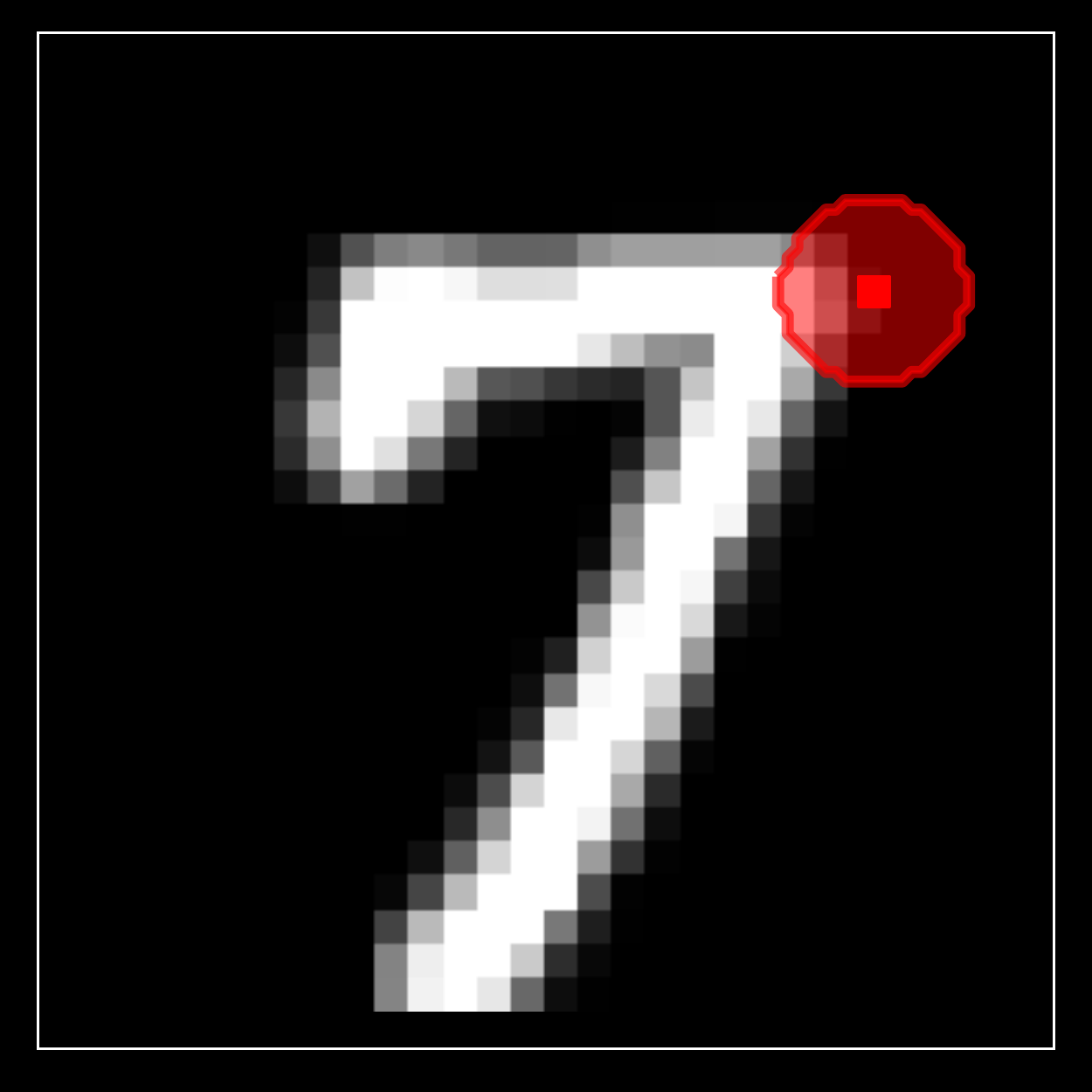}
        &
        \includegraphics[trim=10 10 10 10,clip,width=\panelwidth]{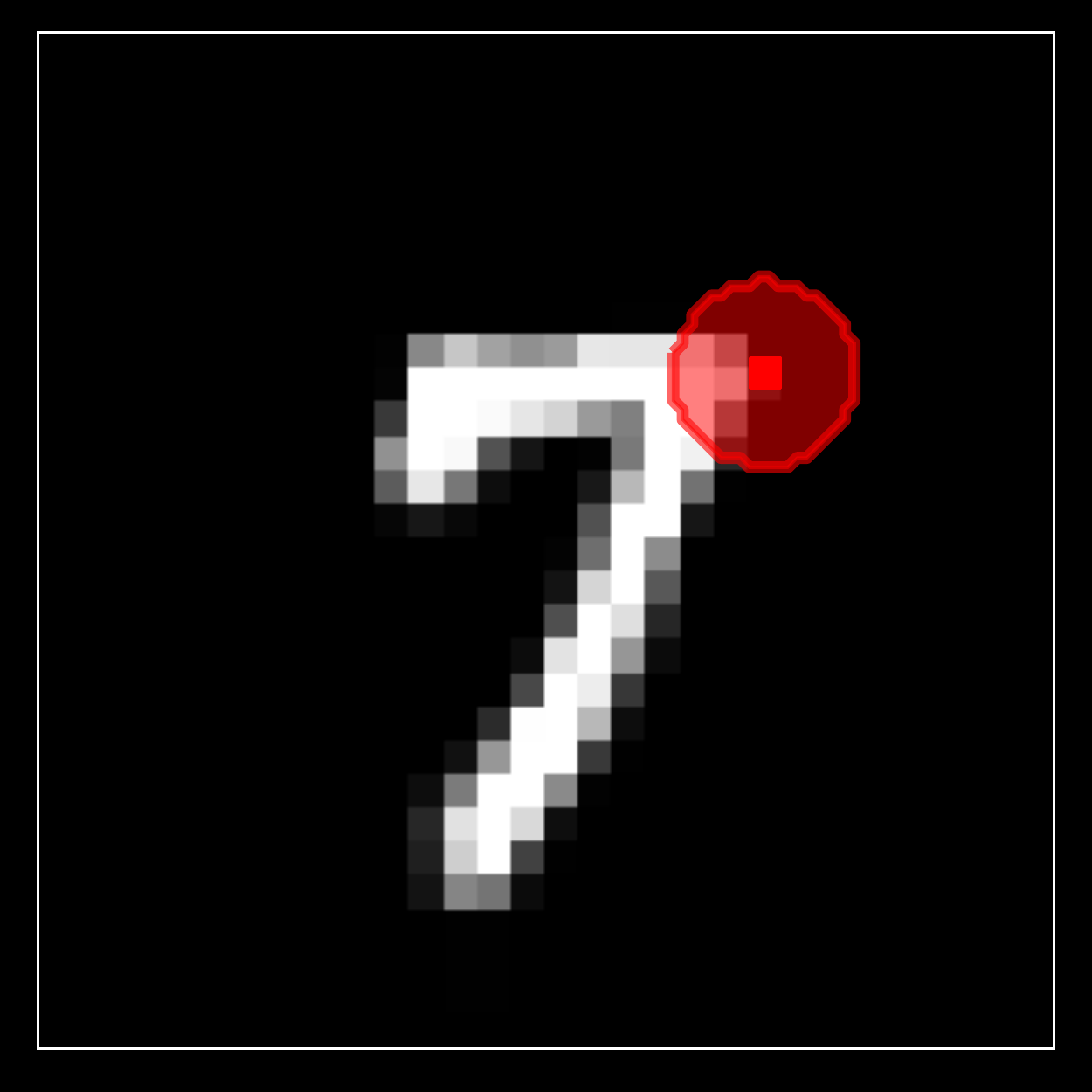}
        &
        \includegraphics[trim=10 10 10 10,clip,width=\panelwidth]{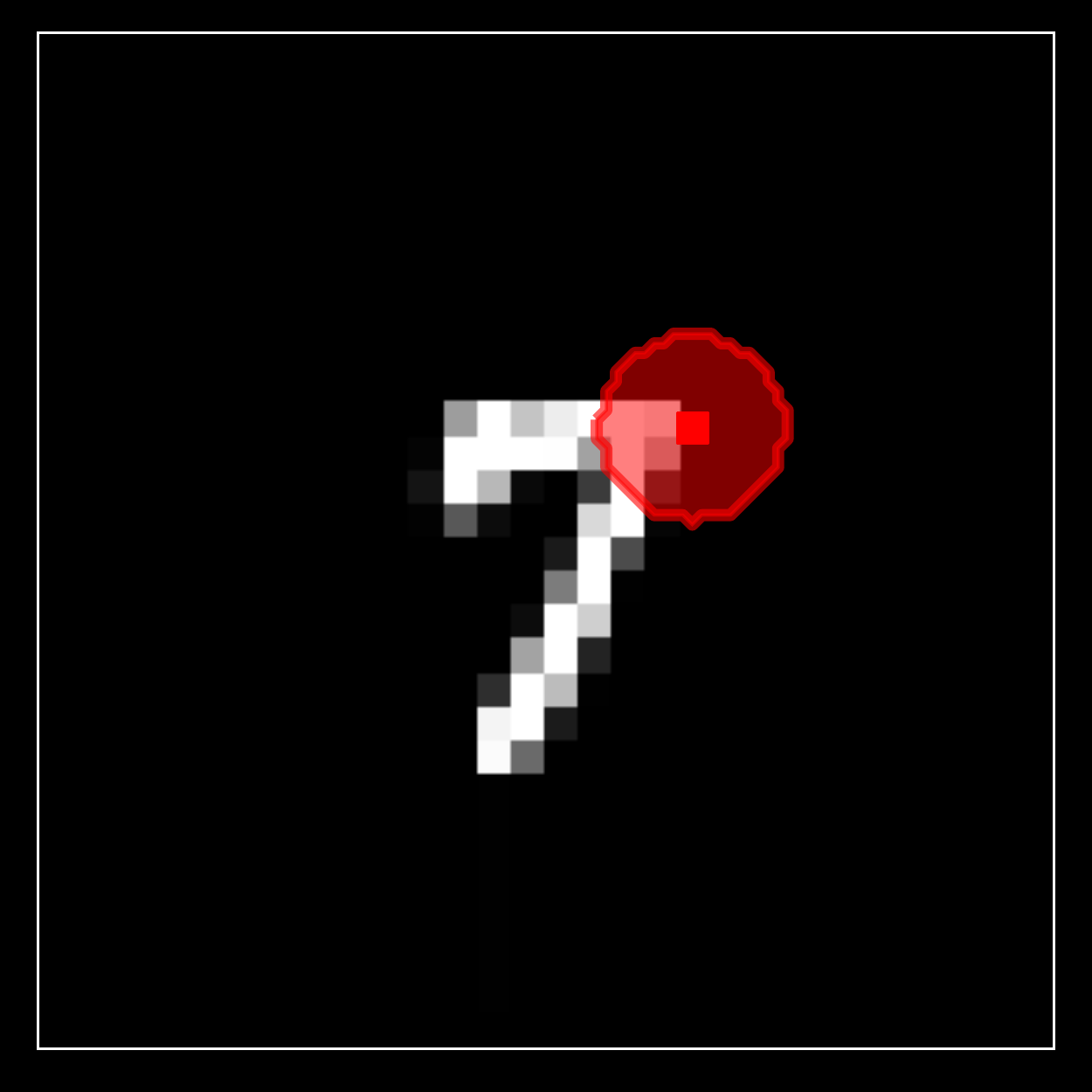}
        \\
    \end{tabular}
    }
    \subfigure[$T(2)$]{
    \begin{tabular}{ccc}
        \includegraphics[trim=10 10 10 10,clip,width=\panelwidth]{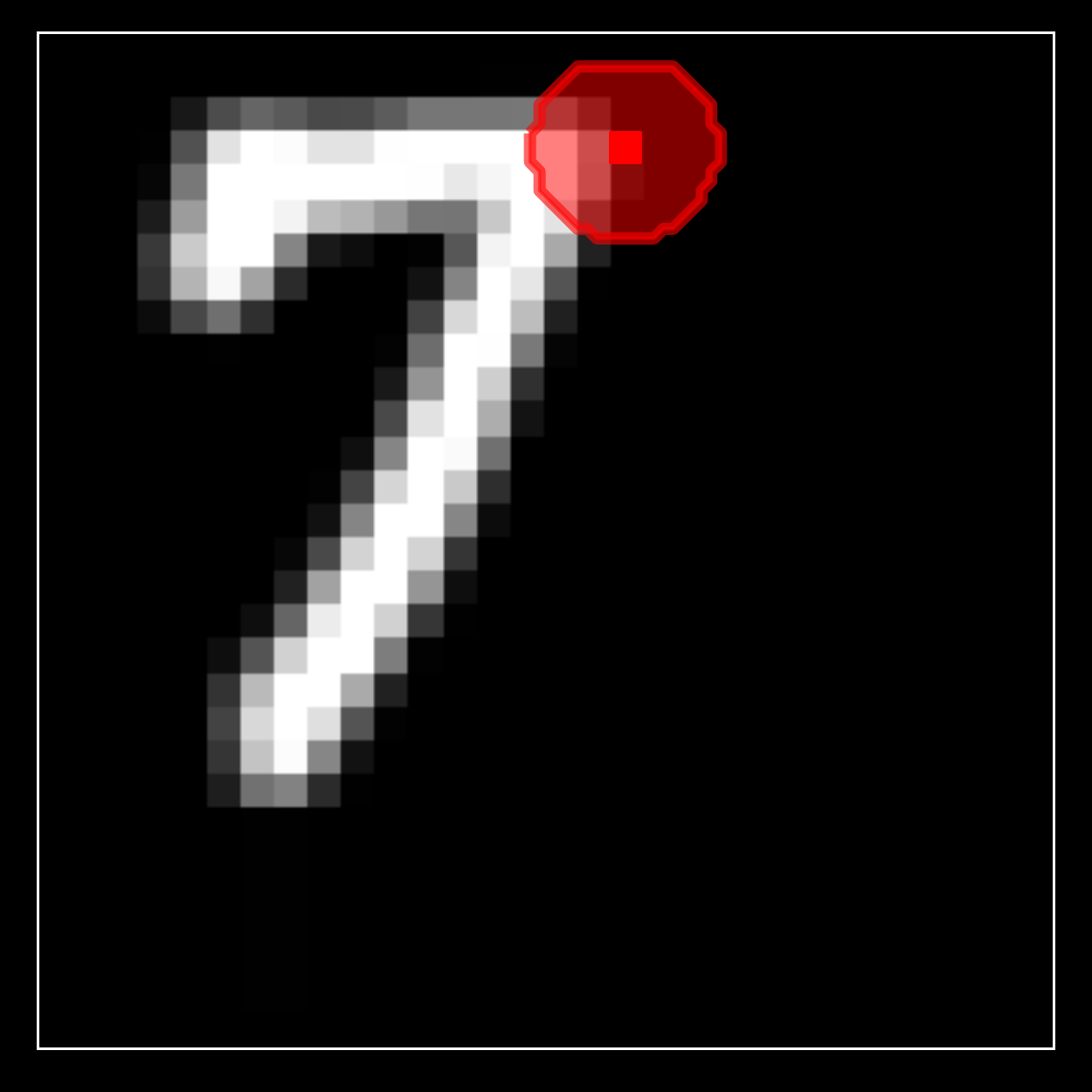}
        &
        \includegraphics[trim=10 10 10 10,clip,width=\panelwidth]{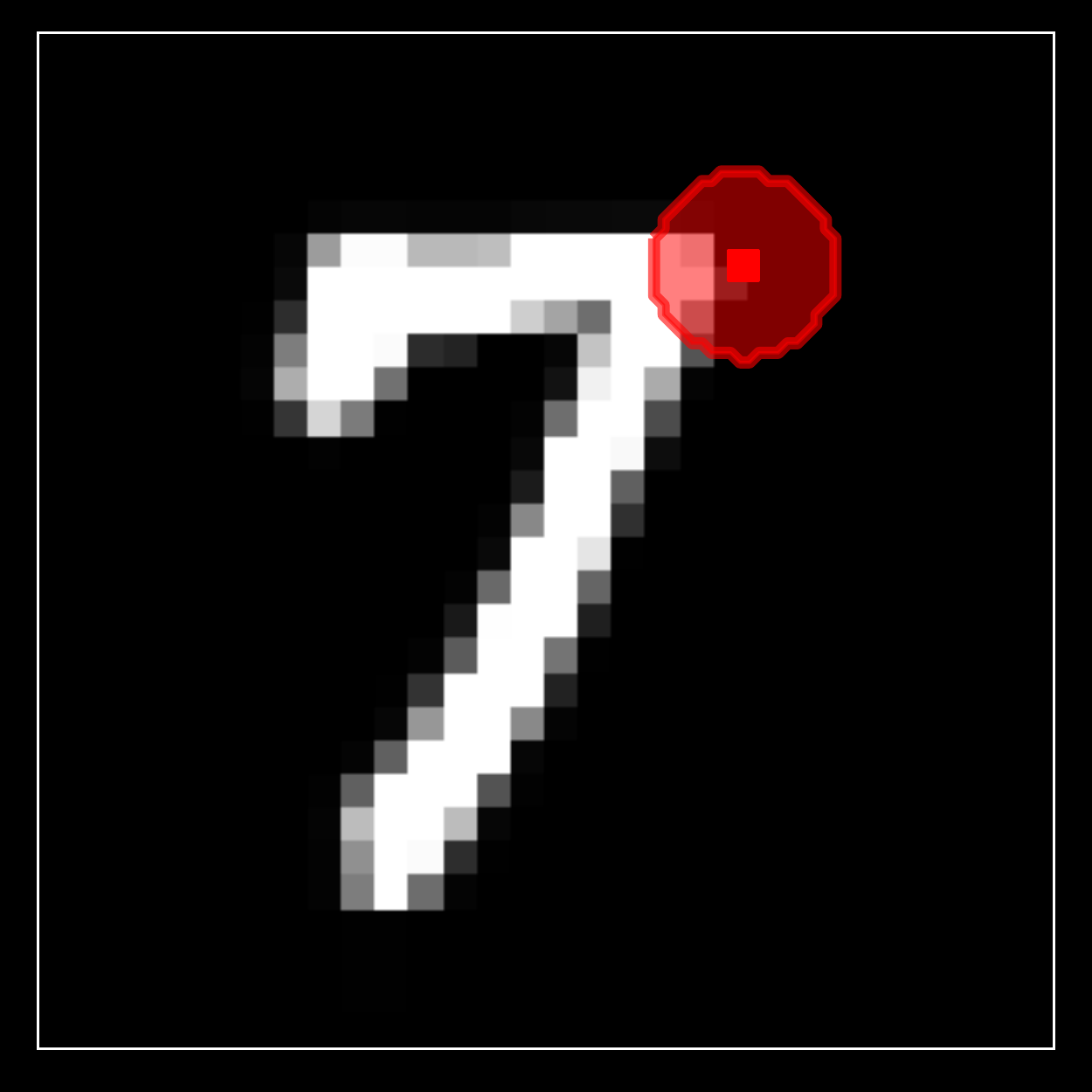}
        &
        \includegraphics[trim=10 10 10 10,clip,width=\panelwidth]{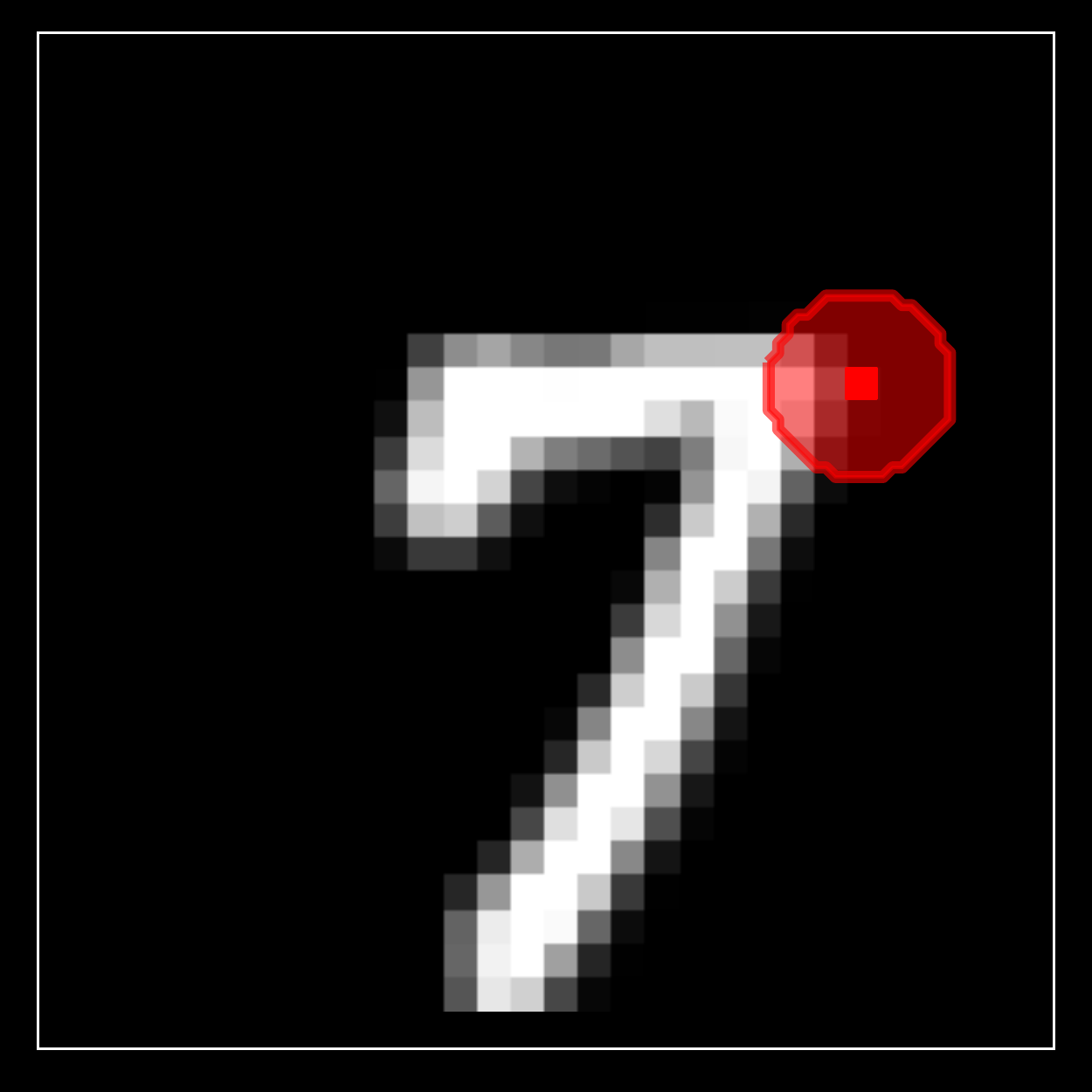}
        \\
        \includegraphics[trim=10 10 10 10,clip,width=\panelwidth]{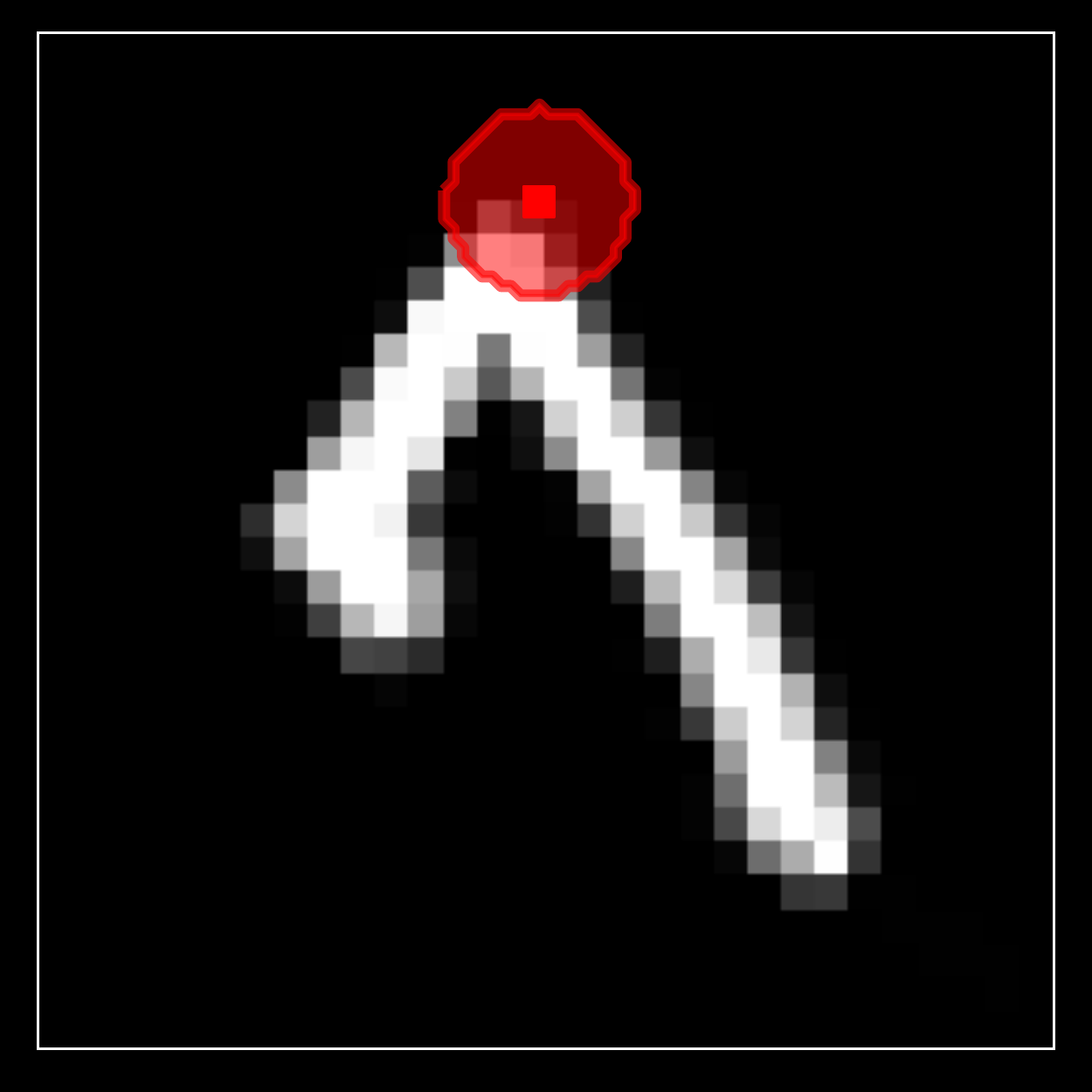}
        &
        \includegraphics[trim=10 10 10 10,clip,width=\panelwidth]{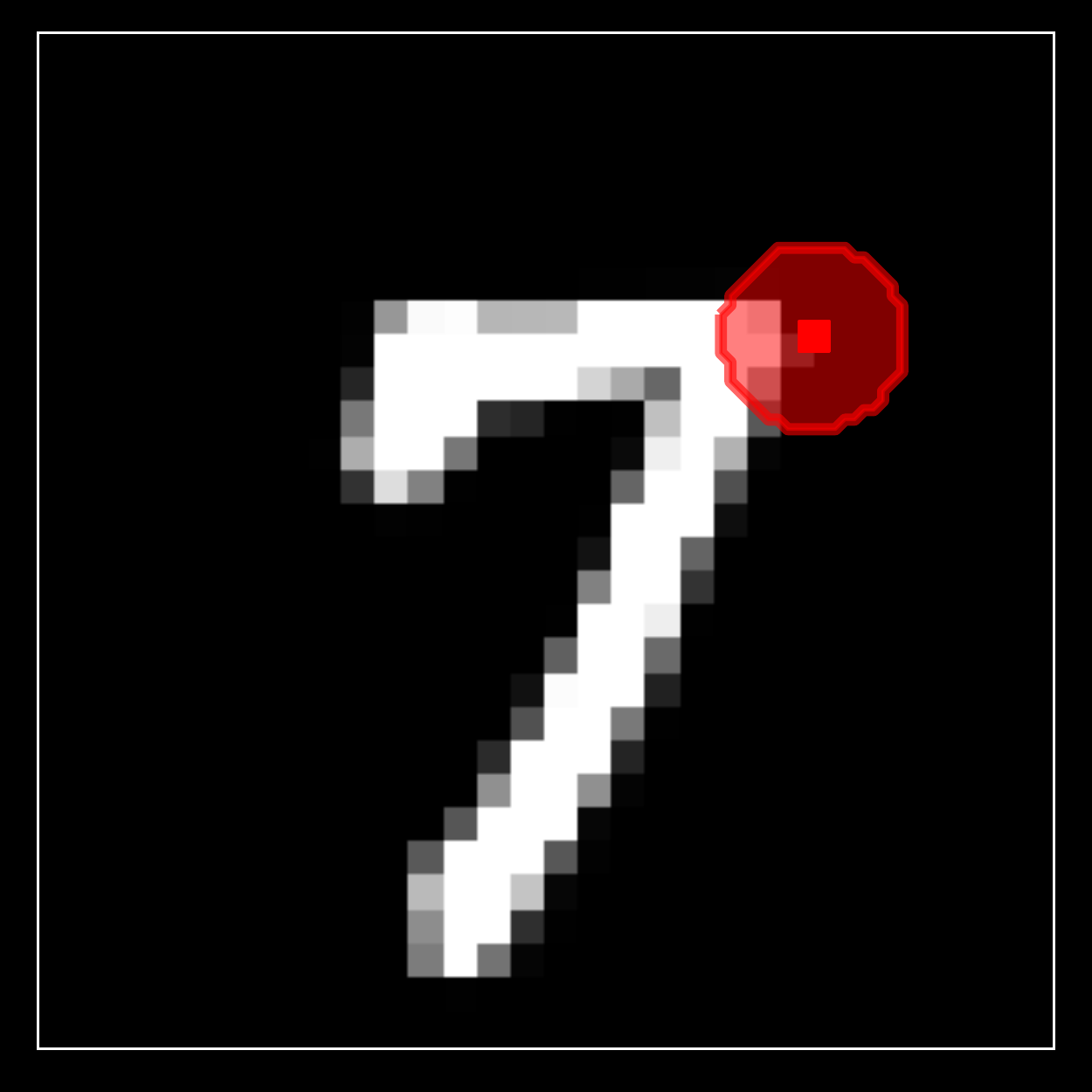}
        &
        \includegraphics[trim=10 10 10 10,clip,width=\panelwidth]{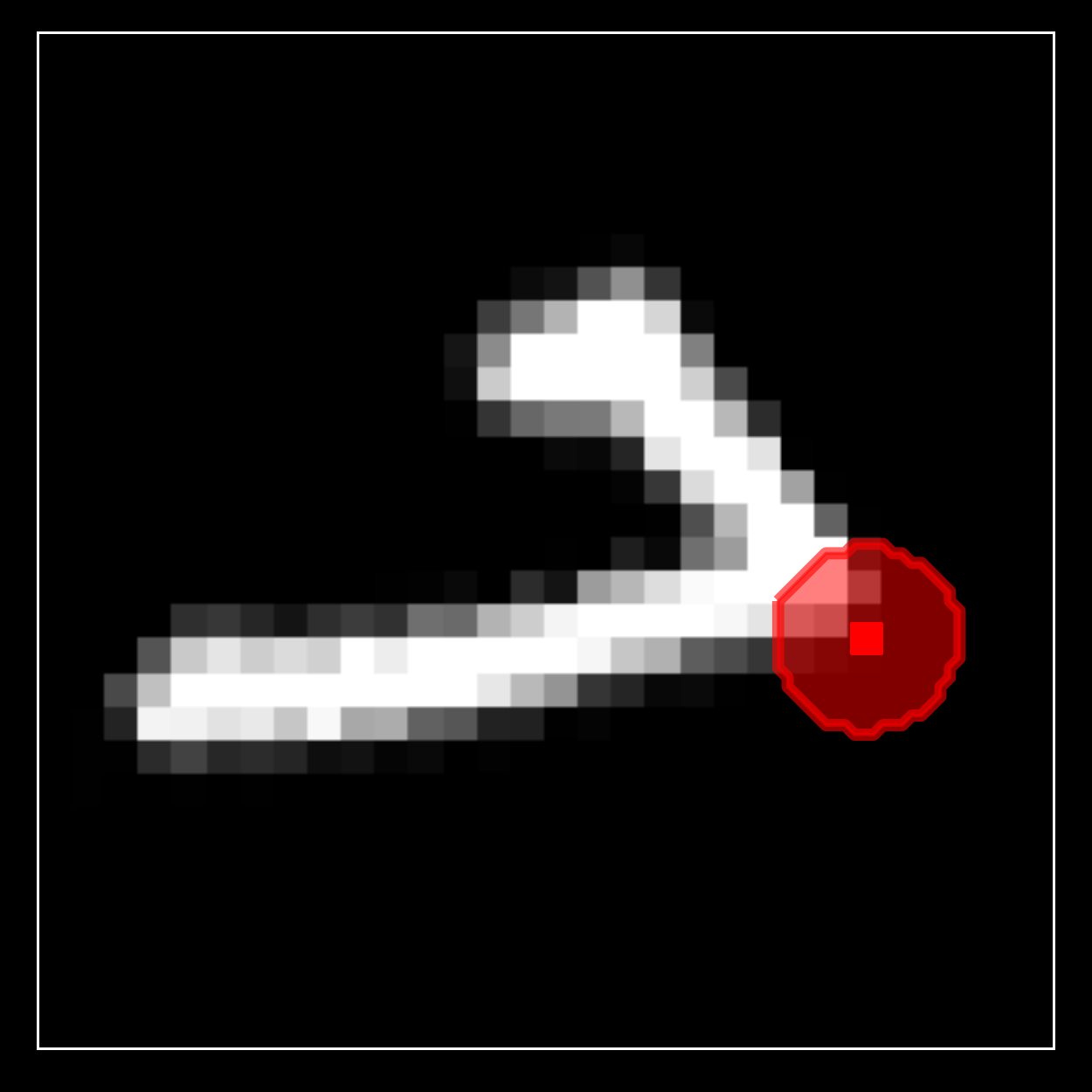}
        \\
        \includegraphics[trim=10 10 10 10,clip,width=\panelwidth]{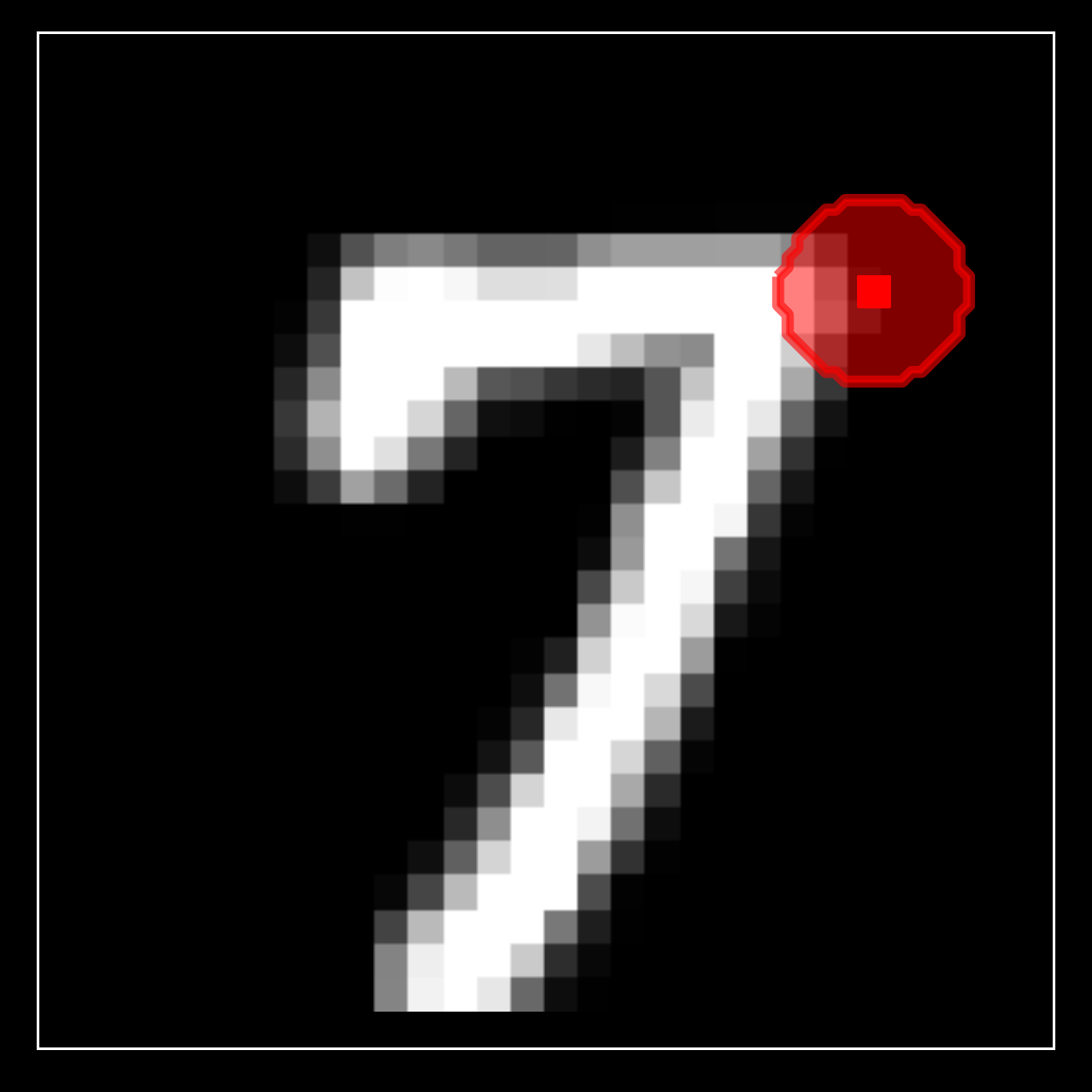}
        &
        \includegraphics[trim=10 10 10 10,clip,width=\panelwidth]{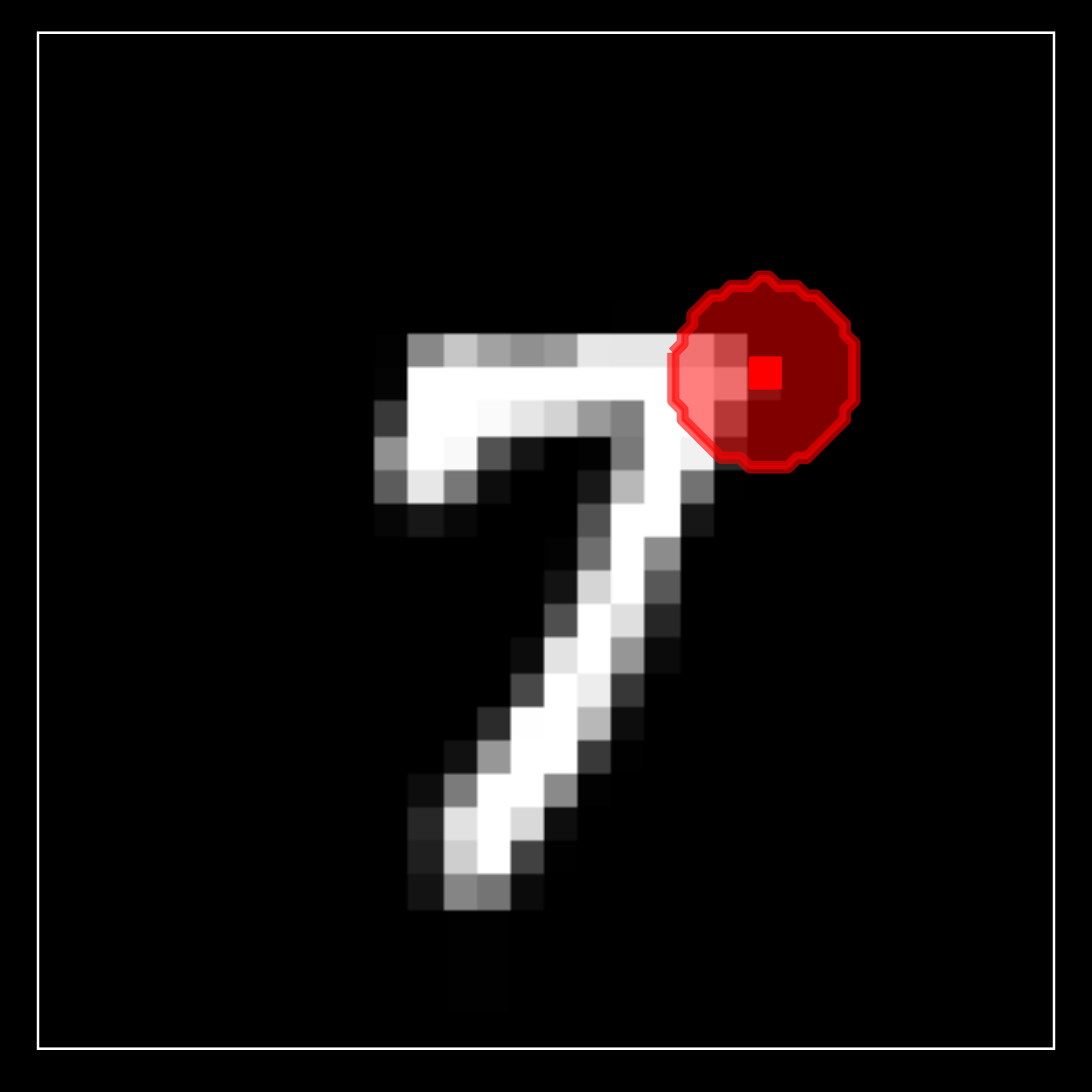}
        &
        \includegraphics[trim=10 10 10 10,clip,width=\panelwidth]{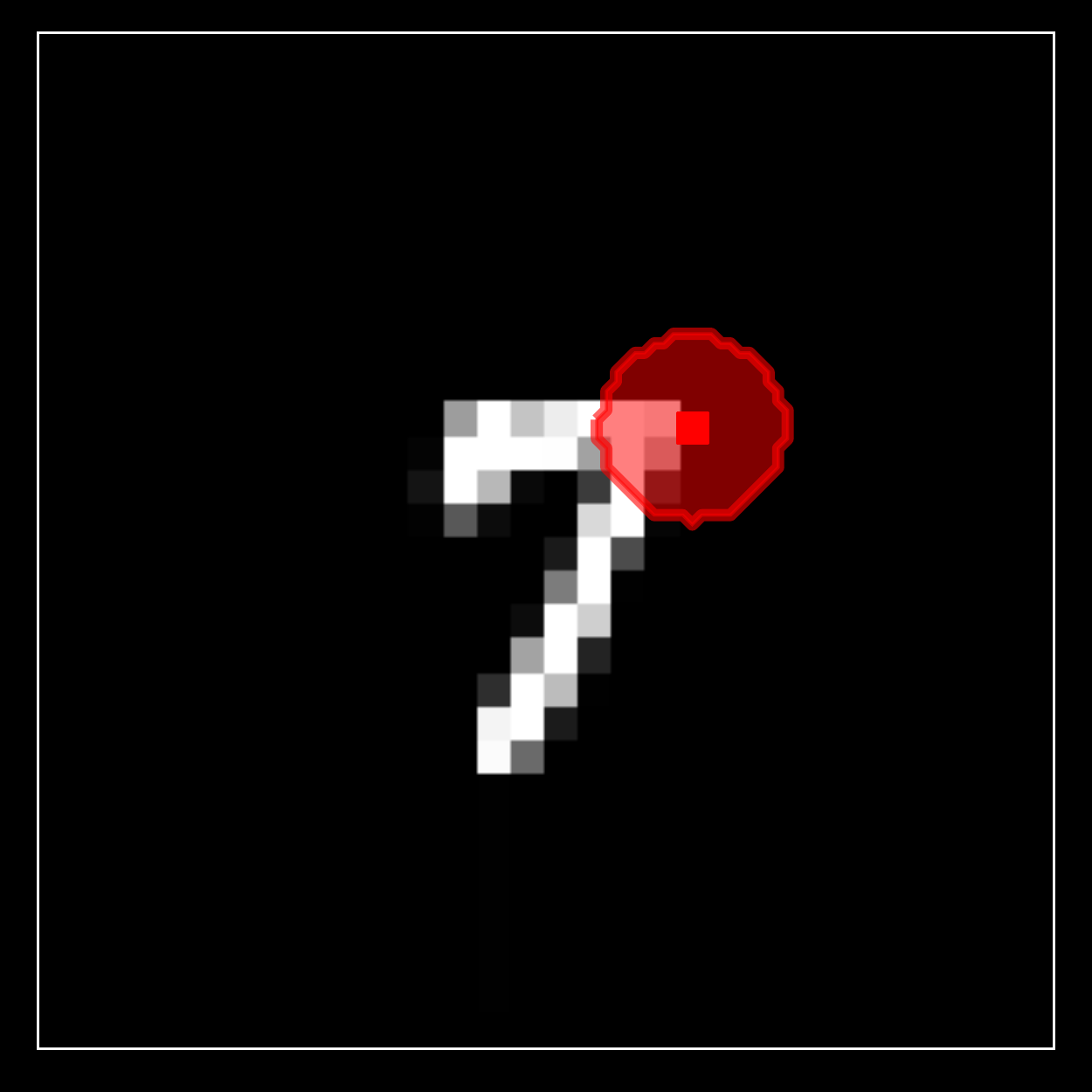}
        \\
    \end{tabular}
    }
    \subfigure[$SO(2)$]{
    \begin{tabular}{ccc}
        \includegraphics[trim=10 10 10 10,clip,width=\panelwidth]{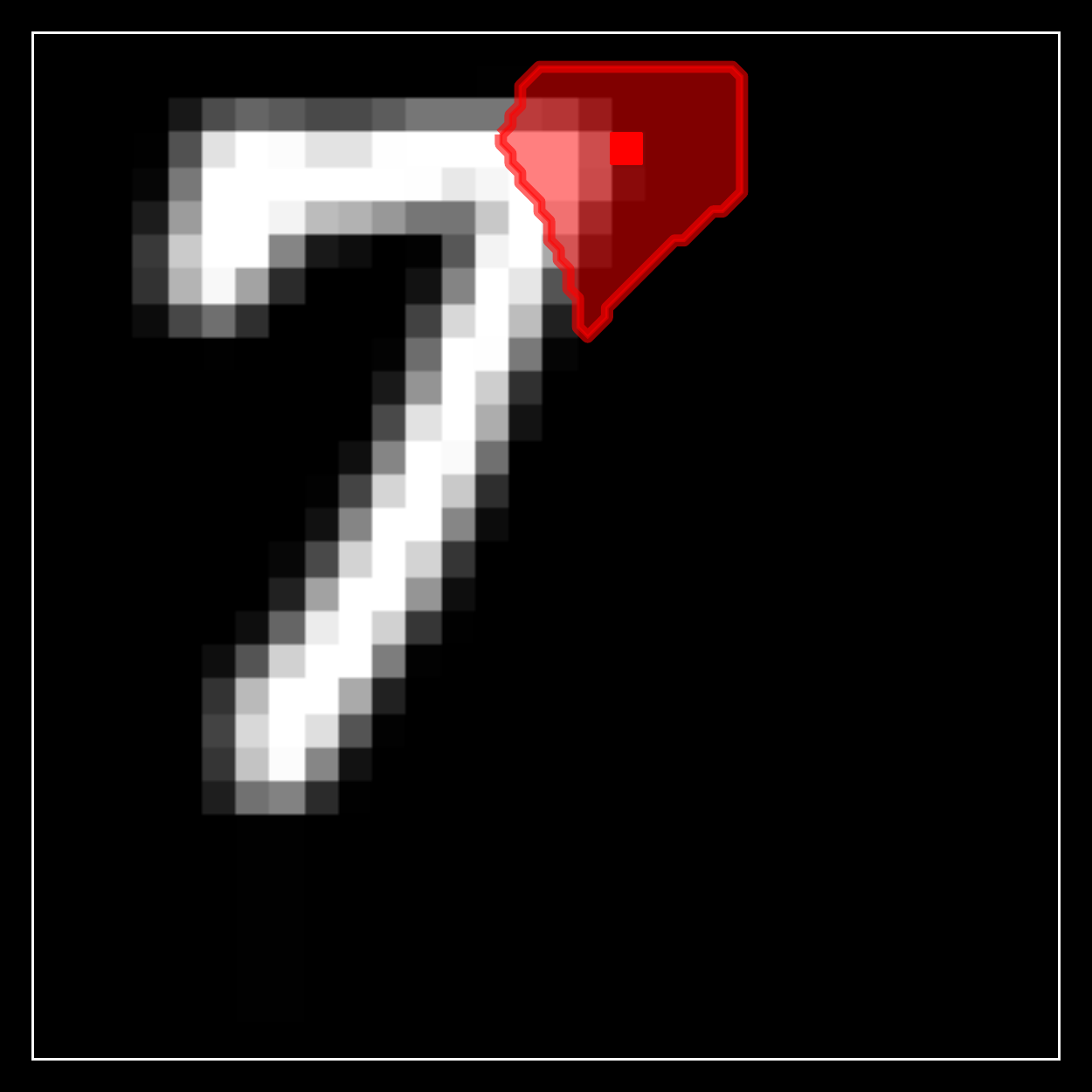}
        &
        \includegraphics[trim=10 10 10 10,clip,width=\panelwidth]{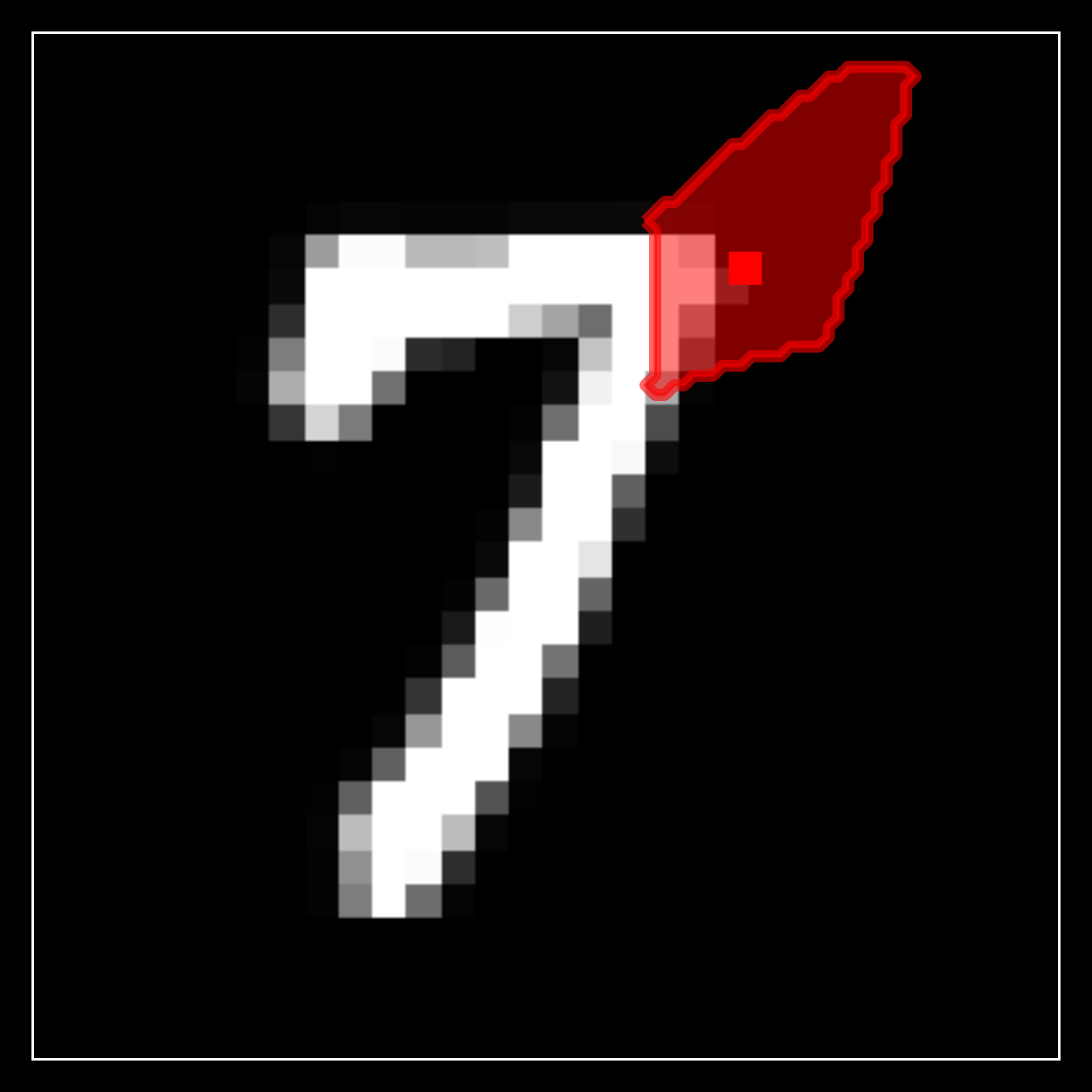}
        &
        \includegraphics[trim=10 10 10 10,clip,width=\panelwidth]{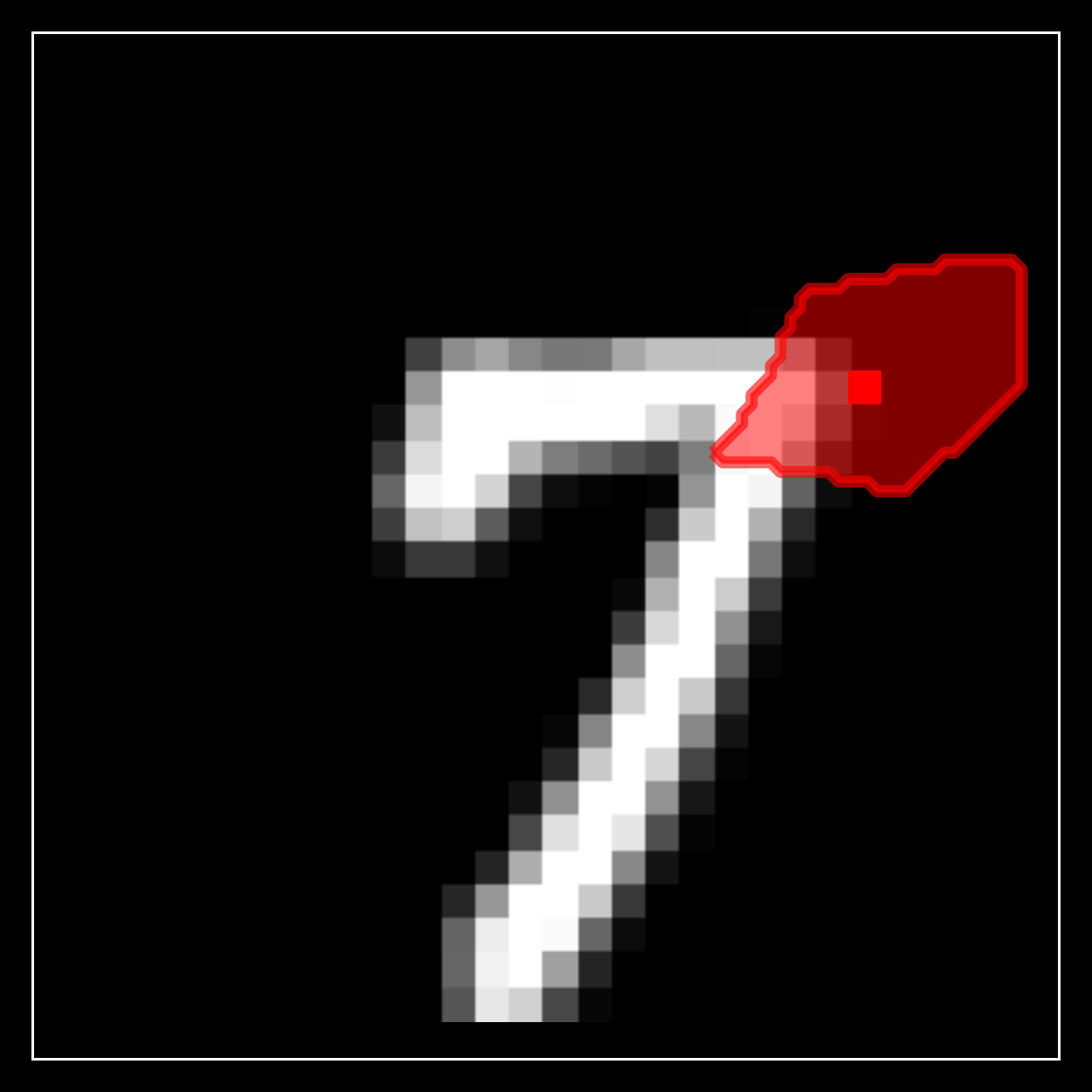}
        \\
        \includegraphics[trim=10 10 10 10,clip,width=\panelwidth]{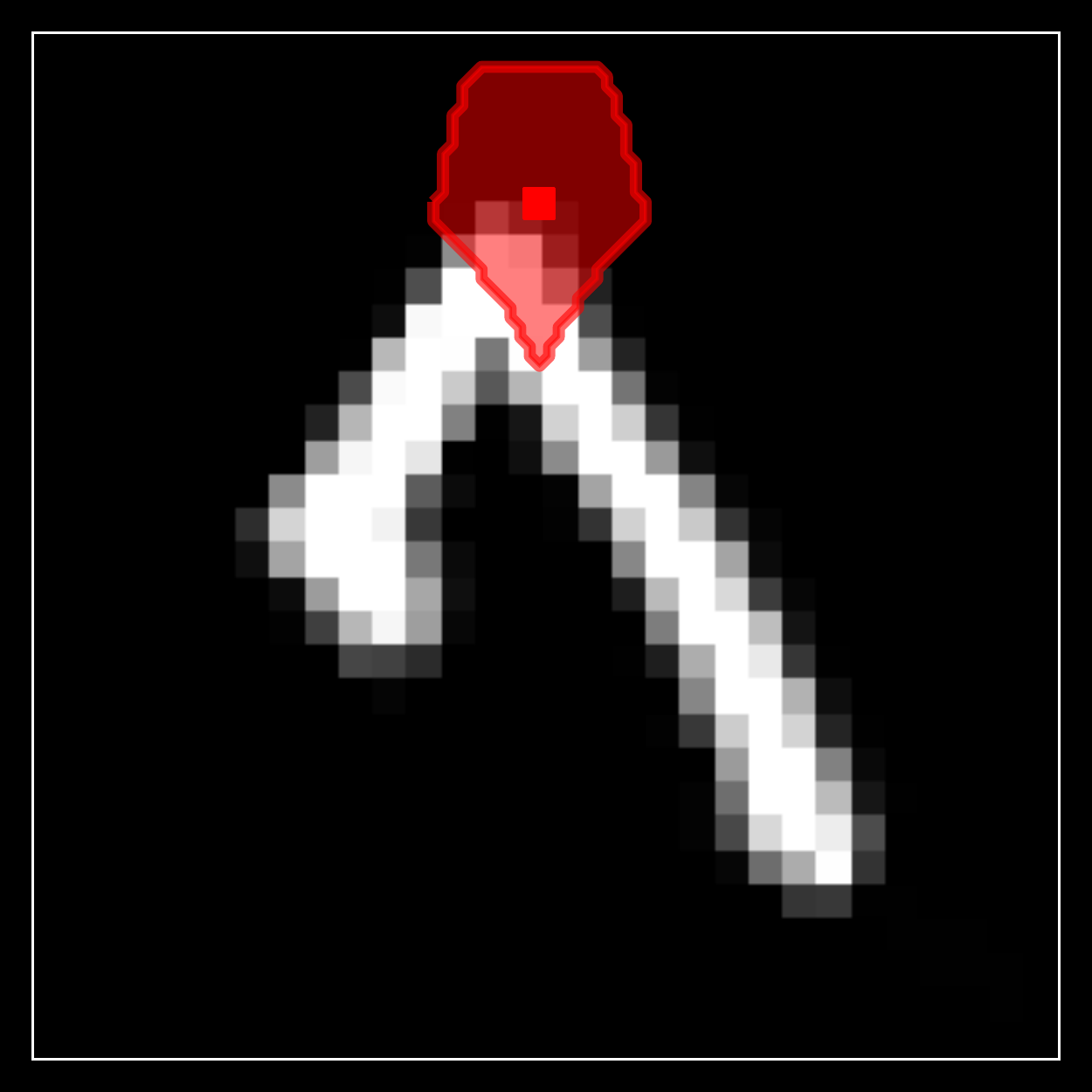}
        &
        \includegraphics[trim=10 10 10 10,clip,width=\panelwidth]{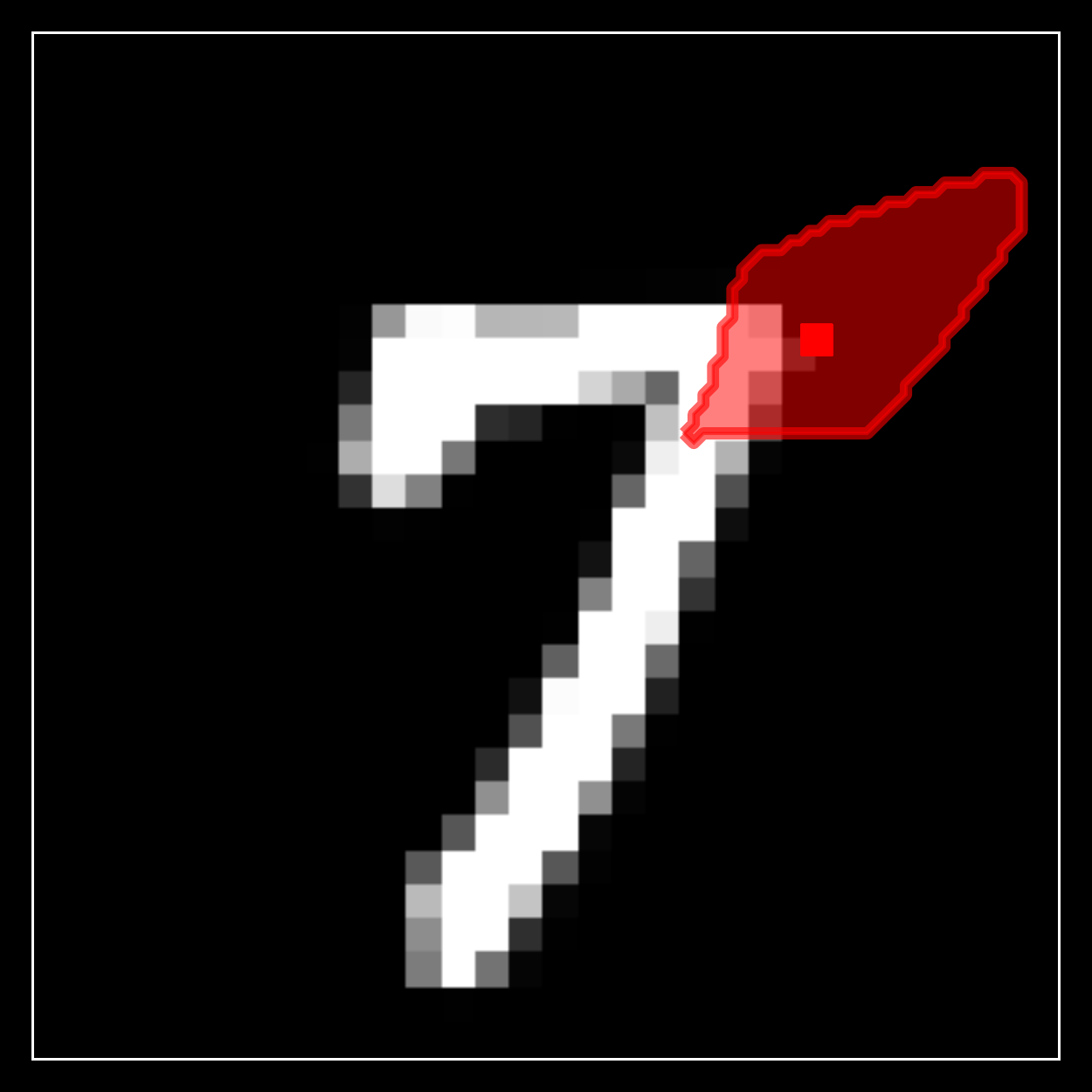}
        &
        \includegraphics[trim=10 10 10 10,clip,width=\panelwidth]{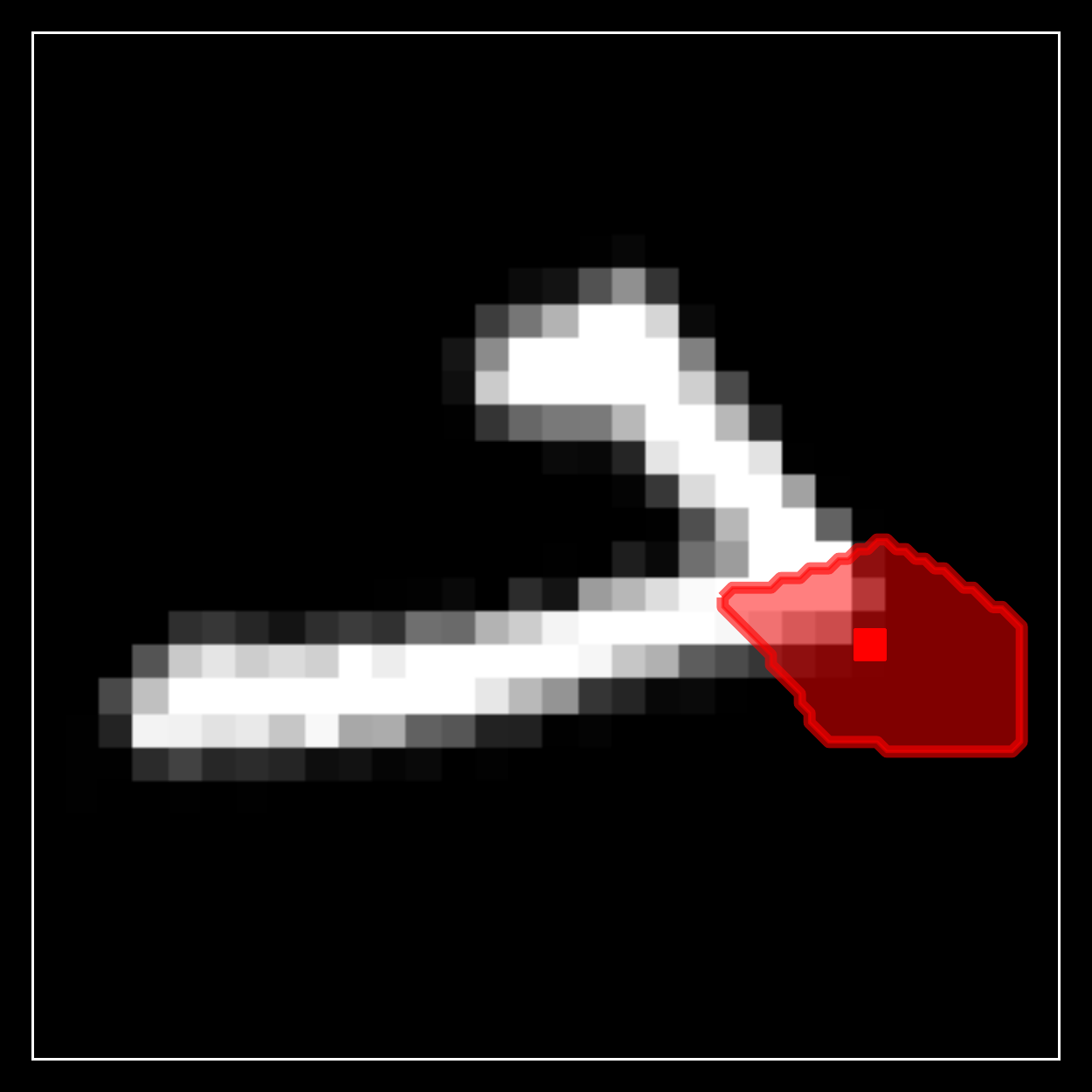}
        \\
        \includegraphics[trim=10 10 10 10,clip,width=\panelwidth]{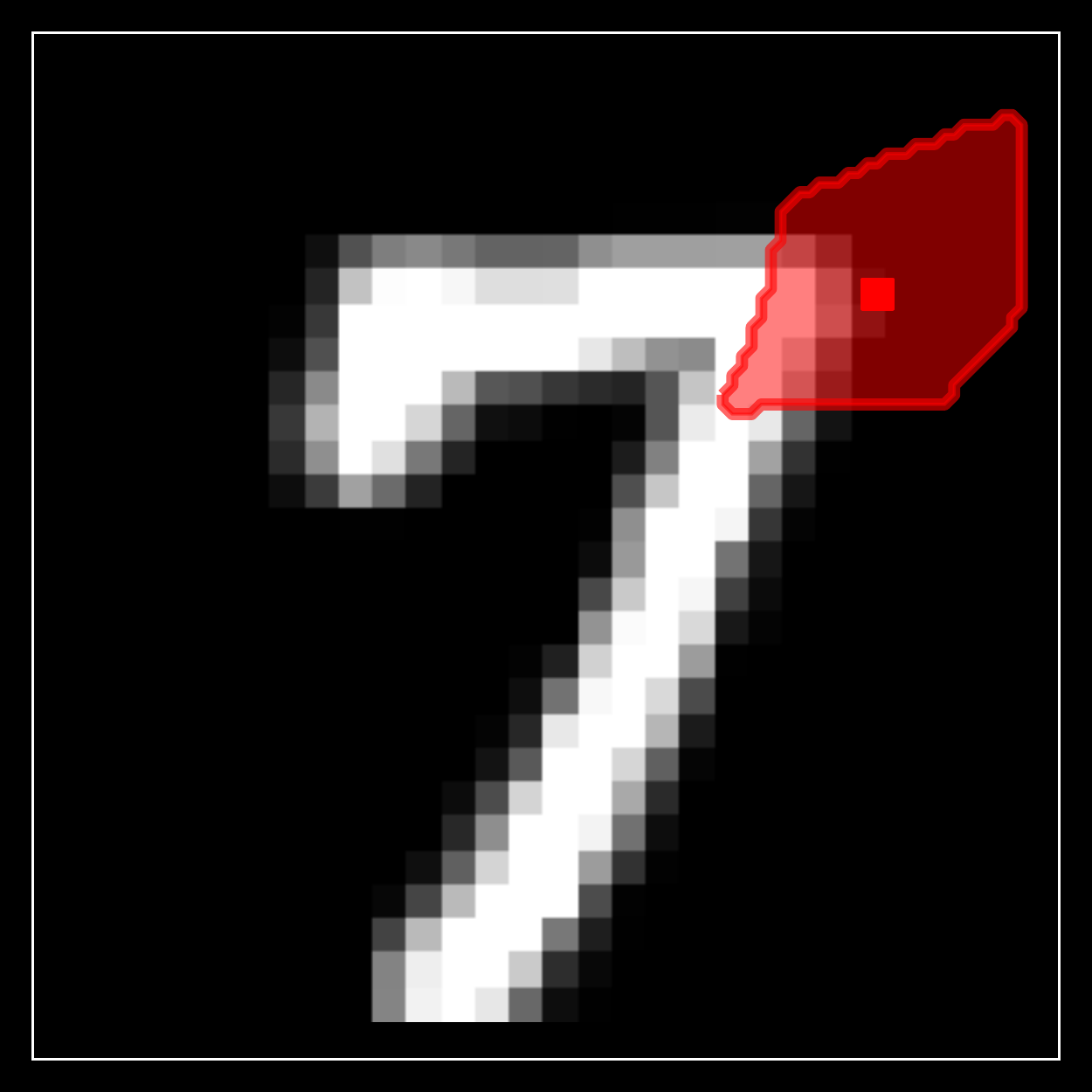}
        &
        \includegraphics[trim=10 10 10 10,clip,width=\panelwidth]{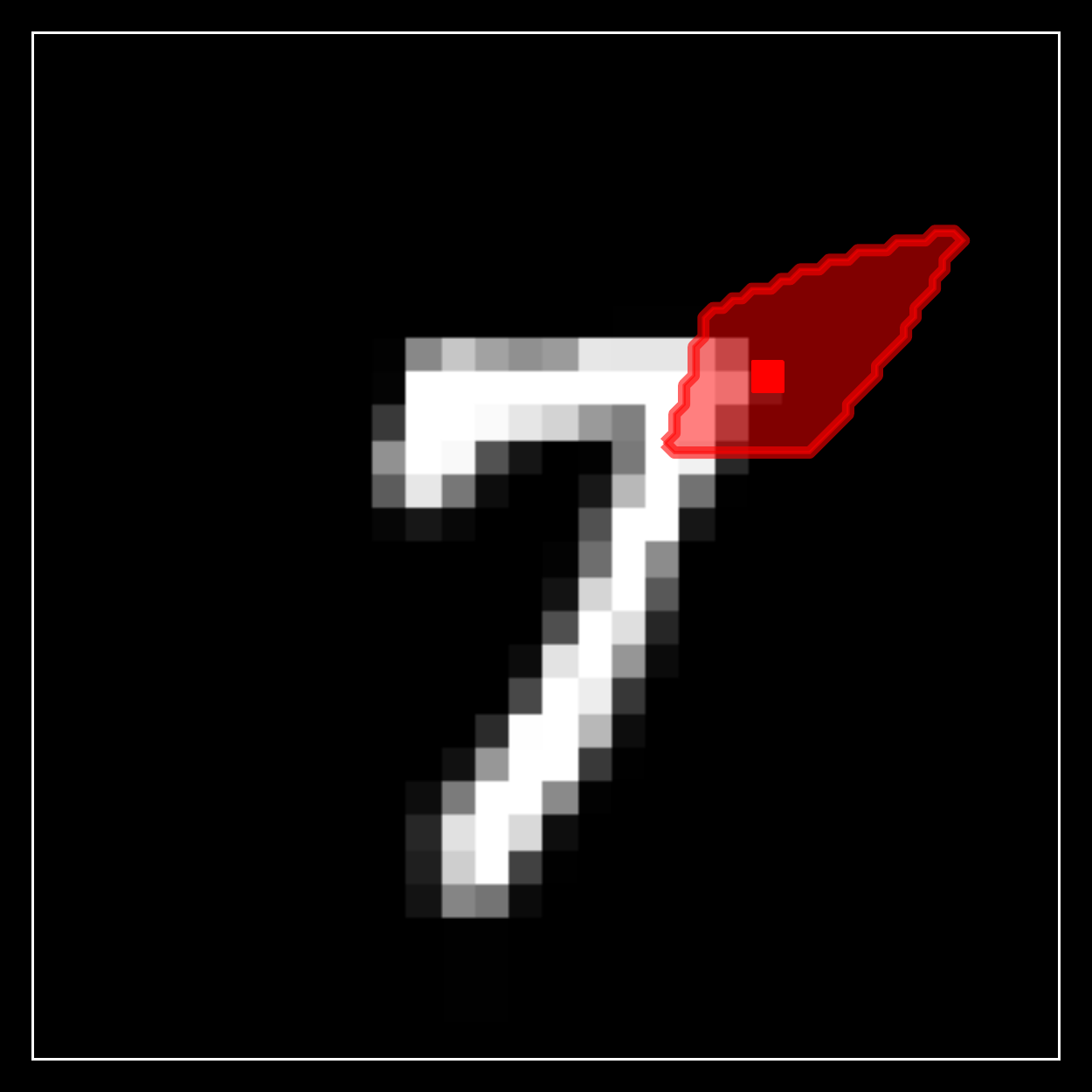}
        &
        \includegraphics[trim=10 10 10 10,clip,width=\panelwidth]{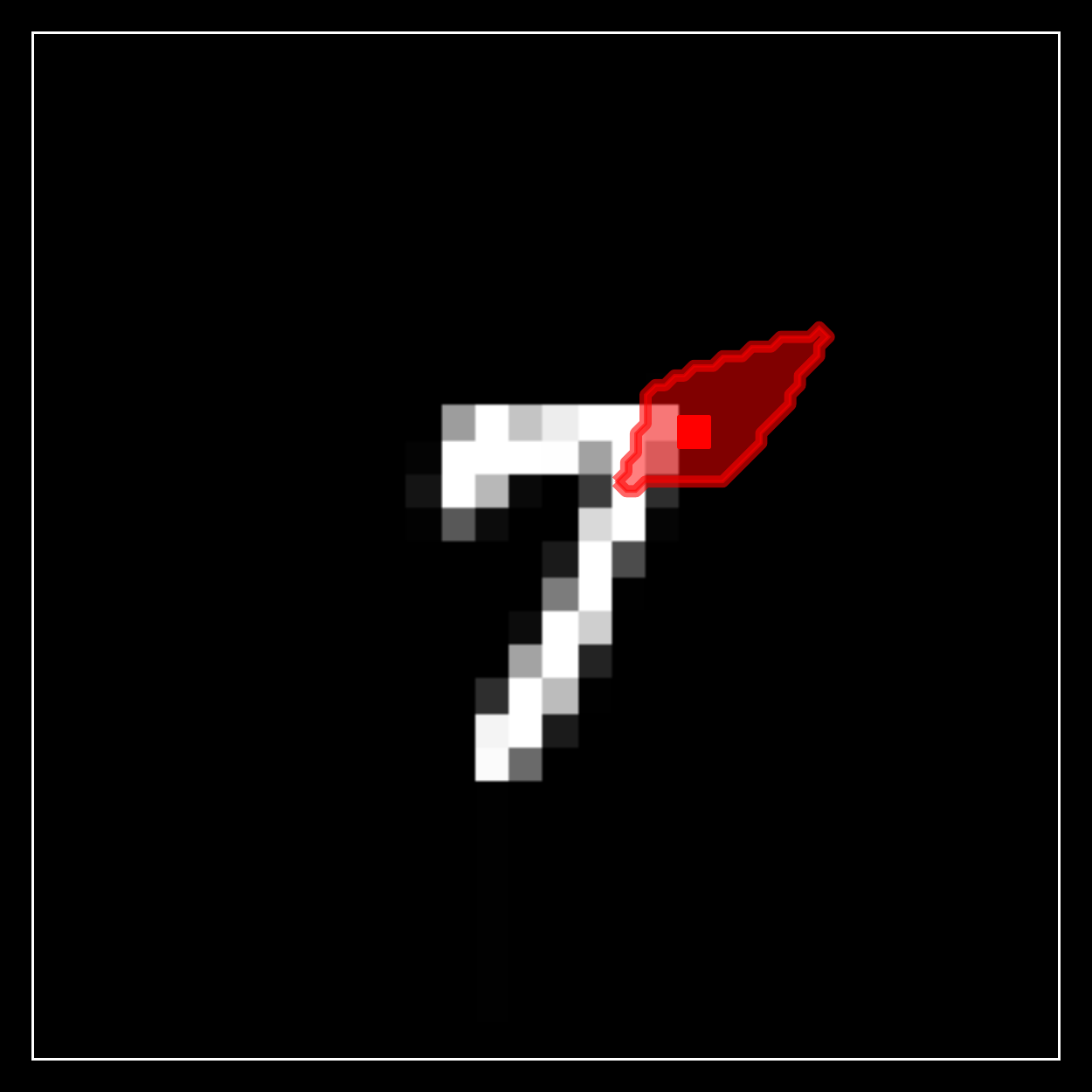}
        \\
    \end{tabular}
    }
    \subfigure[$\mathbb R^*\times SO(2)$]{
    \begin{tabular}{ccc}
        \includegraphics[trim=10 10 10 10,clip,width=\panelwidth]{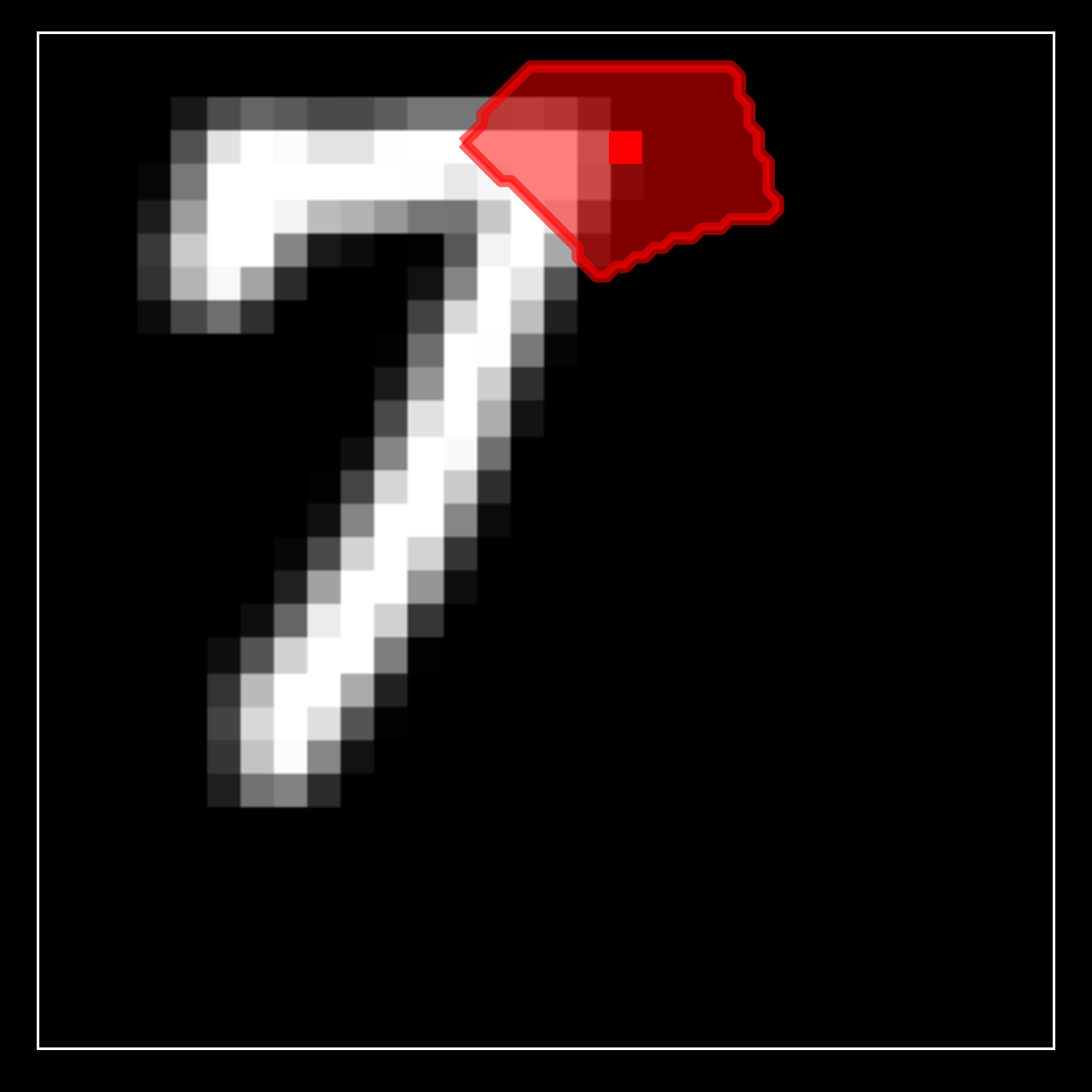}
        &
        \includegraphics[trim=10 10 10 10,clip,width=\panelwidth]{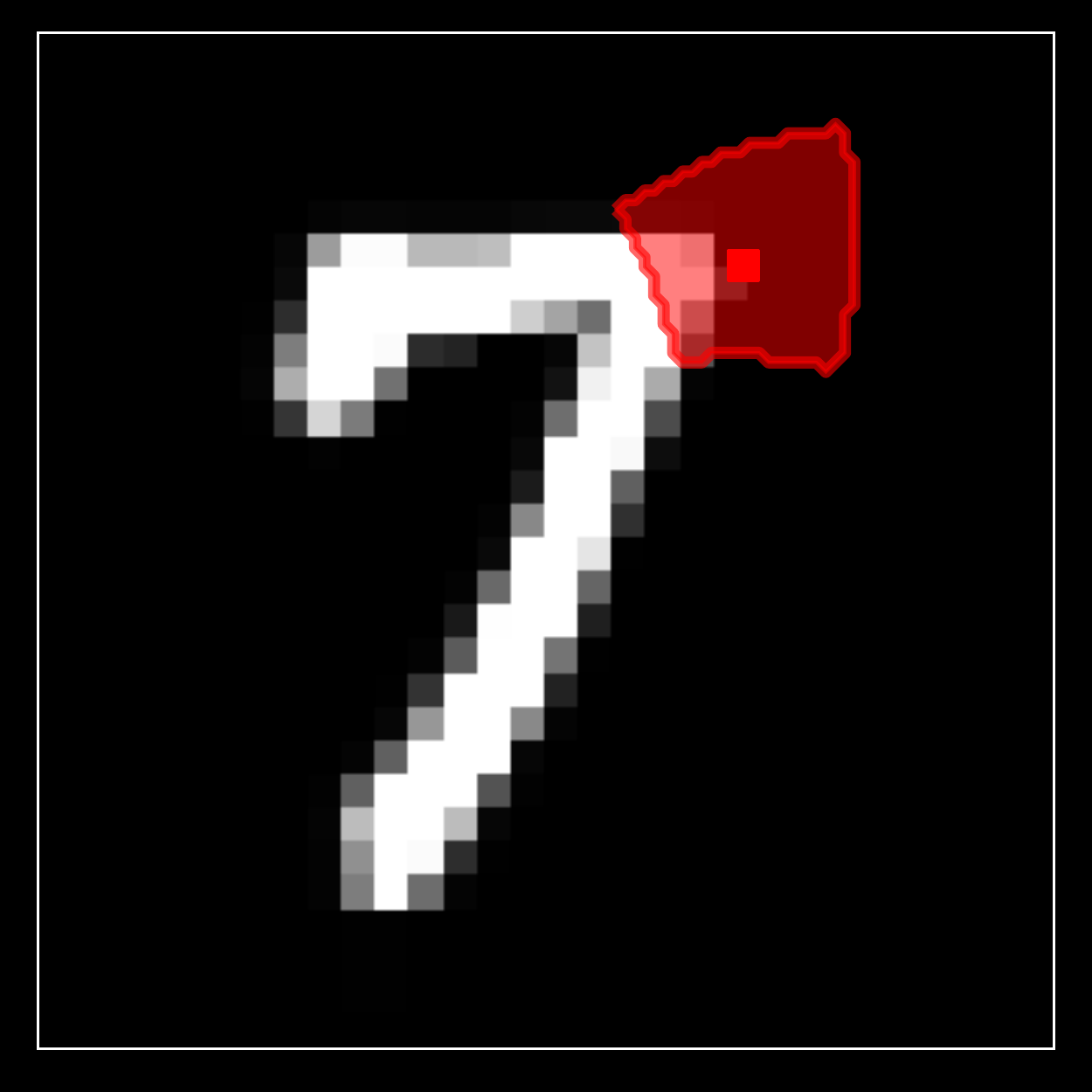}
        &
        \includegraphics[trim=10 10 10 10,clip,width=\panelwidth]{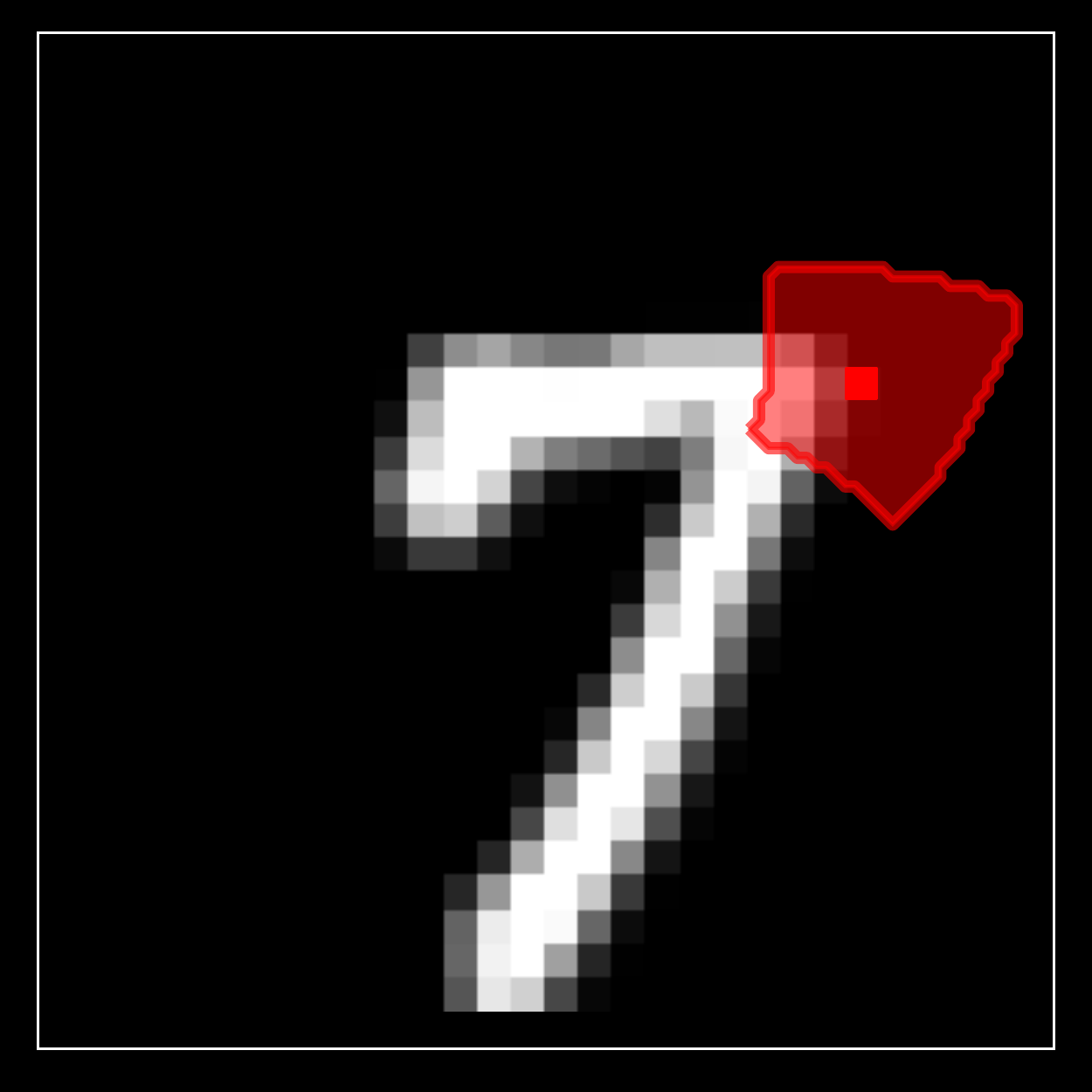}
        \\
        \includegraphics[trim=10 10 10 10,clip,width=\panelwidth]{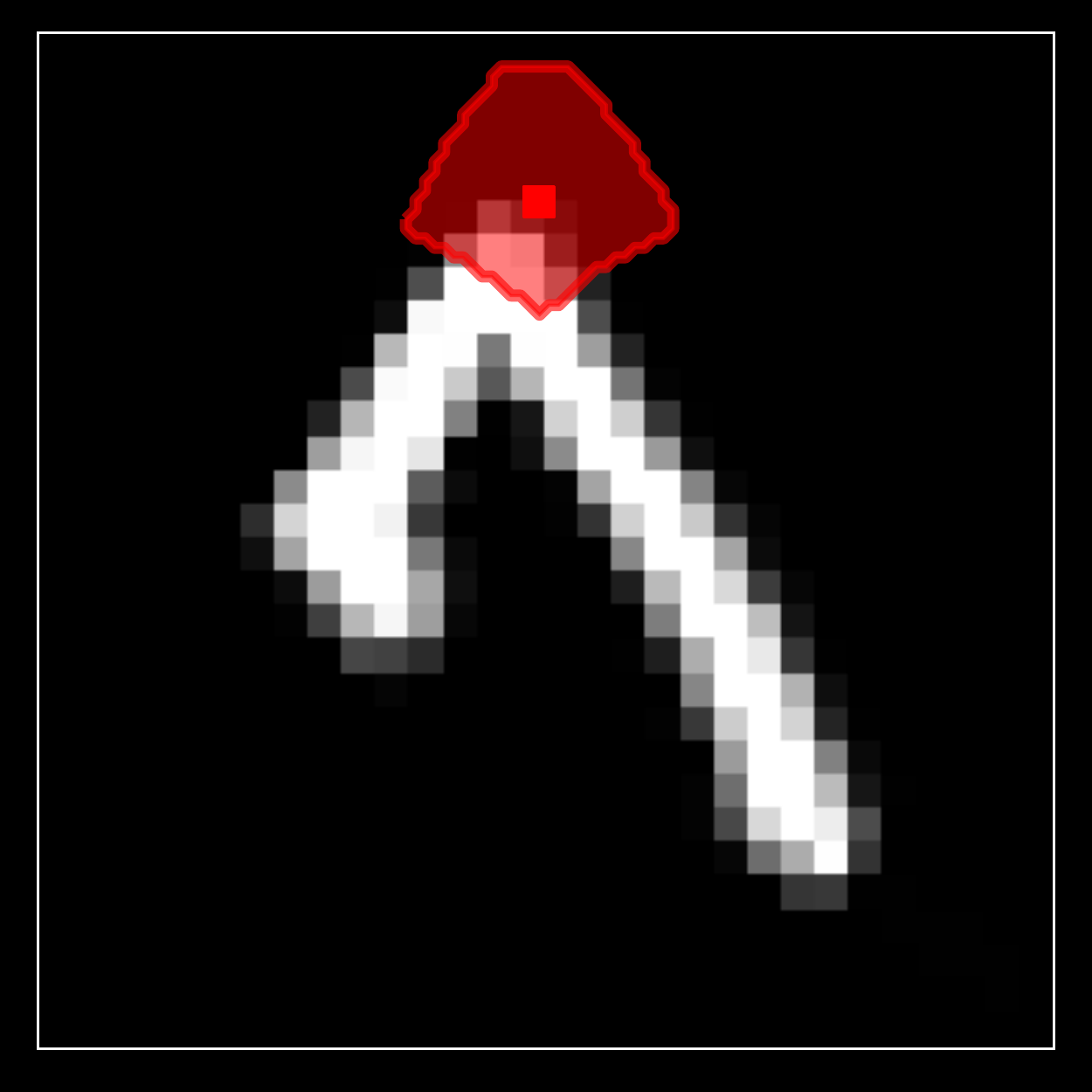}
        &
        \includegraphics[trim=10 10 10 10,clip,width=\panelwidth]{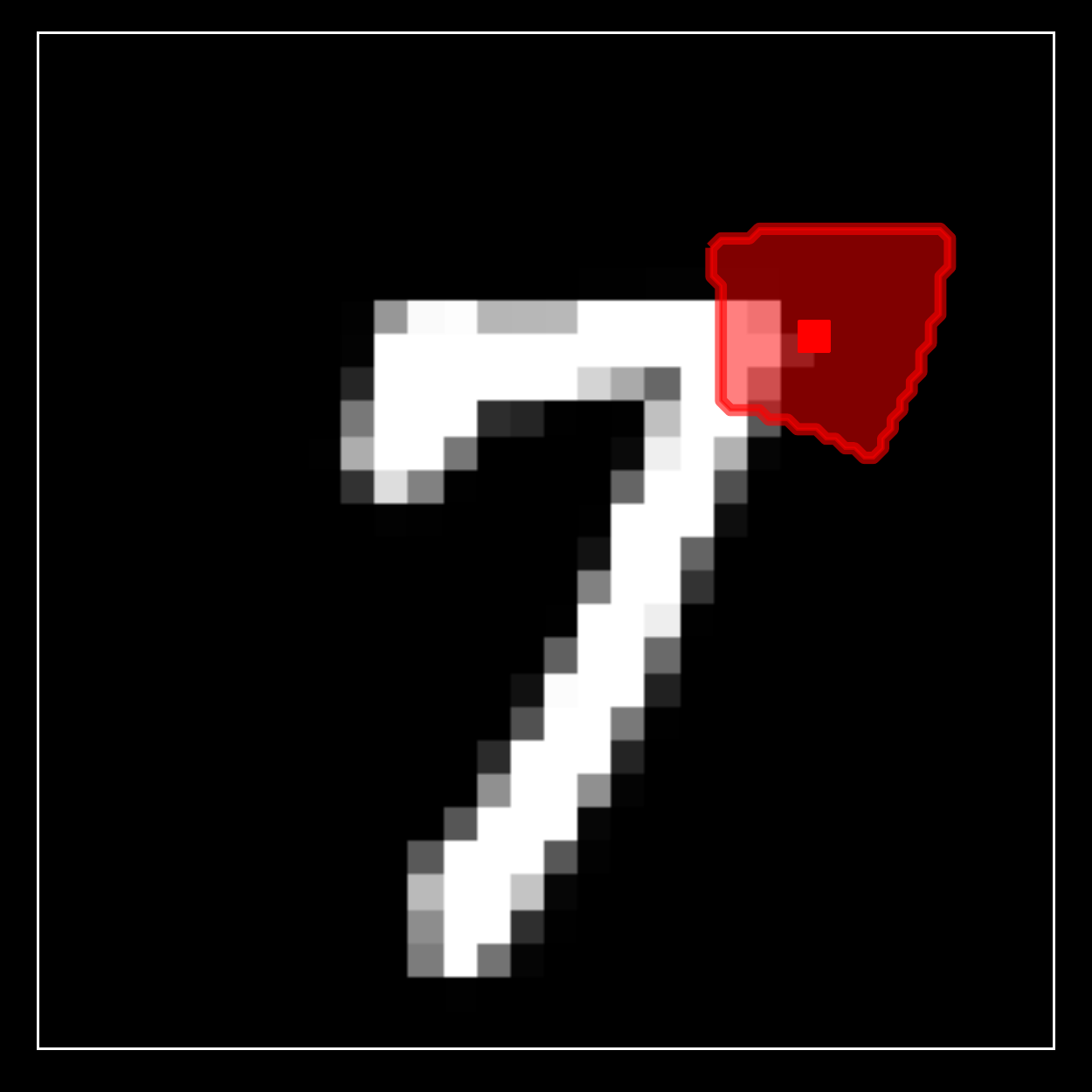}
        &
        \includegraphics[trim=10 10 10 10,clip,width=\panelwidth]{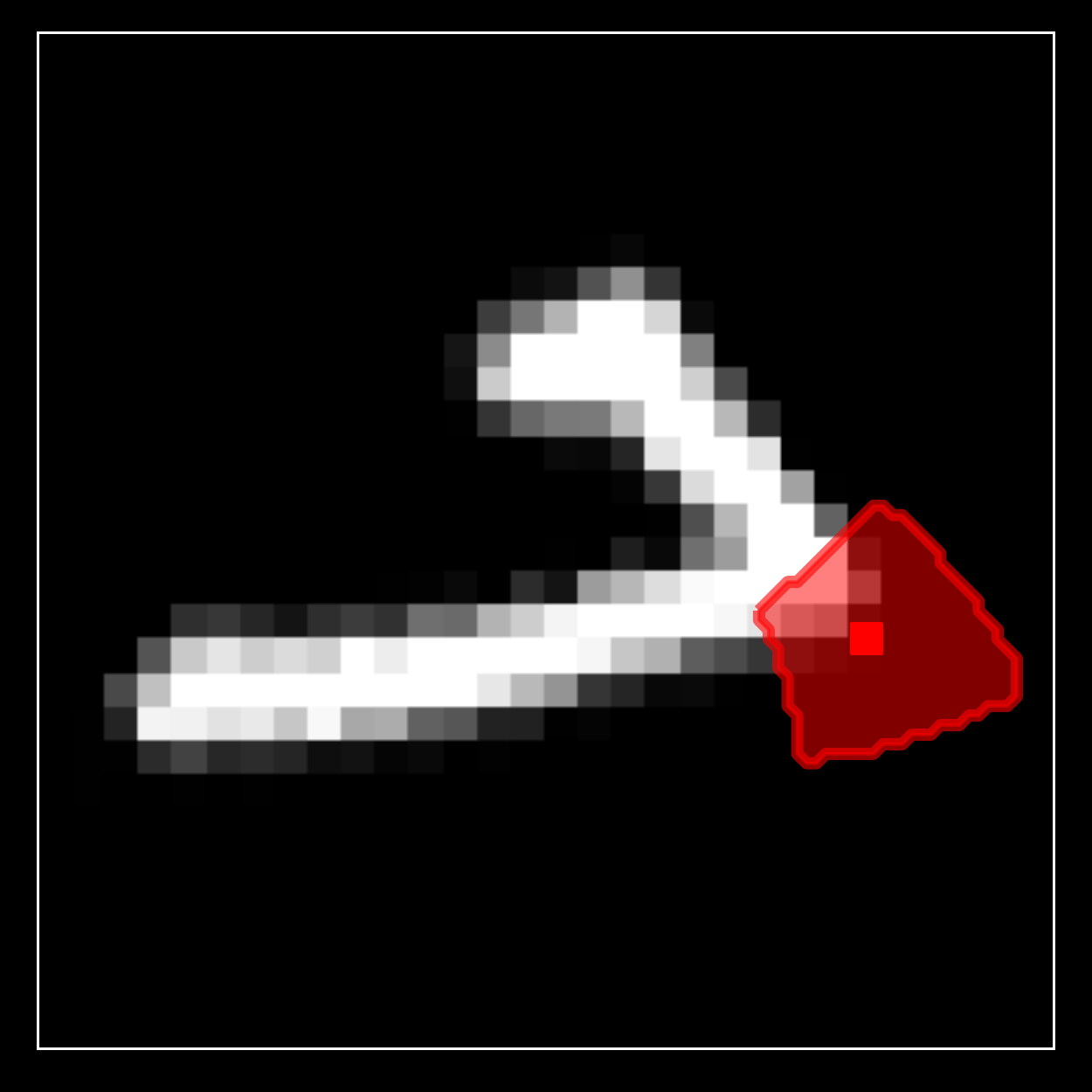}
        \\
        \includegraphics[trim=10 10 10 10,clip,width=\panelwidth]{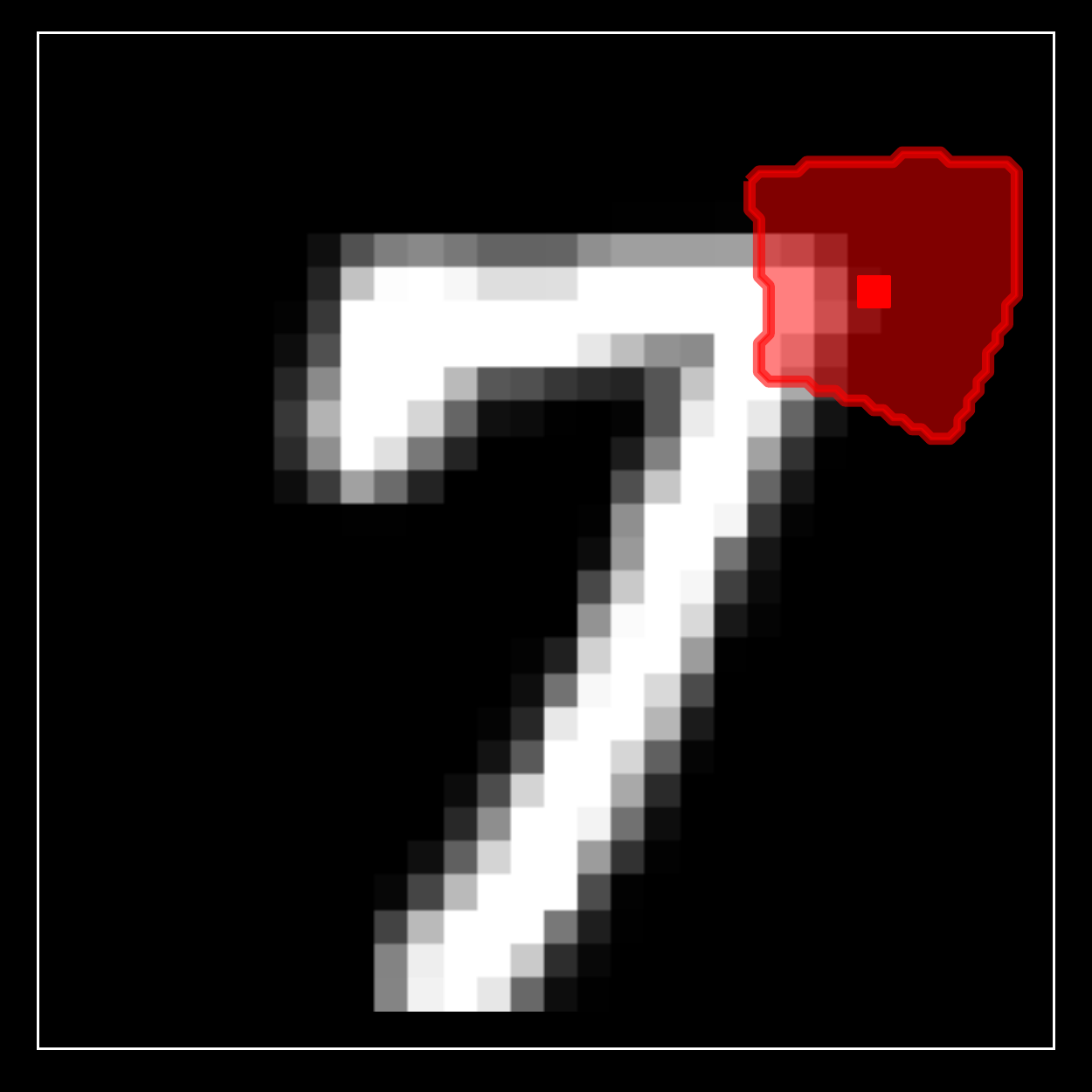}
        &
        \includegraphics[trim=10 10 10 10,clip,width=\panelwidth]{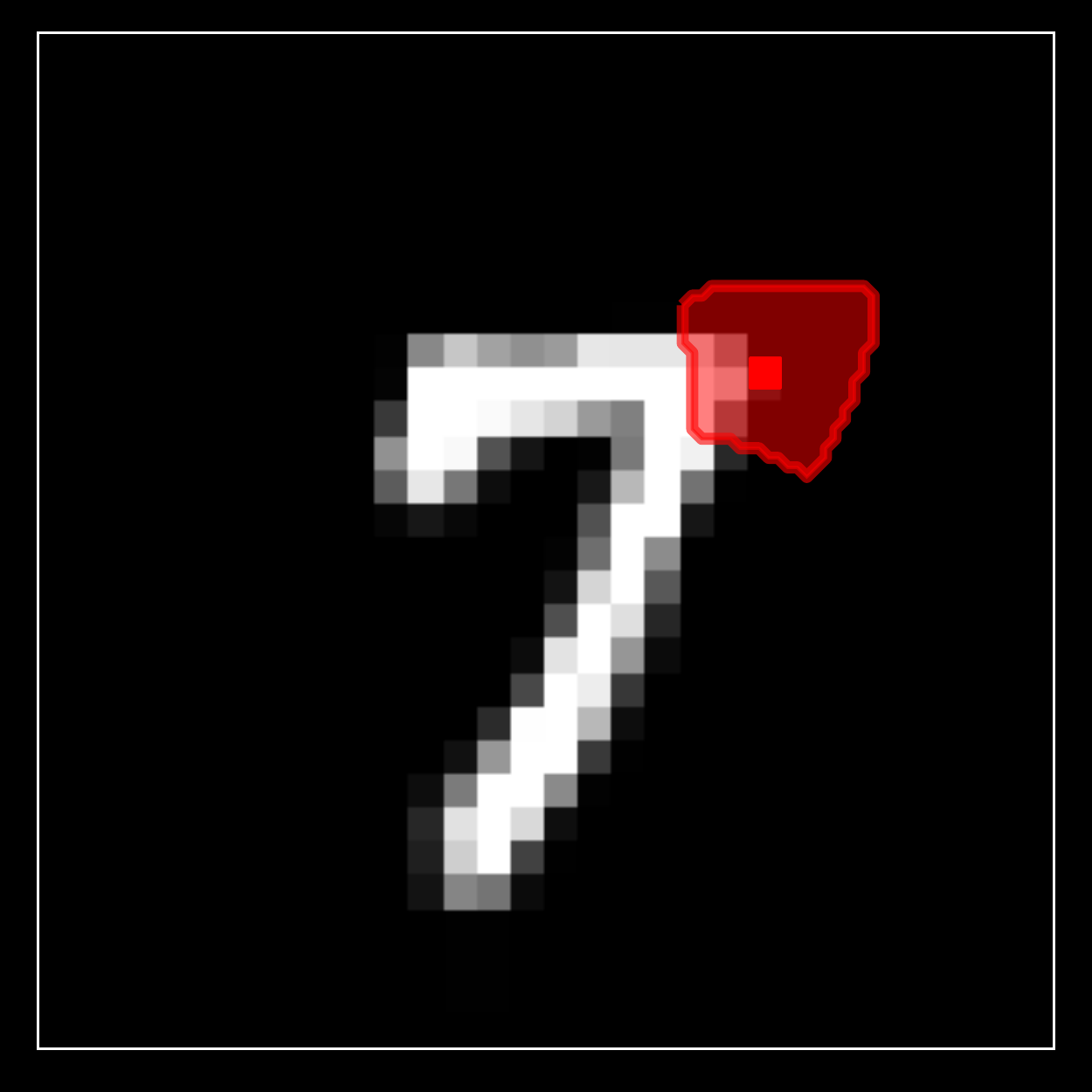}
        &
        \includegraphics[trim=10 10 10 10,clip,width=\panelwidth]{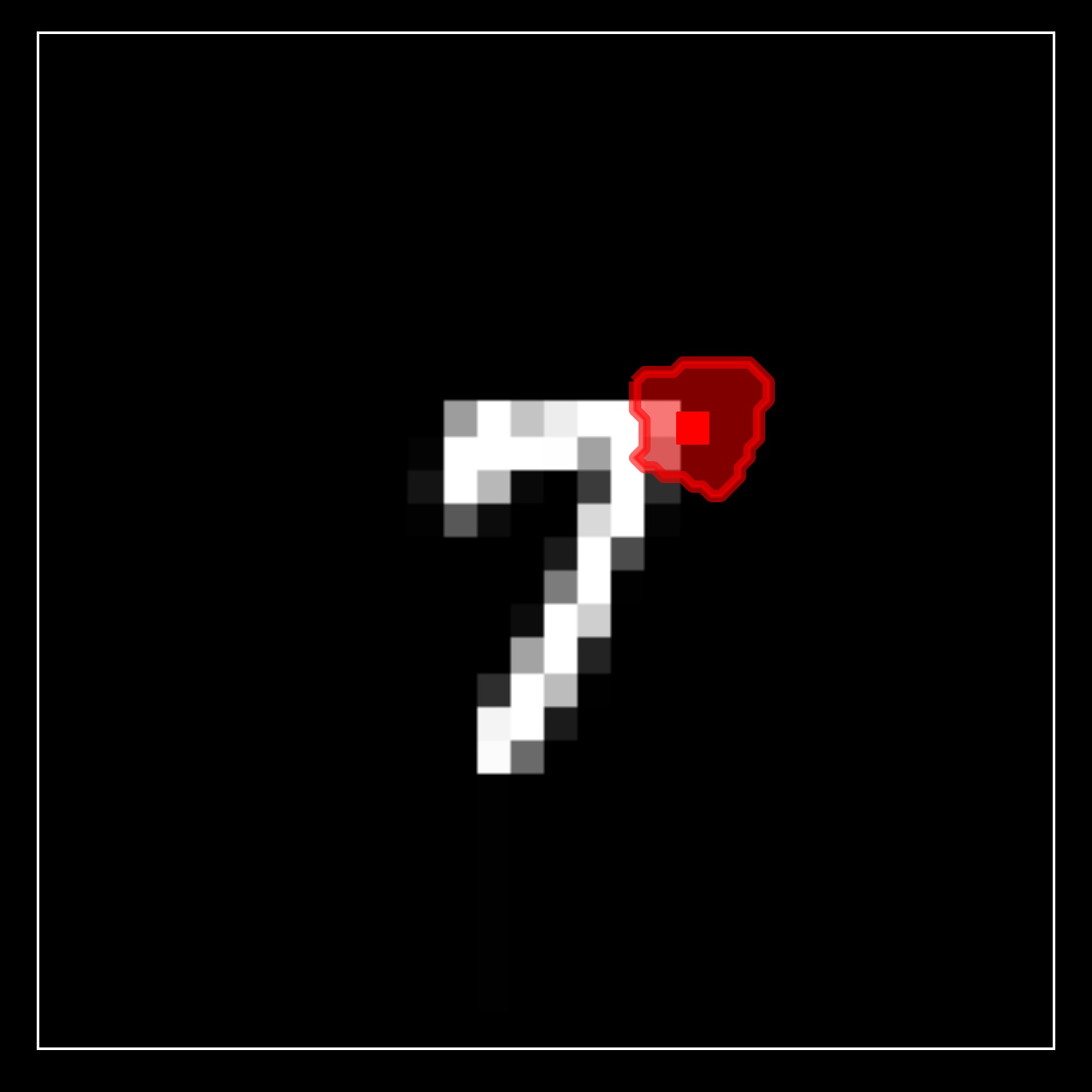}
        \\
    \end{tabular}
    }
    \subfigure[$SE(2)$]{
    \begin{tabular}{ccc}
        \includegraphics[trim=10 10 10 10,clip,width=\panelwidth]{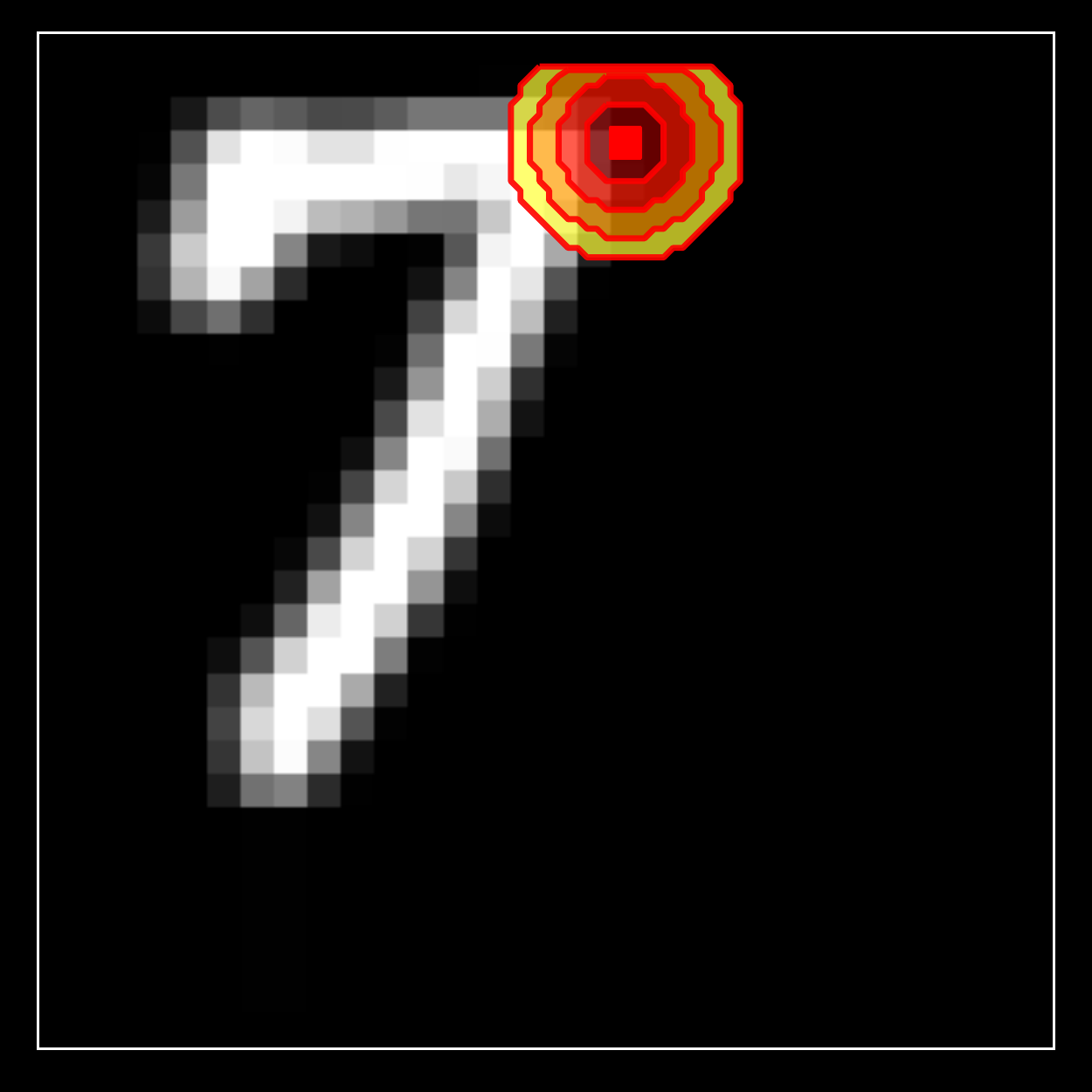}
        &
        \includegraphics[trim=10 10 10 10,clip,width=\panelwidth]{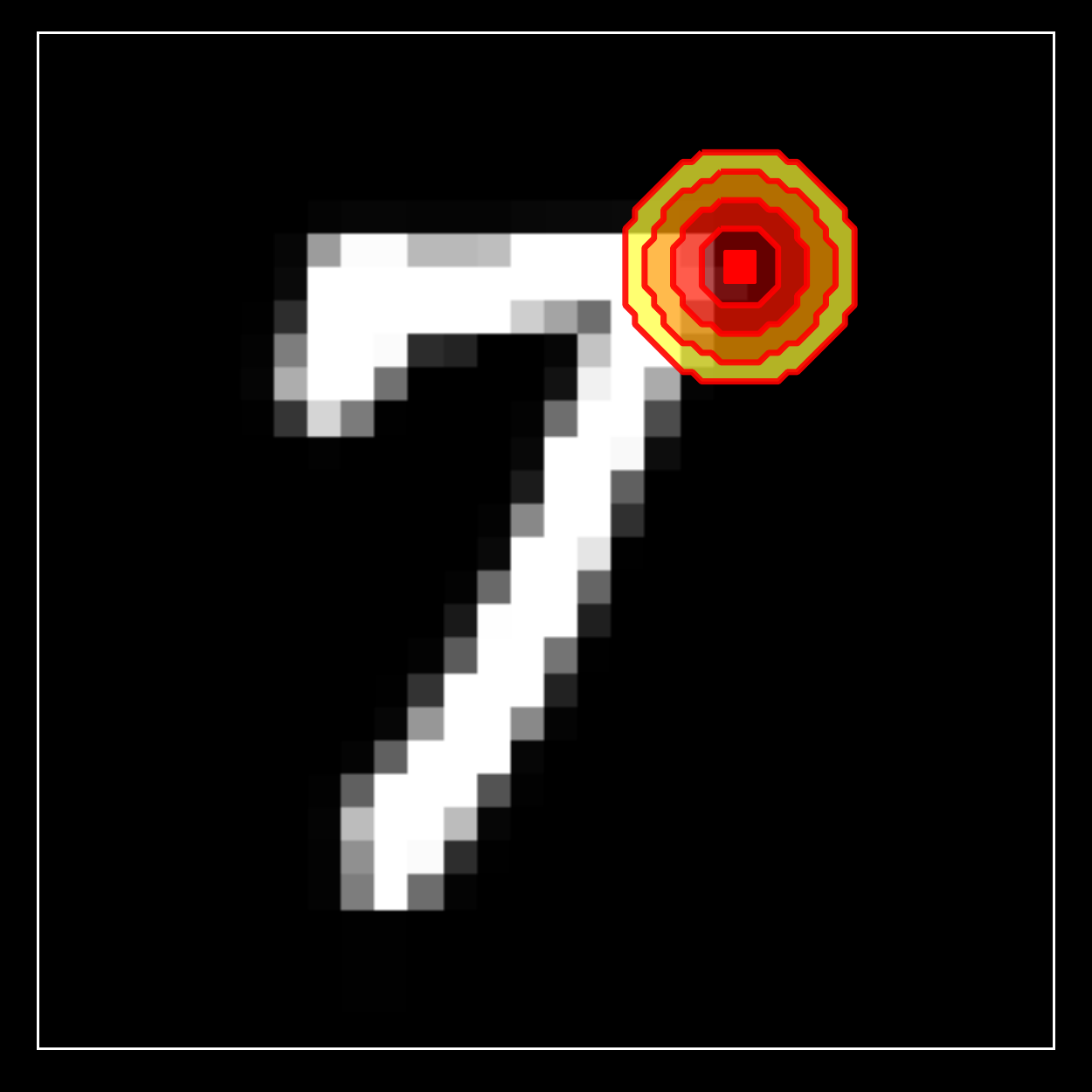}
        &
        \includegraphics[trim=10 10 10 10,clip,width=\panelwidth]{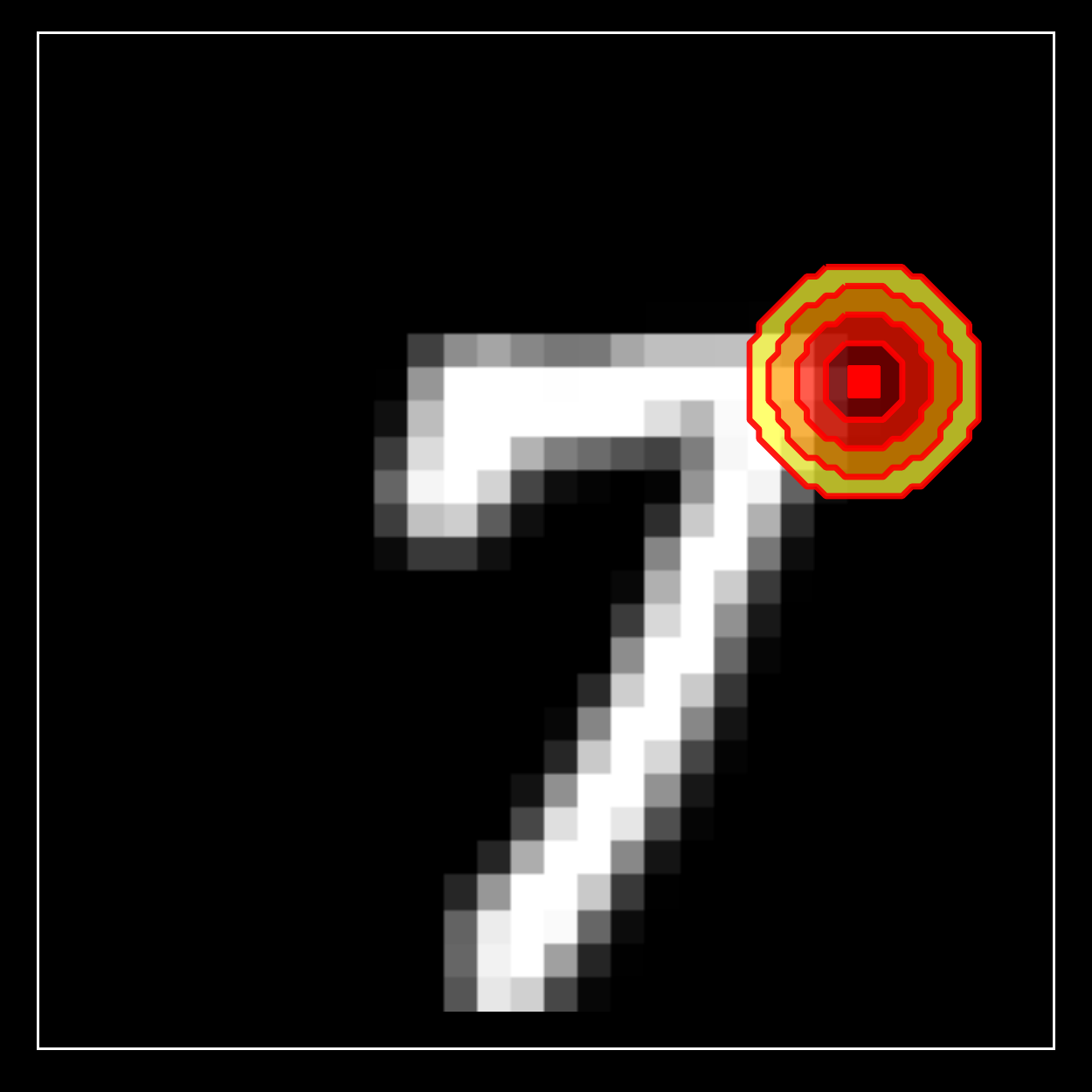}
        \\
        \includegraphics[trim=10 10 10 10,clip,width=\panelwidth]{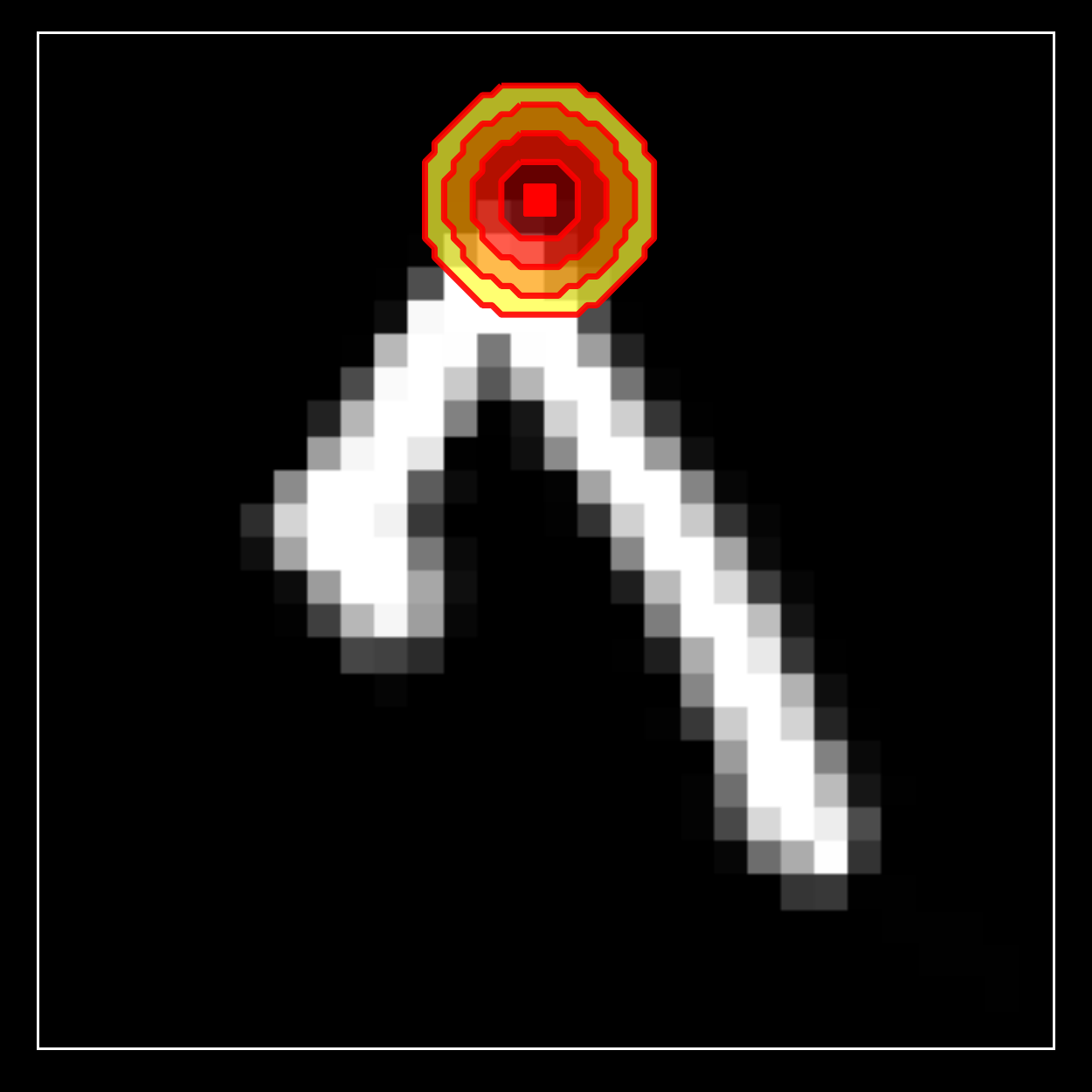}
        &
        \includegraphics[trim=10 10 10 10,clip,width=\panelwidth]{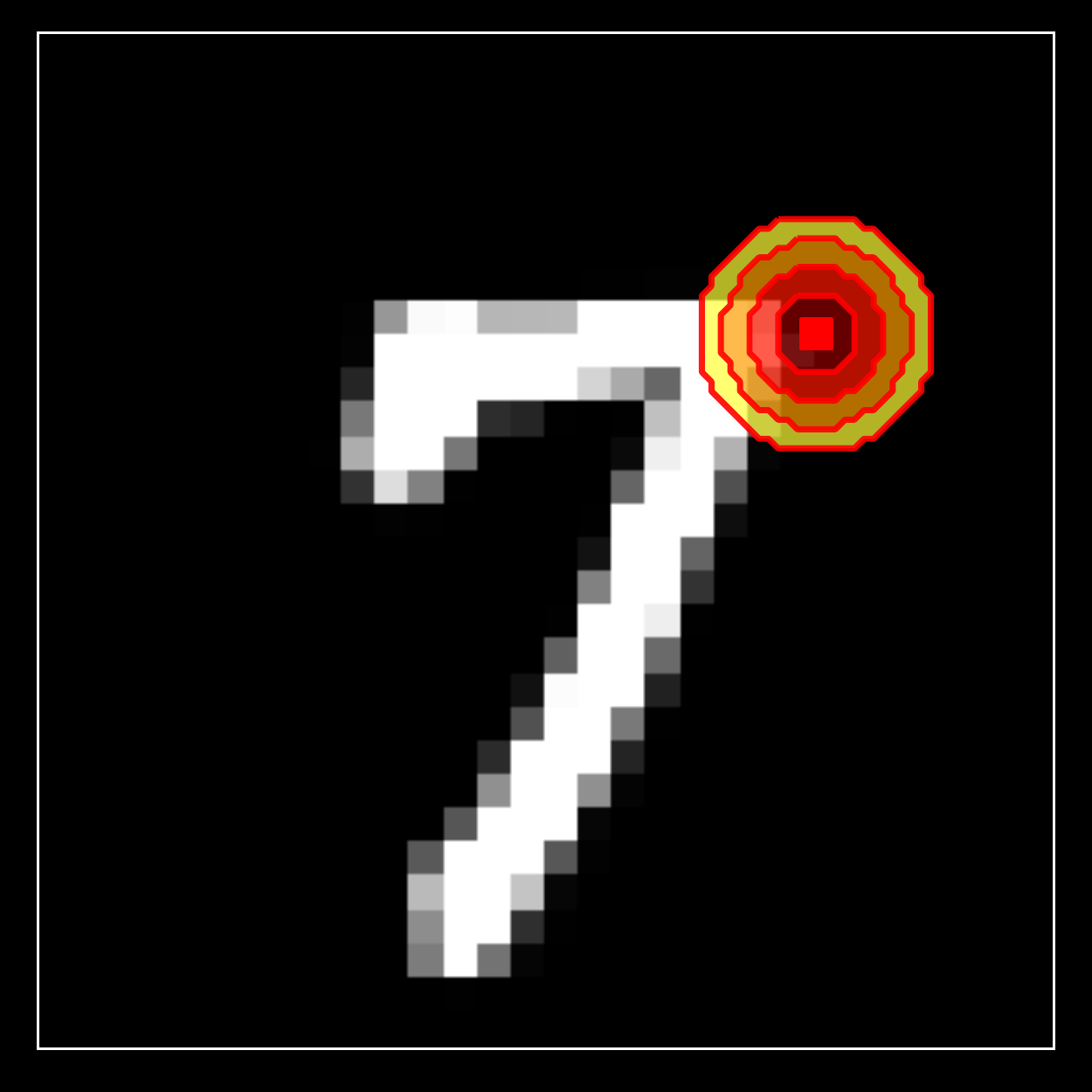}
        &
        \includegraphics[trim=10 10 10 10,clip,width=\panelwidth]{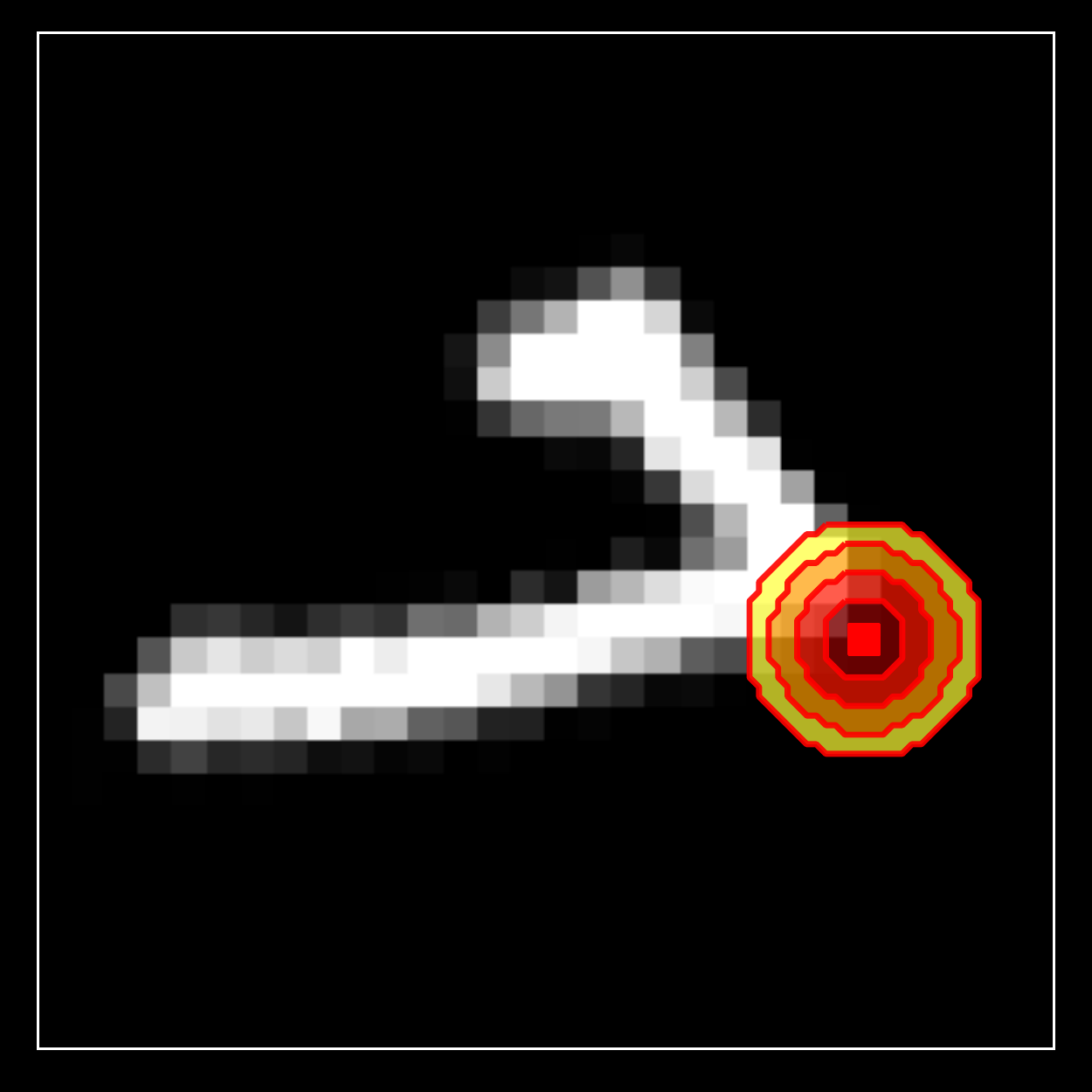}
        \\
        \includegraphics[trim=10 10 10 10,clip,width=\panelwidth]{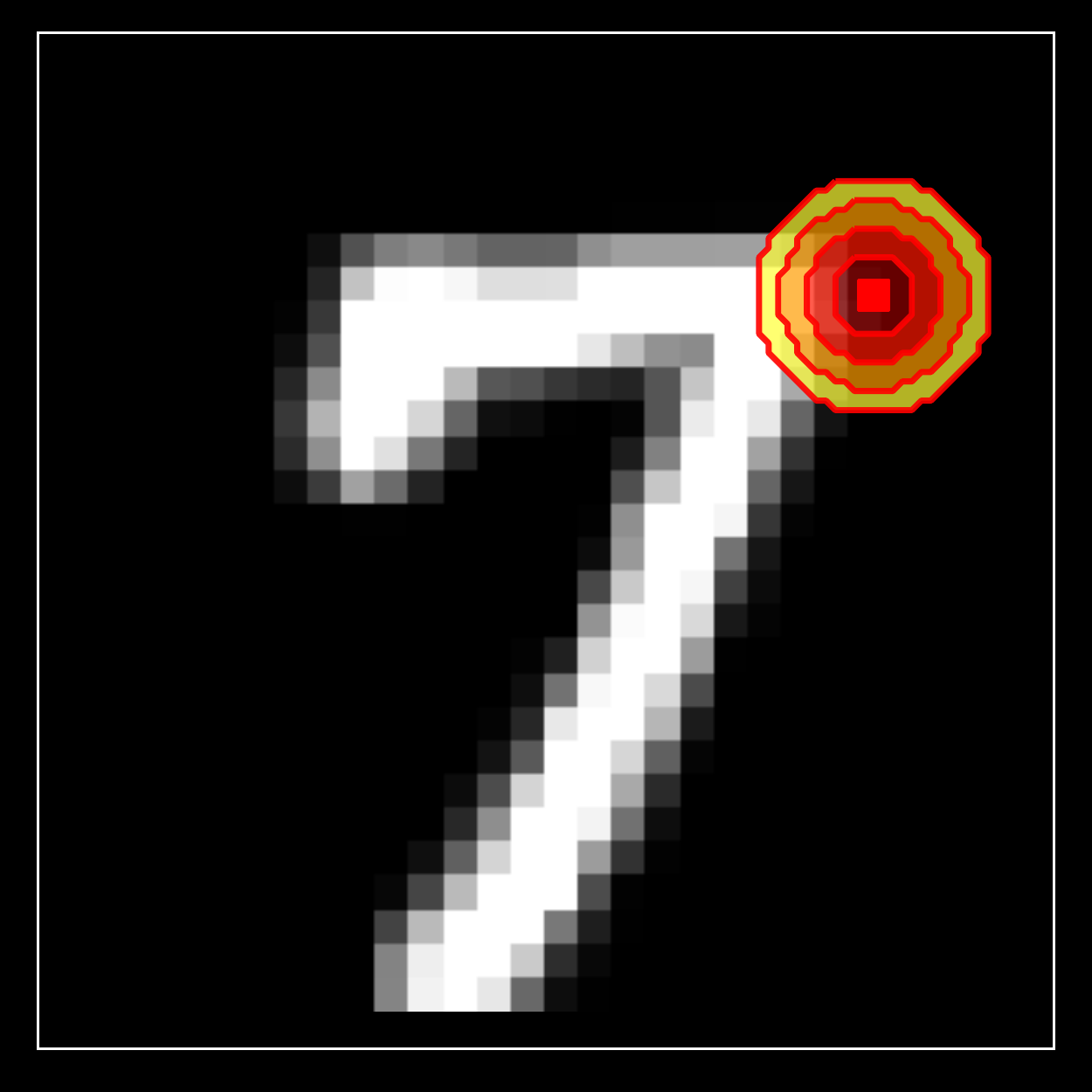}
        &
        \includegraphics[trim=10 10 10 10,clip,width=\panelwidth]{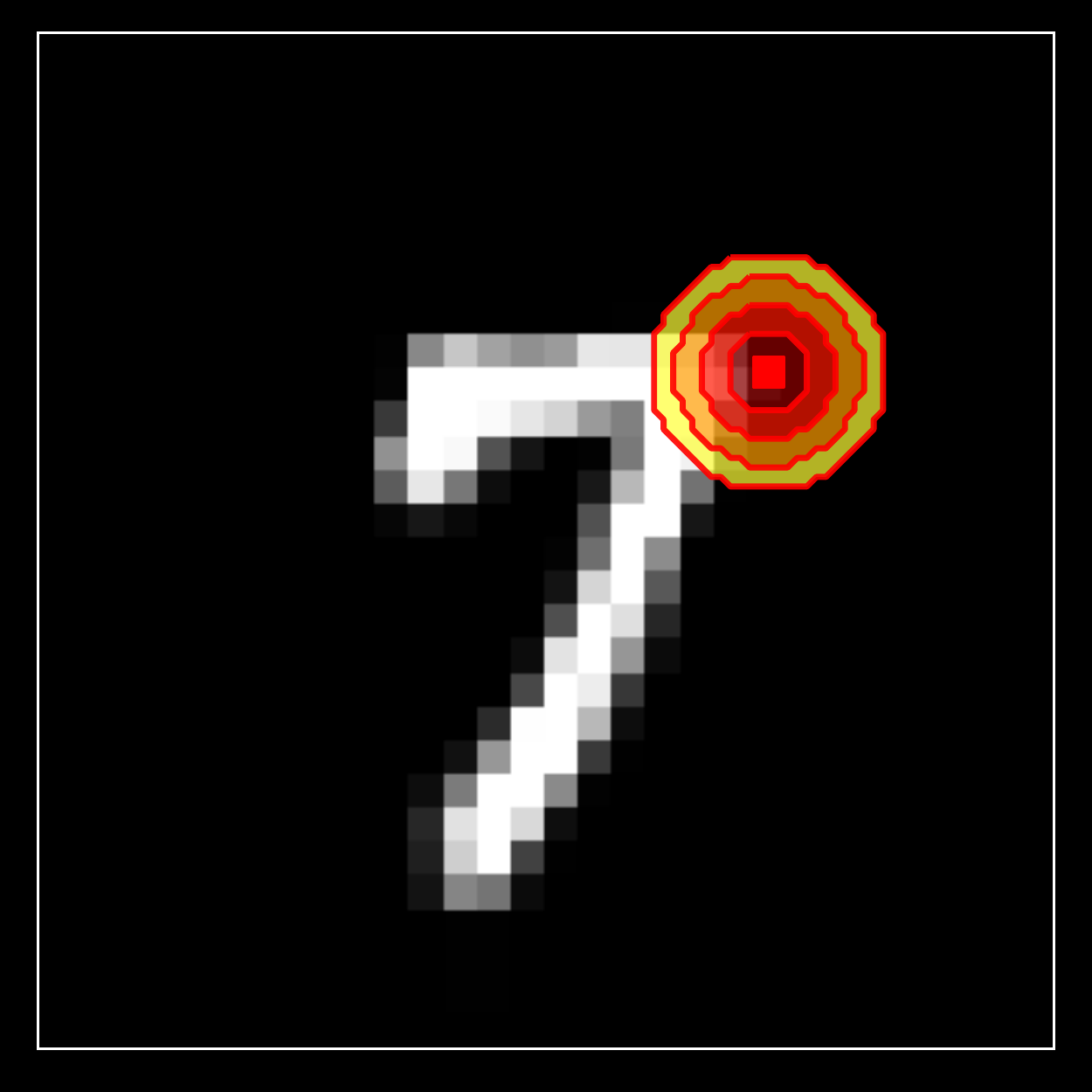}
        &
        \includegraphics[trim=10 10 10 10,clip,width=\panelwidth]{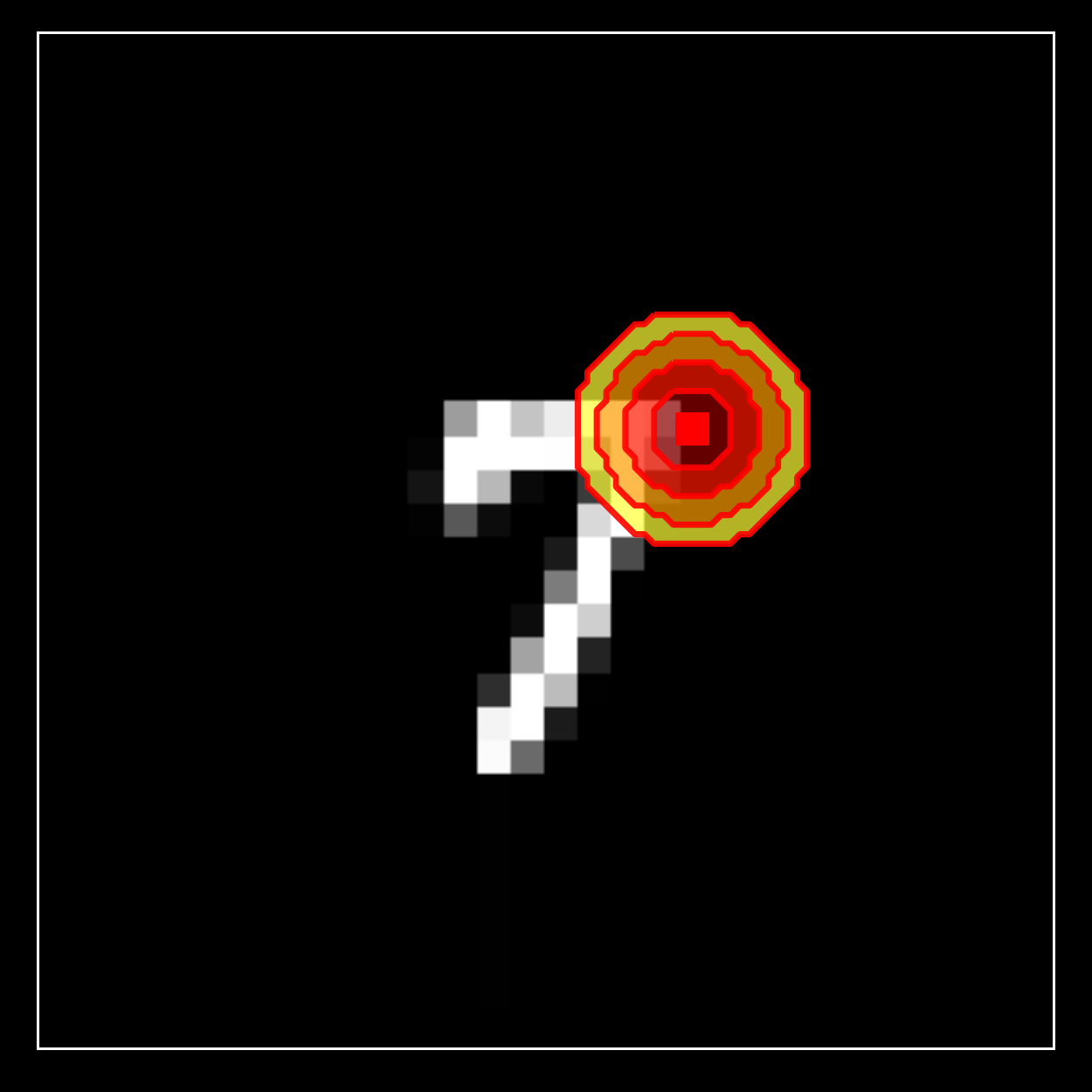}
        \\
    \end{tabular}
    }
    \caption{A visualization of the local neighborhood for different groups, in terms of the points in the input space. 
    For the computation of the convolution at the point in red, elements are sampled from colored region.
    In each panel, the top row shows translations, middle row shows rotations and
    bottom row shows scalings of the same image.
    For \SE2 we  visualize the distribution of points entering the computation 
    of the convolution over multiple lift samples.
    For each of the equivariant models that respects a given symmetry, the points that enter
    into the computation are not affected by the transformation.
    }
    \label{fig:all_nbhds}
\end{figure*}

\subsection{Local Neighborhood Visualizations}
\label{sec:app_nbhd}

In Figure \ref{fig:all_nbhds} we visualize the local neighborhood used with 
different groups under three different types of transformations: translations,
rotations and scaling.
The distance and neighborhood are defined for the tuples of group elements and orbit.
For Trivial, \T2, \SO2, $\mathbb R \times SO(2)$ the correspondence between points and these tuples
is one-to-one and we can identify the neighborhood in terms of the input points.
For \SE2 each point is mapped to multiple tuples, each of which defines its own
neighborhood in terms of other tuples.
In the Figure, for \SE2 for a given point we visualize the distribution of points
that enter the computation of the convolution at a specific tuple.

\end{document}